\documentclass{article}
\usepackage[preprint]{conference}
\usepackage{style}

\usepackage{wrapfig}
\usepackage{microtype}
\usepackage{booktabs}
\usepackage{hyperref}
\usepackage{url}
\usepackage{lineno}
\usepackage{amssymb}
\usepackage{adjustbox}
\usepackage{amsfonts}
\usepackage{amsmath}
\usepackage{amssymb}
\usepackage{bm}
\usepackage{booktabs}
\usepackage{colortbl}
\usepackage{comment}
\usepackage{dsfont}
\usepackage{enumitem}
\usepackage{graphicx} 
\usepackage{hyperref}
\usepackage{mathtools}
\usepackage{microtype}
\usepackage{multicol}
\usepackage{multirow}
\usepackage{nicefrac}
\usepackage{lscape}
\usepackage{scalerel}
\usepackage{subcaption}
\usepackage{tabularx}
\usepackage{url}
\usepackage{xcolor}
\usepackage{xspace}
\usepackage[T1]{fontenc}
\usepackage{wrapfig}
\usepackage{enumitem}
\usepackage{amssymb}
\usepackage{pifont}

\usepackage{tabularx}
\usepackage{makecell}
\usepackage{xcolor}
\usepackage{soul}
\usepackage{upgreek}
\usepackage[most]{tcolorbox}
\tcbuselibrary{breakable,skins,theorems}
\usepackage{caption}
\usepackage{bbm, dsfont}
\usepackage{placeins}
\usepackage{subcaption}
\usepackage{CJKutf8}
\usepackage{newfloat}
\usepackage{caption}
\DeclareCaptionType{example}[Example][List of Examples]
\usepackage{graphicx}
\usepackage{adjustbox}

\tcbset{
	exbox/.style={
		breakable,
		boxrule=0.9pt,
		arc=2pt,
		left=6pt,right=6pt,top=4pt,bottom=4pt, 
		before skip=6pt,after skip=0.1in,
	}
}

\newtcbtheorem[no counter]{hypobox}{}{
    separator sign none,
    enhanced,
    breakable, 
    colback=blue!3, 
    colframe=blue!65!black, 
    coltitle=black, 
    fonttitle=\bfseries, 
    width=\columnwidth,
    arc=2pt, 
    left=8pt, 
    right=8pt, 
    top=6pt, 
    bottom=6pt, 
    parskip=0pt,
    varwidth boxed title*=\columnwidth,
    attach boxed title to top left={xshift=6pt,yshift*=-2mm},
    boxed title style={ 
    enhanced, 
    colback=blue!12, 
    colframe=blue!75!black,
    left=2pt, 
    right=2mm, 
    top=2pt, 
    bottom=2pt, 
    }, 
}{hyp}

\newtcbtheorem[number within=section]{defbox}{Definition}{ enhanced,
    breakable, 
    colback=red!3, 
    colframe=red!65!black, 
    coltitle=black, 
    fonttitle=\bfseries, 
    width=\columnwidth,
    arc=2pt, 
    left=8pt, 
    right=8pt, 
    top=6pt, 
    bottom=6pt, 
    varwidth boxed title=\columnwidth,
    attach boxed title to top left={xshift=6pt,yshift*=-2mm},
    boxed title style={ 
    enhanced, 
    colback=red!12, 
    colframe=red!75!black,
    left=2pt, 
    right=2mm, 
    top=2pt, 
    bottom=2pt, 
    }, 
}{def}

\newcommand{\ohl}[1]{\sethlcolor{orange!25}\hl{#1}}

\newcommand{\ssc}[1]{{\small\textsc{#1}}}

\newcommand{\bssc}[1]{{\small \sc \textbf{#1}}\xspace}

\newcommand{\method}{\ssc{Unlock}\xspace}
\newcommand{\masterkey}{\ssc{MasterKey}\xspace}
\newcommand{\transfer}[2]{$#1\,+\,\text{\method}_{\mathrm{from\;}#2}$}

\newcommand{\SL}{\,\mathcal{S}^{\vphantom{*}}_\mathrm{L}\,}   %
\newcommand{\SU}{\,\mathcal{S}^{\vphantom{*}}_\mathrm{U}\,}   %
\newcommand{\TU}{\,\mathcal{T}^{\vphantom{*}}_\mathrm{U}\,}   %
\newcommand{\TL}{\,\mathcal{T}^{\vphantom{*}}_\mathrm{L}\,}   %
\newcommand{\TPT}{\,\mathcal{T}^*_\mathrm{PT}\,}   %
\newcommand{\mkey}[2]{\mathbf{v}^{(#1,#2)}_{\psi}}  %

\newcommand{\xmark}{\ding{55}}
\newcommand{\cmark}{\ding{51}}

\usepackage[capitalize,noabbrev]{cleveref}
\usepackage{tabularx}   %
\usepackage{array}      %
\usepackage{ragged2e}   %

\newcommand{\github}{\raisebox{-1.5pt}{\includegraphics[height=1.05em]{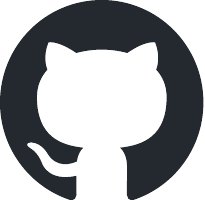}}\xspace}
\newcommand\blankfootnote[1]{%
  \let\thefootnote\relax\footnotetext{#1}%
  \let\thefootnote\svthefootnote%
}

\title{%
    \raisebox{-0.43\height}{%
        \includegraphics[height=2.8cm,width=2.5cm]{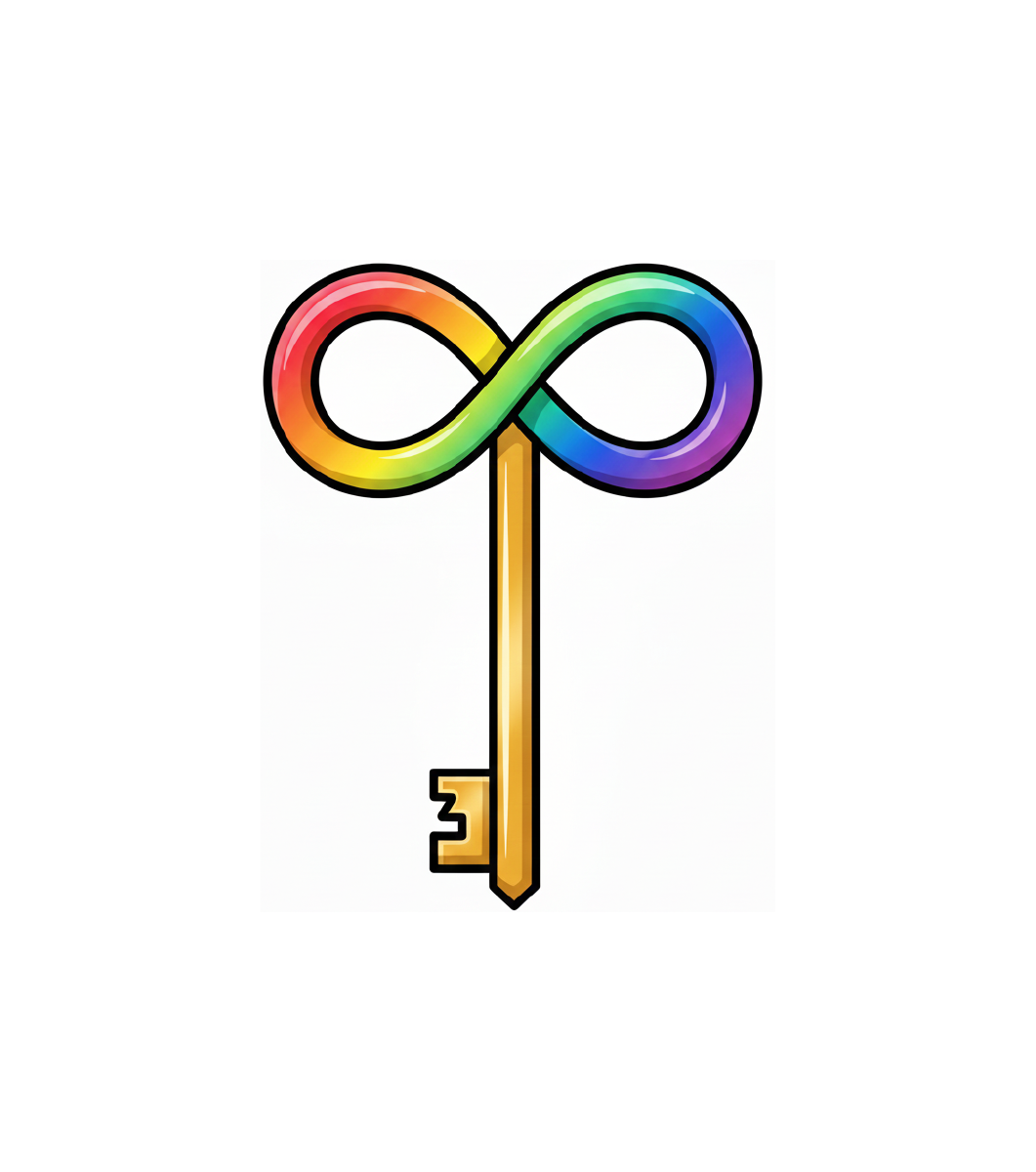}%
    }\hspace{0.3em}%
    \parbox[c]{0.65\linewidth}{%
        {\LARGE\sc\textbf{The Master Key Hypothesis:}}\\[0.3em]%
        {\Large Unlocking Cross-Model Capability Transfer%
        \ via Linear Subspace Alignment}%
    }%
}

\author{Rishab Balasubramanian$^\dagger$ 
\And Pin-Jie Lin$^\dagger$
\And Rituraj Sharma$^\dagger$
\And Anjie Fang$^\ddagger$
\AND Fardin Abdi$^\ddagger$
\And Viktor Rozgic$^\ddagger$
\And Zheng Du$^\ddagger$
\And Mohit Bansal$^\star$
\And Tu Vu$^\dagger$
\AND \normalfont\small 
$^\dagger$ Virginia Tech \quad
$^\ddagger$ Amazon \quad
$^\star$ UNC Chapel Hill
}

\begin{document}

\vspace*{-1cm}
\maketitle

{\vskip -0.2in
\centering
\github \url{https://github.com/rishabbala/Steering-Vector-Transfer}
\vskip 0.2in
}

\begin{abstract}
\begin{adjustbox}{minipage=0.8\textwidth, center}
We investigate whether post-trained capabilities can be transferred across model scales without retraining, and propose the \textbf{Master Key Hypothesis}, which states that model capabilities correspond to directions in a low-dimensional latent subspace that induce specific behaviors and are transferable across models through linear alignment.
Based on the hypothesis, we introduce \method, a \emph{training-free} and \emph{label-free} framework that extracts a capability direction by contrasting activations between capability-present and capability-absent Source variants, aligns it with a Target model through a low-rank linear transformation, and applies it at inference time to elicit the behavior.
Experiments on reasoning behaviors, including Chain-of-Thought (CoT) and mathematical reasoning, demonstrate substantial improvements across model scales without training. 
For example, transferring CoT reasoning from \texttt{Qwen1.5-14B} to \texttt{Qwen1.5-7B} yields an accuracy gain of 12.1\% on MATH, and transferring a mathematical reasoning direction from \texttt{Qwen3-4B-Base} to \texttt{Qwen3-14B-Base} improves AGIEval Math accuracy from 61.1\% to 71.3\%, surpassing the 67.8\% achieved by the 14B post-trained model.
Our analysis shows that the success of transfer depends on the capabilities learned during pre-training, and that our intervention amplifies latent capabilities by sharpening the output distribution toward successful reasoning trajectories. 

\end{adjustbox}
\end{abstract}

\section{Introduction}
Training modern language models involves two stages: pre-training, which instills general linguistic structure, and post-training, which aligns the model to desired behaviors.
Pre-training data often overlaps across model families and sizes, typically varying only in composition.
However, each new model often requires substantial data, computation, and engineering effort to instill useful behaviors in the post-training phase.
As models proliferate, this redundancy creates a fundamental inefficiency: \emph{capabilities are costly to learn, yet difficult to reuse across models.}
This inefficiency is compounded by evidence that post-training methods such as reinforcement learning do not incorporate new knowledge or reasoning capabilities, but rather act as a distribution-sharpening mechanism that pushes the base model towards narrow yet correct output trajectories~\citep{yue2025doesreinforcementlearningreally, wang20258020rulehighentropyminority, cui2025entropymechanismreinforcementlearning}.

To bridge this gap between pre- and post-training, prior work introduces reasoning and question answering data in between the two training stages, creating an additional \emph{mid-training} regime~\cite{olmo20252olmo2furious, wang2025octothinkermidtrainingincentivizesreinforcement, akter2025frontloadingreasoningsynergypretraining, liu2026midtrainingbridgespretrainingposttraining}.\footnote{In this work we use pre-training to collectively refer to both pre- and mid-training regimes.}
A mechanism that transfers capability-inducing representations across models to reliably elicit post-training behaviors without the need for retraining could reduce training costs, accelerate development, and enable modular reuse of existing model capabilities.

Concretely, we ask: \emph{can a desired capability that is expressed in one model be isolated and transferred to another model without gradient-based training or labeled supervision?}
We define a capability as a reproducible property of model behavior, such as step-by-step reasoning or mathematical problem solving.
The key challenge is that capabilities are implicitly encoded as high-dimensional representations, and successful transfer requires bridging differences in architecture, scale, and latent structure.\footnote{In the remainder of the paper, we use \emph{capability} and \emph{behavior} interchangeably.}

\begin{figure*}[t]
    \centering
    \begin{subfigure}[t]{0.49\textwidth}
        \centering
        \includegraphics[width=\linewidth]{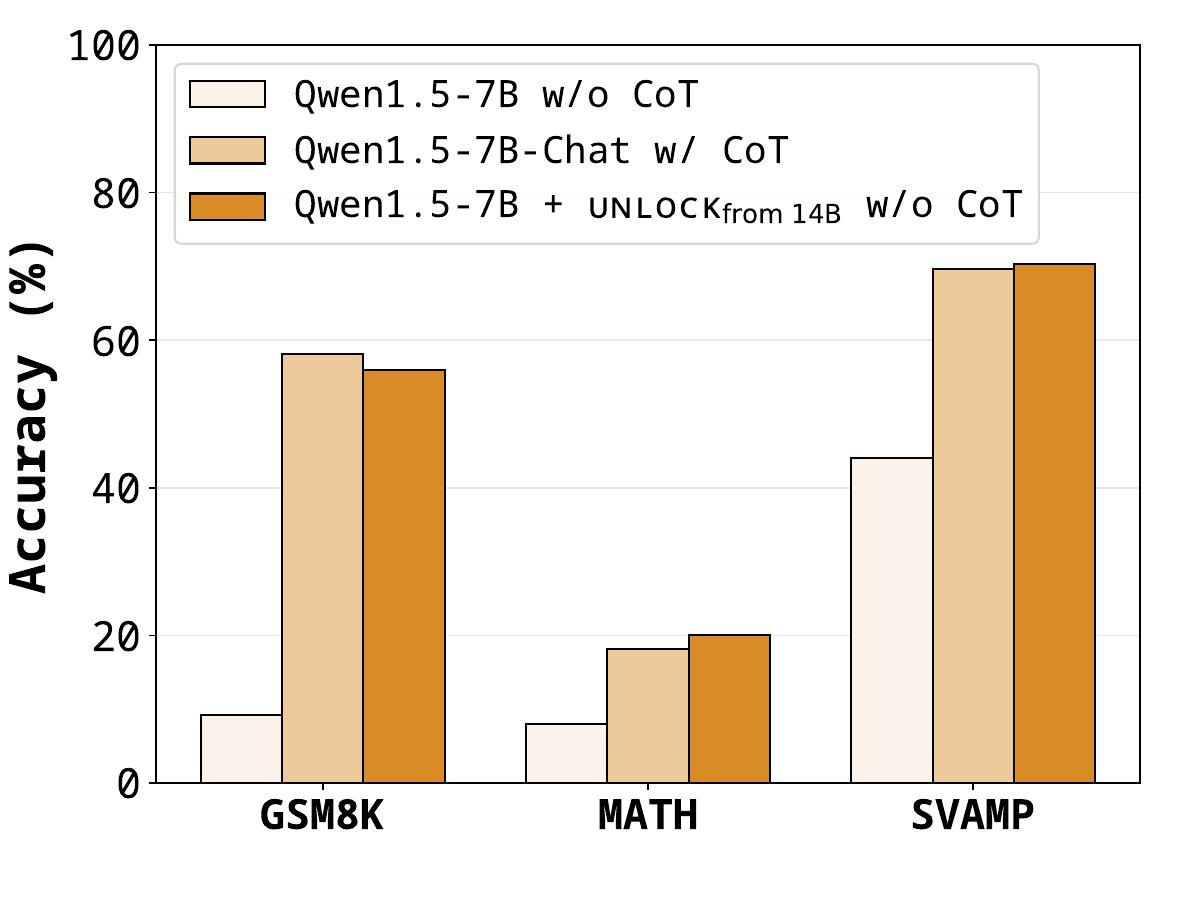}
        \label{fig:cot-subplot}
    \end{subfigure}
    \hfill
    \begin{subfigure}[t]{0.49\textwidth}
        \centering
        \includegraphics[width=\linewidth]{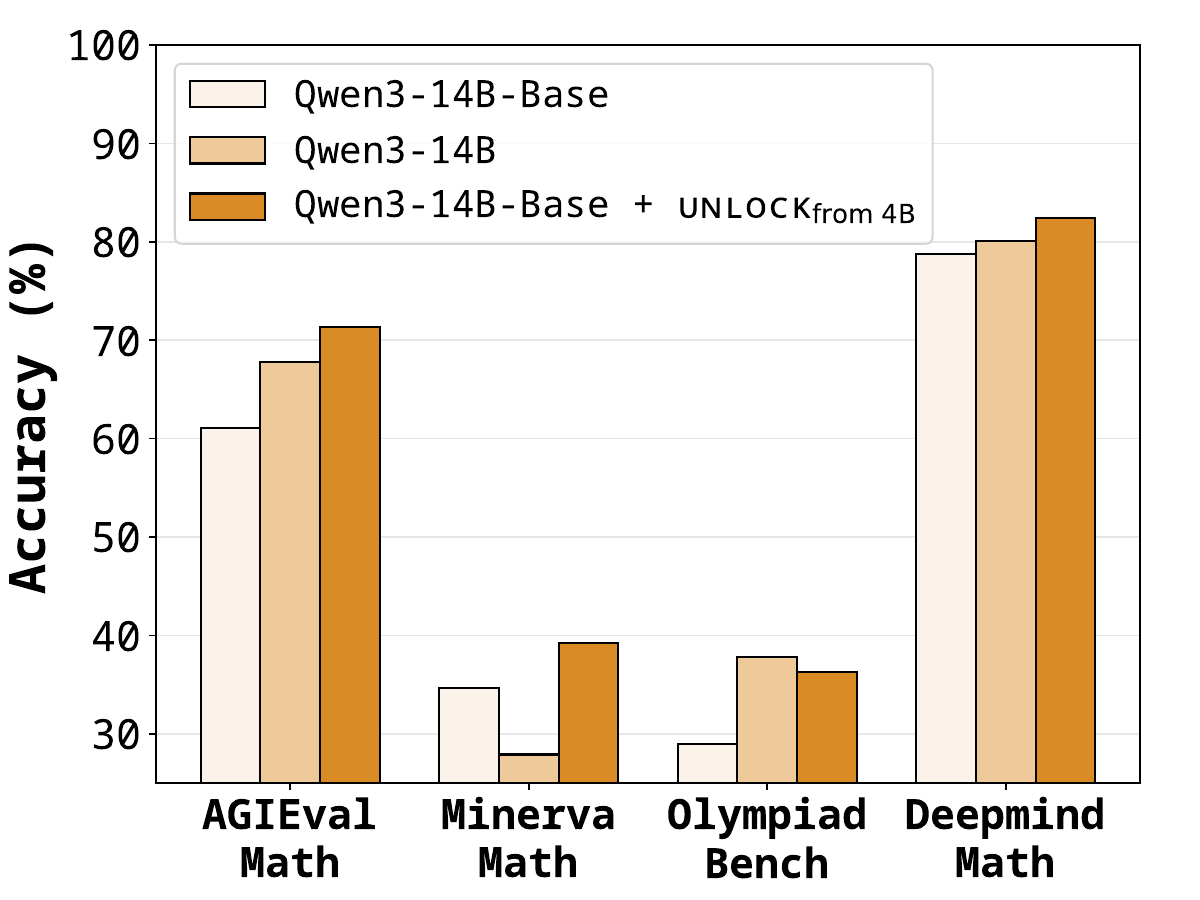}
        \label{fig:math-subplot}
    \end{subfigure}
    \vspace{-5mm}
    \caption{\textbf{Performance improvements from \mbox{\method}} when transferring (a) Chain-of-Thought capabilities from \texttt{Qwen1.5-14B} onto \texttt{Qwen1.5-7B}; and (b) Math reasoning capabilities from \texttt{Qwen3-4B-Base} onto \texttt{Qwen3-14B-Base}.
    Capability transfer substantially improves the base model \emph{without additional training}, approaching the performance of the post-trained model.}
    \label{fig:main_figure}
\end{figure*}

Existing methods for capability transfer can be decomposed into two steps:
(1) extracting a transformation from two Source variants that differ in an intended behavior --- arising either from different models (\textit{model-driven}, e.g., base and fine-tuned) or different prompting of the same model (\textit{prompt-driven}, e.g., with vs. without chain-of-thought prompting); 
and (2) applying this transformation to a Target model to reproduce the desired behavior.
These approaches differ primarily in the space in which this transformation is represented, and can be broadly categorized into three distinct frameworks --- 
(1) \textbf{Weight-space transfer}~\citep{ilharco2023editingmodelstaskarithmetic}:
The parameter-level difference between two Source models is added directly to the Target model, which typically requires architectural compatibility between the Source and Target models, or additional pruning or corrective training for cross-model alignment;
(2) \textbf{Output-space transfer}
~\citep{liu2024tuninglanguagemodelsproxy}:
The logit difference between two Source variants is applied at each generation step to adjust the output distribution of the Target model without modifying its parameters.
This avoids shape mismatches, but incurs substantial inference cost due to per-token logit computation for multiple models, and requires identical tokenization between the source and target; and
(3) \textbf{Latent-space transfer}~\citep{panickssery2024steeringllama2contrastive}: This strategy involves intervening on internal representations to steer the Target model towards the  desired output.

In this work, we focus on the \emph{latent space}, where capabilities are encoded as shifts in internal activations.
Existing methods typically construct steering directions from labeled contrastive examples (positive vs. negative \cite{turner2024steeringlanguagemodelsactivation, panickssery2024steeringllama2contrastive}) using a \emph{single} Source model and apply them at inference time to similar prompts.
Furthermore, these methods are largely focused on alignment and surface-level behavioral control (e.g., safety, toxicity, bias, and stylistic shaping \cite{liu2024incontextvectorsmakingcontext, tan2025analyzinggeneralizationreliabilitysteering, ghandeharioun2024patchscopesunifyingframeworkinspecting, chen2024selfieselfinterpretationlargelanguage, stolfo2025improvinginstructionfollowinglanguagemodels}) rather than advanced capabilities such as reasoning.

We address these limitations by proposing \method{} --- a \emph{training-free} and \emph{label-free} framework for cross-model capability transfer.
Our method involves three main stages: \emph{First}, we extract a \masterkey{} --- a capability direction in a Source model's representation space by contrasting internal activations between capability-present and capability-absent variants using a small set of unlabeled prompts (Section~\ref{sec:steer_vec}).
\emph{Second}, we estimate a low-rank linear transformation that aligns this direction with the latent space of a Target model (Section~\ref{sec:low_dim_match}).
\emph{Finally}, we apply the transferred direction as a normalized inference-time intervention to elicit the corresponding behavior (Section~\ref{sec:unlock_target}).
The entire procedure is \emph{training-free, label-free, architecture-agnostic}, and requires \emph{only forward passes}. 

As case studies, we evaluate our capability transfer approach on reasoning behaviors, including Chain-of-Thought (Section~\ref{sec:cot-main}) and mathematical reasoning (Section~\ref{sec:math-main}).
We find that the transferred capability leads to substantial performance improvements, and can match the gains from post-training.
As shown in Figure~\ref{fig:main_figure}, transferring a Chain-of-Thought (CoT) direction from \texttt{Qwen1.5-14B} to \texttt{Qwen1.5-7B} improves the accuracy on MATH from $8.0\%$ to $20.1\%$ \emph{without} explicit CoT prompting, which outperforms the $18.2\%$ achieved by the 7B instruction-tuned model \emph{with} CoT prompting.
Notably, transferring mathematical reasoning from \texttt{Qwen3-4B} to \texttt{Qwen3-14B} improves AGIEval Math accuracy from $61.1\%$ to $71.3\%$, which surpasses the $67.8\%$ achieved by the 14B instruction-tuned model.
Lastly, we provide preliminary experiments for cross-family transfer of CoT behavior (Appendix~\ref{sec:model-family}), and observe consistent performance gains, offering initial evidence for the convergence of capability representations across model families as postulated by~\cite{huh2024platonic}.

Our analysis reveals several consistent patterns.
First, capability transfer exhibits a directional asymmetry: small-to-large transfer typically yields larger relative improvements than large-to-small transfer.
Second, for both CoT and mathematical reasoning, \method{} amplifies capabilities that are dormant in the model, yielding greater gains when those capabilities are more strongly represented.
Finally, we provide evidence that \method{} sharpens the output distribution and directs generation toward reasoning trajectories that are more likely to succeed --- in line with the findings from~\cite{yue2025doesreinforcementlearningreally}.
Based on our results and observations, we introduce the \emph{Master Key Hypothesis} below and defer a formal definition to Section~\ref{sec:hypothesis-sec}.

\begin{hypobox}{The Master Key Hypothesis}{informal-hypothesis}
Capabilities exist as directions in a low-dimensional latent subspace, such that shifting the representations along these directions induces the desired behaviors.
These directional vectors can then be isolated and mapped into the representation space of another model via linear transformations, thus eliciting the intended capability.
\end{hypobox}

To summarize, our main contributions are:
\begin{itemize}
    \item \textbf{Training-free Capability Transfer:}  We propose a method that extracts a capability-inducing direction (\masterkey) from a pair of Source models and transfers it to a Target model via low-rank linear subspace alignment, enabling capability reuse without additional training.
    \item \textbf{The Master Key Hypothesis:} We hypothesize that model capabilities correspond to directions in a shared low-dimensional latent subspace, which can be isolated and transferred across models via linear transformations.
    \item \textbf{Empirical Validation Across Model Sizes:} Through extensive experiments, we demonstrate that reasoning behaviors, including Chain-of-Thought and mathematical reasoning, can be transferred across models of different sizes, yielding substantial improvements that approach or match gains typically obtained through post-training.
    \item \textbf{Analysis of Transfer Dynamics:} We analyze the factors that influence capability transfer, namely, the effect of model family and scale on transfer, and the effect of steering on the model's output distribution.
\end{itemize}

\section{Method}
\label{sec:method}

\begin{figure*}
    \centering
    \includegraphics[width=\textwidth,height=0.6\textwidth]{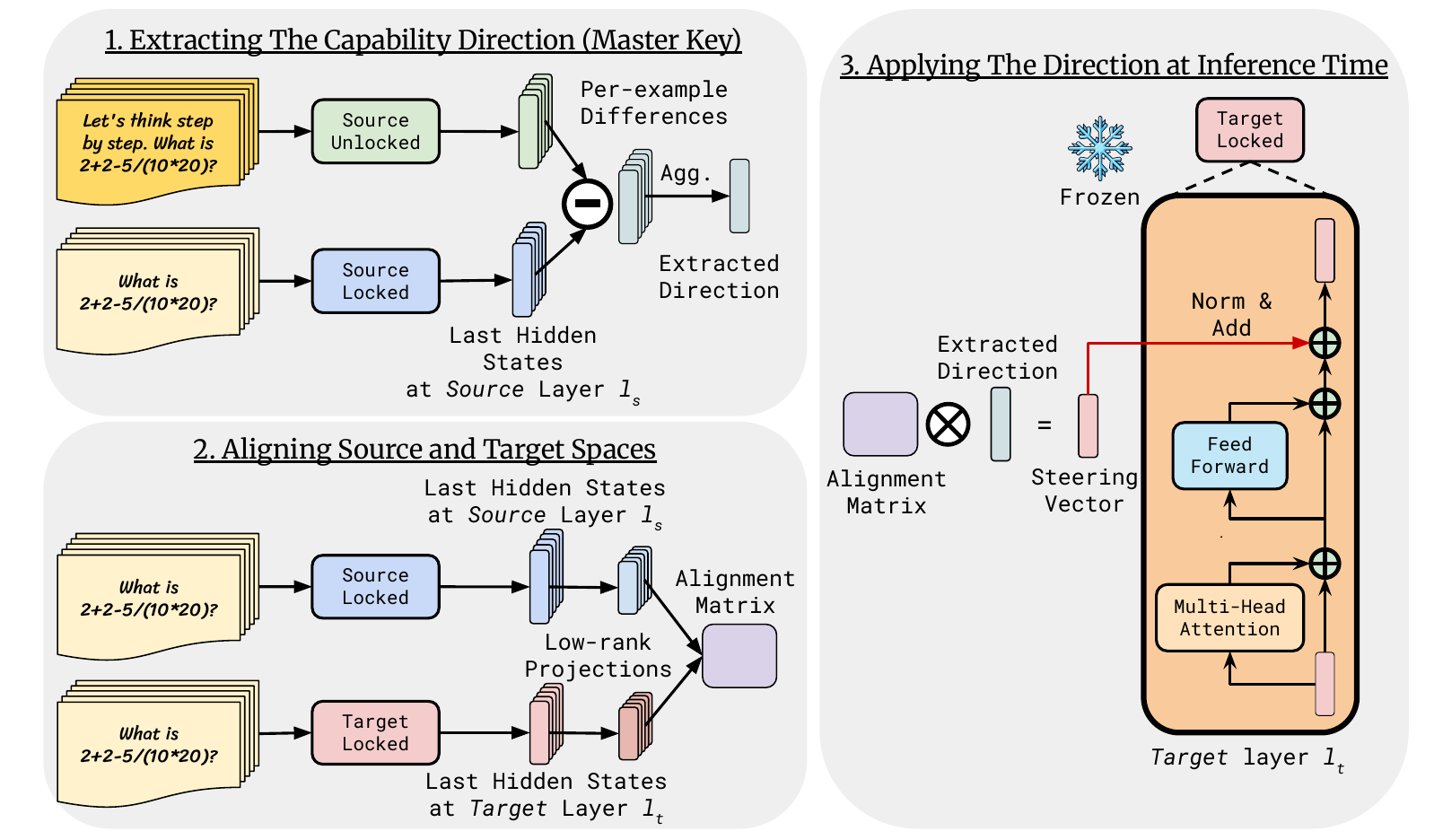}
    \caption{\textbf{Illustration of \method:} Our method consists of three stages: (1) Calculating the difference in hidden states in the Source space; (2) Learning a linear transformation between the Source Locked and Target Locked models; and (3) Projecting the \masterkey from Source to Target space and applying as a test-time intervention to the residual stream at every layer.}
    \label{fig:method}
\end{figure*}

In this section, we introduce \method, a training-free and label-free framework for transferring capability-inducing directions across models.
The core idea is to represent a capability as a direction in representation space that shifts a model from a state where the behavior is weak or absent to one where it reliably emerges.
\method extracts this direction (referred to as \masterkey) by contrasting two Source variants that differ in the presence of the capability (e.g., \texttt{Qwen3-4B-Base} vs. its post-trained counterpart \texttt{Qwen3-4B}) and transfers it to a Target model (e.g., \texttt{Qwen3-14B-Base}) to elicit the behavior.

\newpage
The framework involves three conceptual models:
\begin{itemize}
    \item \textbf{Source Locked $\SL$:} a Source variant in which the desired capability is weak or absent.
	\item \textbf{Source Unlocked $\SU$:} a Source variant that reliably exhibits the capability.
	\item \textbf{Target Locked $\TL$:} the Target model in which we aim to elicit the capability.
\end{itemize}

The Source variants $\SL$ and $\SU$ share the same architecture and tokenizers, allowing their internal activations to be directly compared. Their contrast isolates a capability direction in the \emph{Source representation space}. This direction is then mapped into the \emph{Target representation space} and applied during inference to $\TL$, producing the \textbf{Target Unlocked} model $\TU$.

At a high level, \method consists of \emph{three} stages: 
\begin{itemize}
    \item extracting a capability direction (\masterkey) from the Source variants (Section~\ref{sec:steer_vec}),
	\item aligning representation spaces between the Source and Target models (Section~\ref{sec:low_dim_match}), and 
	\item applying the transferred direction as an inference-time intervention (Section~\ref{sec:unlock_target}).
\end{itemize}

The entire procedure requires only forward passes on a small set of unlabeled prompts.

\subsection{Problem Setup}
\label{sec:problem_setup}

Let $\mathcal{M}$ denote a language model with hidden dimension $d^{\vphantom{*}}_{\mkern-3mu\mathcal{M}}$. For a \emph{model-specific} prompt $P^{\vphantom{*}}_{\mkern-3mu\mathcal{M}}$ (e.g., task instructions and/or a few demonstrations) and query $q^{\vphantom{*}}_i$, we denote the final-token hidden state at layer $l$ as 
\[
\mathbf{h}_{\mathcal{M}}^{(l)}\:(P^{\vphantom{*}}_{\mkern-3mu\mathcal{M}} \oplus q^{\vphantom{*}}_i) \in \mathbb{R}^{d^{\vphantom{*}}_{\mkern-3mu\mathcal{M}}},
\]
where $\oplus$ denotes sequence concatenation. We assume access to a small set of unlabeled queries $\mathcal{D} = \{q^{\vphantom{*}}_i\}_{i=1}^{n}$. These queries are used both to extract the capability direction and to estimate the cross-model alignment.

\subsection{Extracting The \bssc{MasterKey}}
\label{sec:steer_vec}
We first isolate a direction from the Source variants that elicits the desired capability $\psi$. Intuitively, the difference in their internal representations captures the shift required to induce the behavior.
For each query $q^{\vphantom{*}}_i$ and layer $l$, we compute a \emph{per-example} representation difference between the Unlocked and Locked representations
\begin{equation}
v^{(\mathcal{S}, l, i)}_\psi \;=\; \mathbf{h}_{\SU}^{(l)}(P^{\vphantom{*}}_{\mkern-3mu\SU} \mkern-6mu\oplus q^{\vphantom{*}}_i) \;-\; \mathbf{h}_{\SL}^{(l)}(P^{\vphantom{*}}_{\mkern-3mu\SL} \mkern-6mu\oplus q^{\vphantom{*}}_i).
\end{equation} 
This vector represents the activation shift required to move the Locked model towards the Unlocked behavior for that example. To obtain a dataset-level direction, we aggregate these differences across queries as 
\(
\mkey{\mathcal{S}}{l,\Phi} \;=\; \Phi\left(v^{(\mathcal{S}, l, i)}_\psi\right)
\)
where $\Phi$ is any aggregator function.
In this work, we consider two aggregator functions --- the \emph{mean} aggregator, defined as the average of the differences
\vspace{-2pt}
\begin{equation}
\mkey{\mathcal{S}}{l,\text{avg}} \;=\; \frac{1}{n}\sum_{i=1}^{n}v^{(\mathcal{S}, l, i)}_\psi,
\end{equation}
and the \emph{principal component} aggregator, defined as the first principal component of the centered differences~\citep{nguyen2025grains}
\vspace{-2pt}
\begin{equation}
\mkey{\mathcal{S}}{l,\text{pca}} \;=\; \operatorname{PCA}^{\vphantom{*}}_{1}\!\left(\left\{v^{(\mathcal{S}, l, i)}_\psi - \mkey{\mathcal{S}}{l,\text{avg}}\right\}_{i=1}^{n}\right).
\end{equation}

Our formulation is entirely unsupervised and requires no labeled supervision (e.g., positive or negative examples). The contrast between the Source variants may arise from \textit{prompt-driven} differences (e.g., with vs. without CoT prompting) or \textit{model-driven} differences (e.g., base vs. post-trained models).

\subsection{Cross-model Subspace Alignment}
\label{sec:low_dim_match}
The capability direction \masterkey extracted in Section~\ref{sec:steer_vec} lies in the Source representation space. Since the Target model may have a different hidden size and latent geometry, we compute a mapping that transfers this \masterkey onto the Target representation space.

Specifically, we first collect hidden representations from the Source Locked model $\SL$ and the Target Locked model $\TL$. Let $l^{\vphantom{*}}_s$ denote the layer of $\SL$ used to extract Source representations, and $l^{\vphantom{*}}_t$ the layer of $\TL$ where the transferred \masterkey will be applied (Section~\ref{sec:unlock_target} describes how $l^{\vphantom{*}}_s$ is selected for a given $l^{\vphantom{*}}_t$). To reduce prompt-induced variance, we use a shared prompt $p = P^{\vphantom{*}}_{(\SL,\TL)}$ for both models.\footnote{We note that the shared prompt is to minimize noise from the prompts. Our framework theoretically allows any combination of prompts to be applied between $\SL$ and $\TL$.} Given the query set $\mathcal{D} = \{q^{\vphantom{*}}_i\}_{i=1}^{n}$, we stack representations across queries for both models to obtain two matrices
\begin{equation*}
\begin{aligned}
X_{\mathcal{S}} &=
\begin{bmatrix}
\big(\mathbf{h}_{\SL}^{(l_s)}(p \oplus q^{\vphantom{*}}_1)\big)^\top\\[5pt]
\big(\mathbf{h}_{\SL}^{(l_s)}(p \oplus q^{\vphantom{*}}_2)\big)^\top\\[5pt]
\vdots\\[5pt]
\big(\mathbf{h}_{\SL}^{(l_s)}(p \oplus q^{\vphantom{*}}_n)\big)^\top\\[5pt]
\end{bmatrix}
\in\mathbb{R}^{n\times d^{\vphantom{*}}_{\mathcal{S}}},
\qquad
X_{\mathcal{T}} &=
\begin{bmatrix}
\big(\mathbf{h}_{\TL}^{(l_t)}(p \oplus q^{\vphantom{*}}_1)\big)^\top\\[5pt]
\big(\mathbf{h}_{\TL}^{(l_t)}(p \oplus q^{\vphantom{*}}_2)\big)^\top\\[5pt]
\vdots\\[5pt]
\big(\mathbf{h}_{\TL}^{(l_t)}(p \oplus q^{\vphantom{*}}_n)\big)^\top\\[5pt]
\end{bmatrix}
\in\mathbb{R}^{n\times d^{\vphantom{*}}_{\mathcal{T}}},
\end{aligned}
\end{equation*}
where $d^{\vphantom{*}}_{\mathcal{S}}$ and $d^{\vphantom{*}}_{\mathcal{T}}$ denote the hidden size of the Source and Target models, respectively, and $n \ll \min(d^{\vphantom{*}}_{\mathcal{S}}, d^{\vphantom{*}}_{\mathcal{T}})$.
Instead of aligning the full hidden spaces, we align \emph{low-rank} subspaces that capture the dominant structure of the representations.
To find these low-rank subspaces, we perform Singular Value Decomposition (SVD) on both matrices
\[
X^{\vphantom{*}}_{\mathcal{S}} = U^{\vphantom{*}}_{\mathcal{S}}\Sigma^{\vphantom{*}}_{\mathcal{S}}V_{\mathcal{S}}^{\top}, \qquad
X^{\vphantom{*}}_{\mathcal{T}} = U^{\vphantom{*}}_{\mathcal{T}}\Sigma^{\vphantom{*}}_{\mathcal{T}}V_{\mathcal{T}}^{\top},
\]
and retain the top-$k$ right singular vectors $V_{\mathcal{S}}^{k}\in\mathbb{R}^{d_{\mathcal{S}}\times k}$, $V_{\mathcal{T}}^{k}\in\mathbb{R}^{d_{\mathcal{T}}\times k}$, where $k \leq n$. 
Projecting the representations into these subspaces yields
\[
\hat{X}^{\vphantom{*}}_{\mathcal{S}} = X^{\vphantom{*}}_{\mathcal{S}}V_{\mathcal{S}}^{k}\in\mathbb{R}^{n\times k},
\qquad
\hat{X}^{\vphantom{*}}_{\mathcal{T}} = X^{\vphantom{*}}_{\mathcal{T}}V_{\mathcal{T}}^{k}\in\mathbb{R}^{n\times k}.
\]
We then learn a linear transformation $W\in\mathbb{R}^{k\times k}$ that aligns the projected representations by minimizing the Frobenius norm loss
\begin{equation}
W^{\star} \;=\; \arg\min_{W\in\mathbb{R}^{k\times k}} \left\lVert \hat{X}^{\vphantom{*}}_{\mathcal{S}} W - \hat{X}^{\vphantom{*}}_{\mathcal{T}} \right\rVert_{F}^{2}.
\end{equation}

This problem has the closed-form solution $W^{\star}=\hat{X}_{\mathcal{S}}^{\dagger}\hat{X}_{\mathcal{T}}^{\vphantom{\dagger}} \in\mathbb{R}^{k\times k}$, where $\lVert\cdot\rVert_F$ denotes the Frobenius norm and $\hat{X}_{\mathcal{S}}^{\dagger}\in\mathbb{R}^{k\times n}$ is the Moore--Penrose pseudoinverse of $\hat{X}^{\vphantom{*}}_{\mathcal{S}}$. 

Using this alignment, we define a lifted cross-model operator
\begin{equation}
R_{l_t} \;=\; V_{\mathcal{T}}^{k}\,(W^{\star})^{\top}\,(V_{\mathcal{S}}^{k})^{\top} \in \mathbb{R}^{d_{\mathcal{T}}\times d_{\mathcal{S}}},
\end{equation}

which maps vectors from the Source representation space to the Target representation space.
Applying this operator to the source \masterkey $\mkey{\mathcal{S}}{l_s, \Phi} \in\mathbb{R}^{d_{\mathcal{S}}}$ yields the transferred capability direction
\begin{equation}
\label{eqn:transform}
\mkey{\mathcal{T}}{l_t,\Phi} \;=\; R^{\vphantom{*}}_{l_t}\,\mkey{\mathcal{S}}{l_s,\Phi} \in\mathbb{R}^{d_{\mathcal{T}}}.
\end{equation}

\subsection{Unlocking The Target Model}
\label{sec:unlock_target}
Since the Source and Target models may have different depths, we align layers by relative position. Let $L_{\mathcal{S}}$ and $L_{\mathcal{T}}$ denote the number of layers in the Source and Target models, respectively. For each target layer $l_t$, we choose the corresponding Source layer as
\[
l_s \;=\; \min\!\left(L_{\mathcal{S}},\ \max\!\left(1,\ \left\lfloor \frac{L_{\mathcal{S}}}{L_{\mathcal{T}}}\, l_t \right\rfloor \right)\right).
\]
This mapping aligns layers at similar relative depths, following prior observations that representations maintain structural relationships across model scales~\citep{csordas2025do}.

For a new input query, we compute the \masterkey in Target space using Equation~\ref{eqn:transform}, and  apply this direction during inference to steer the hidden representations of the Target model $\TL$. 
At each layer $l_t$, the final-token hidden state $\mathbf{h}_{\TL}^{(l_t)}$ is modified as

\begin{equation*}
\begin{aligned}
\tilde{\mathbf{h}}_{\TL}^{(l_t)} = \frac{\mathbf{h}_{\TL}^{(l_t)}}{\left\lVert \mathbf{h}_{\TL}^{(l_t)}\right\rVert_{2}} \;+\; \alpha\frac{\mkey{\mathcal{T}}{l_t,\Phi}}{\left\lVert \mkey{\mathcal{T}}{l_t,\Phi}\right\rVert_{2}},
\end{aligned}
\end{equation*}
where $\alpha$ controls the strength of the intervention.
The resulting vector is rescaled to preserve the magnitude of the original hidden state.
\begin{equation}
\mathbf{h}_{\TU}^{(l_t)} = \frac{\tilde{\mathbf{h}}_{\TL}^{(l_t)}}{\left\lVert \tilde{\mathbf{h}}_{\TL}^{(l_t)}\right\rVert_{2}} \times \left\lVert \mathbf{h}_{\TL}^{(l_t)}\right\rVert_{2}.
\end{equation}
Applying this intervention across layers during generation yields the Target Unlocked model $\TU$, which exhibits the transferred capability without requiring additional training.
Figure~\ref{fig:method} shows the three stages of our method and how the intervention is applied at test-time.

We denote the transfer of a specific capability $\psi$ from the Source pair of models to the Target model as $\TL\,+\,\text{\method}_{\mathrm{from\;}\SL}$.\footnote{Since we always deploy a base version for the Locked models, we use the model name and size to represent $\SL$ and $\TL$, and drop additional suffixes such as \texttt{-Base} or \texttt{-pt}.}
To avoid confusion, we refer to the Target model that has undergone extensive post-training as the \textbf{post-trained Target} model $\TPT$.
We treat the aggregation method $\Phi$, subspace rank $k$, number of queries $n$, and steering strength $\alpha$ as hyperparameters, which are selected via grid search on a held-out development set.
We provide a discussion of these hyperparameters and the low-rank nature of the capability subspace in Appendix~\ref{sec:hyperparams-cot}.

\section{Atomic And Non-Atomic Capabilities}

We formally define a capability as:

\begin{defbox}{Capability}{capability}
For a model $\mathcal{M}$, a query distribution $\mathcal{Q}$, and a prompt distribution $\mathcal{P}^{\vphantom{*}}_{\mkern-5mu\mathcal{M}}$, a \textbf{capability} $\psi$ is defined as a measurable behavior associated with a metric $m^{\vphantom{*}}_\psi\left(\mathcal{M}; q, p\right)$. We say the capability $\psi$ is present in $\mathcal{M}$ if the aggregate performance 
\[ \mathbb{E}^{\vphantom{*}}_{q \sim \mathcal{Q}, \, p \sim \mathcal{P}^{\vphantom{*}}_{\mkern-5mu\mathcal{M}}} \left[ m^{\vphantom{*}}_\psi(\mathcal{M}; q, p) \right] \geq \ \tau \]
remains significantly above a baseline $\tau$ and is robust to minor semantic perturbations in the query and prompt spaces.
\end{defbox}

Conceptually, we define a capability as any model behavior that can be consistently observed across a distribution of semantically similar inputs and prompt templates, remaining invariant to minor changes in them.
We further distinguish between capabilities that are \emph{latent} within a model (elicitable via steering or prompting) and those that are \emph{absent} (requiring explicit training to acquire).

\begin{defbox}{(Non-)Atomic Capabilities}{atomicity}
Let $\mathcal{M}$ be a base model and $\mathcal{M}^{*}$ its post-trained variant. Let $q \sim \mathcal{Q}$ and $p \sim \mathcal{P}$ denote queries and prompts drawn from their respective evaluation distributions. For a capability $\psi$ present in $\mathcal{M}^{*}$ and measurable with metric $m_\psi(\mathcal{M}; q, p)$, we define the average post-training gain as:
\[
\Delta^{\mathcal{Q}}_\psi = \mathbb{E}^{\vphantom{*}}_{q \sim \mathcal{Q}, \, p \sim \mathcal{P}} \left[ m^{\vphantom{*}}_\psi(\mathcal{M}^{*}; q, p) - m^{\vphantom{*}}_\psi(\mathcal{M}; q, p) \right].
\]

For a small threshold $\epsilon\ge 0$, we say that $\psi$ is an \textbf{atomic capability} if $\Delta^{\mathcal{Q}}_\psi \leq \epsilon$ (i.e. the capability is present to nearly the same degree before and after post-training), and a \textbf{non-atomic capability} if $\Delta^{\mathcal{Q}}_\psi > \epsilon$ (i.e. post-training yields a substantial improvement that cannot be explained by an already-present behavior).

\end{defbox}

Under this view, atomicity is inherently relative to a model's pre-training distribution: a capability is atomic only to the extent that it is supported by the data and objectives encountered during pre-training. 
In Sections~\ref{sec:cot-main},~\ref{sec:math-main} we show that the atomicity of the capability impacts the gains from \method.

The atomicity of a capability also depends on the learning capacity of the language model and thus would be impacted by size and architecture.
While we explore the effects of model scale on transferability (Appendix~\ref{sec:cot-appendix}), we leave a more systematic study on the impact of architectures to future work.
Lastly, atomicity also depends on the nature of the data. In this work we focus on transferring post-training capabilities onto a base model version, and thus we consider the capabilities present within the base model version (i.e. learned during pre-training).
When transferring capabilities between two post-trained models, the definitions above should be modified to reflect this change in data distribution.
We emphasize that atomicity is a function of not only the capability required, but also the architecture, scale, and data.
We intentionally leave Definitions~\ref{def:capability},~\ref{def:atomicity} vague to reflect this gap in our understanding of the representation space in language models.

\section{Atomic Capability Transfer}
\label{sec:cot-main}

Having established the necessary framework for transferring capability-inducing directions across models, we now ask whether such directions can be extracted from prompt-induced representational changes within a single model (i.e., $\SU \equiv \SL$) and then transferred across model scales.
This setting provides a controlled test of the Master Key Hypothesis: since the model weights remain fixed, any behavioral change must arise from shifts in internal representations. If a capability corresponds to a direction in representation space, then contrasting activations from prompts that encourage the capability and those that do not should reveal the corresponding direction.
We study this question using Chain-of-Thought (CoT) reasoning, which often emerges in sufficiently capable base language models and can be elicited through prompting alone.
We therefore treat CoT as an \emph{atomic capability}, meaning that the underlying reasoning ability is already present in the base model but is not always expressed without the appropriate prompt.
Empirically, we find that \method makes step-by-step thinking more consistently expressed, improving reasoning performance across model families and benchmarks even in the absence of explicit CoT prompting.

\subsection{Experimental Setup}
We evaluate prompt-induced capability transfer across model scales within five model families: \texttt{Qwen1.5}~\citep{bai2023qwen}, \texttt{Qwen2.5}~\citep{yang2025qwen25}, \texttt{Qwen3}~\citep{yang2025qwen3}, \texttt{OLMo-2}~\citep{walsh2025}, and \texttt{gemma-2}~\citep{team2024gemma}. For each model, we construct Source variants using two prompts: a \emph{Direct} prompt that requests only the final answer and a \emph{CoT} prompt that encourages step-by-step reasoning (e.g., \emph{``Let's think step by step''}, see Appendix~\ref{sec:prompts} for details). 
The \masterkey is extracted from the difference in activations between these two prompts and then transferred to a Target model following the \method\ procedure described in Section~\ref{sec:method}. 
We evaluate performance on three reasoning benchmarks --- GSM8K~\citep{cobbe2021training}, MATH~\citep{hendrycks2021measuring}, and SVAMP~\citep{patel-etal-2021-nlp}.\footnote{We use a maximum generation length of 512 tokens across datasets.}

\subsection{Results \& Discussion}
\NewDocumentCommand{\graybox}{O{2.2em} O{1.5cm} m}{%
  \begingroup
  \setlength{\fboxsep}{0pt}%
  \colorbox{gray!15}{%
    \parbox[c][#1][c]{\dimexpr #2 + 2\tabcolsep\relax}{%
      \centering
      \strut\texttt{#3}%
    }%
  }%
  \endgroup
}

\NewDocumentCommand{\plainbox}{O{2.2em} O{1.4cm} m}{%
  \begingroup
  \setlength{\fboxsep}{0pt}%
    \parbox[c][#1][c]{\dimexpr #2 + 2\tabcolsep\relax}{%
      \centering
      \strut\texttt{#3}%
  }%
  \endgroup
}

\begin{table}
\caption{\textbf{Chain-of-Thought Capability Transfer.}
Transfer performance across model families.
\sethlcolor{gray!15}\hl{Accuracies of base model with Direct prompting ($\TL$) and base model with explicit CoT prompting shown in gray.}}
\label{tab:cot-results}
\centering

\begin{tabular}{
    >{\centering\arraybackslash}m{1.4cm}
    >{\centering\arraybackslash}m{1.6cm}
    cccccc
}
\toprule
\textbf{Model} & \textbf{Prompt} & $\TL$ & $\SU \equiv \SL$ & GSM8K & MATH & SVAMP \\
\midrule

\multirow{7}{*}{\rotatebox[origin=c]{90}{\texttt{Qwen1.5}}}
& \multirow{2}{*}{\graybox{Direct}}
& \cellcolor{gray!15}7B  & \cellcolor{gray!15}--  & \cellcolor{gray!15}9.2  & \cellcolor{gray!15}8.0  & \cellcolor{gray!15}44.0 \\
& 
& \cellcolor{gray!15}14B & \cellcolor{gray!15}--  & \cellcolor{gray!15}16.0 & \cellcolor{gray!15}16.0 & \cellcolor{gray!15}58.3 \\

\cmidrule(lr){2-7}

& \multirow{2}{*}{\graybox{CoT}}
& \cellcolor{gray!15}7B  & \cellcolor{gray!15}--  & \cellcolor{gray!15}64.4  & \cellcolor{gray!15}17.9  & \cellcolor{gray!15}73.0 \\
& 
& \cellcolor{gray!15}14B & \cellcolor{gray!15}--  & \cellcolor{gray!15}77.3 & \cellcolor{gray!15}26.8 & \cellcolor{gray!15}79.0 \\

\cmidrule(lr){2-7}

& \multirow{2}{*}{\plainbox{Direct}}
& 7B  & +\method$_{\text{from 14B}}$  & 56.0  & 20.1  & 70.3 \\
& 
& 14B & +\method$_{\text{from 7B}}$  & 74.4 & 31.2 & 78.3 \\

\midrule

\multirow{5}{*}{\rotatebox[origin=c]{90}{\texttt{OLMo-2}}}
& \multirow{1}{*}{\graybox[1.1em][1.4cm]{Direct}}
& \cellcolor{gray!15}7B  & \cellcolor{gray!15}--  & \cellcolor{gray!15}10.0  & \cellcolor{gray!15}9.7  & \cellcolor{gray!15}43.7 \\

\cmidrule(lr){2-7}

& \multirow{1}{*}{\graybox[1.1em][1.4cm]{CoT}}
& \cellcolor{gray!15}7B  & \cellcolor{gray!15}--  & \cellcolor{gray!15}53.8  & \cellcolor{gray!15}15.3  & \cellcolor{gray!15}71.0 \\

\cmidrule(lr){2-7}

& \multirow{2}{*}{\plainbox{Direct}}
& 7B  & +\method$_{\text{from 1B}}$  & 63.4  & 15.1  & 59.7 \\
& 
& 7B & +\method$_{\text{from 13B}}$  & 36.1 & 14.3 & 58.7 \\

\midrule

\multirow{7}{*}{\rotatebox[origin=c]{90}{\texttt{gemma-2}}}
& \multirow{2}{*}{\graybox{Direct}}
& \cellcolor{gray!15}2B & \cellcolor{gray!15}--  & \cellcolor{gray!15}5.8 & \cellcolor{gray!15}6.2 & \cellcolor{gray!15}36.7 \\
&
& \cellcolor{gray!15}9B  & \cellcolor{gray!15}--  & \cellcolor{gray!15}3.0  & \cellcolor{gray!15}3.5  & \cellcolor{gray!15}21.0 \\ 

\cmidrule(lr){2-7}

& \multirow{2}{*}{\graybox{CoT}}
& \cellcolor{gray!15}2B  & \cellcolor{gray!15}--  & \cellcolor{gray!15}13.3  & \cellcolor{gray!15}8.9  & \cellcolor{gray!15}31.7 \\
& 
& \cellcolor{gray!15}9B & \cellcolor{gray!15}--  & \cellcolor{gray!15}66.6 & \cellcolor{gray!15}26.4 & \cellcolor{gray!15}79.3 \\

\cmidrule(lr){2-7}

& \multirow{2}{*}{\plainbox{Direct}}
& 2B & +\method$_{\text{from 9B}}$  & 9.5 & 6.4 & 37.7 \\
& 
& 9B  & +\method$_{\text{from 2B}}$  & 60.1  & 26.4  & 74.3 \\

\bottomrule
\end{tabular}
\end{table}

We provide our results in Table~\ref{tab:cot-results}, and additional results in Appendix~\ref{sec:cot-appendix}. 
We find that \method (i) consistently improves reasoning performance and displays structured reasoning traces; (ii) is asymmetric in its impact: small-to-large transfer outperforms large-to-small transfer; and (iii) is most effective when the desired capability is already present in latent space.

\paragraph{\bssc{Unlock} Consistently Improves Reasoning Performance:}
Across all evaluated model families and datasets, the \emph{Target Unlocked} model $\TU$ consistently outperforms the baseline $\TL$ under \emph{Direct} prompting.
In the \texttt{Qwen1.5} model family, large-to-small (\transfer{\texttt{Qwen1.5-7B}}{14B}) and small-to-large (\transfer{\texttt{Qwen1.5-14B}}{7B}) produce average accuracy gains of 25.0\% and 31.2\%, respectively.
The performance of $\TU$ is also comparable to the performance obtained from prompting $\TL$ with \emph{explicit CoT} instructions.

\begin{wrapfigure}{r}{0.5\linewidth}
    \centering
    \includegraphics[width=\linewidth]{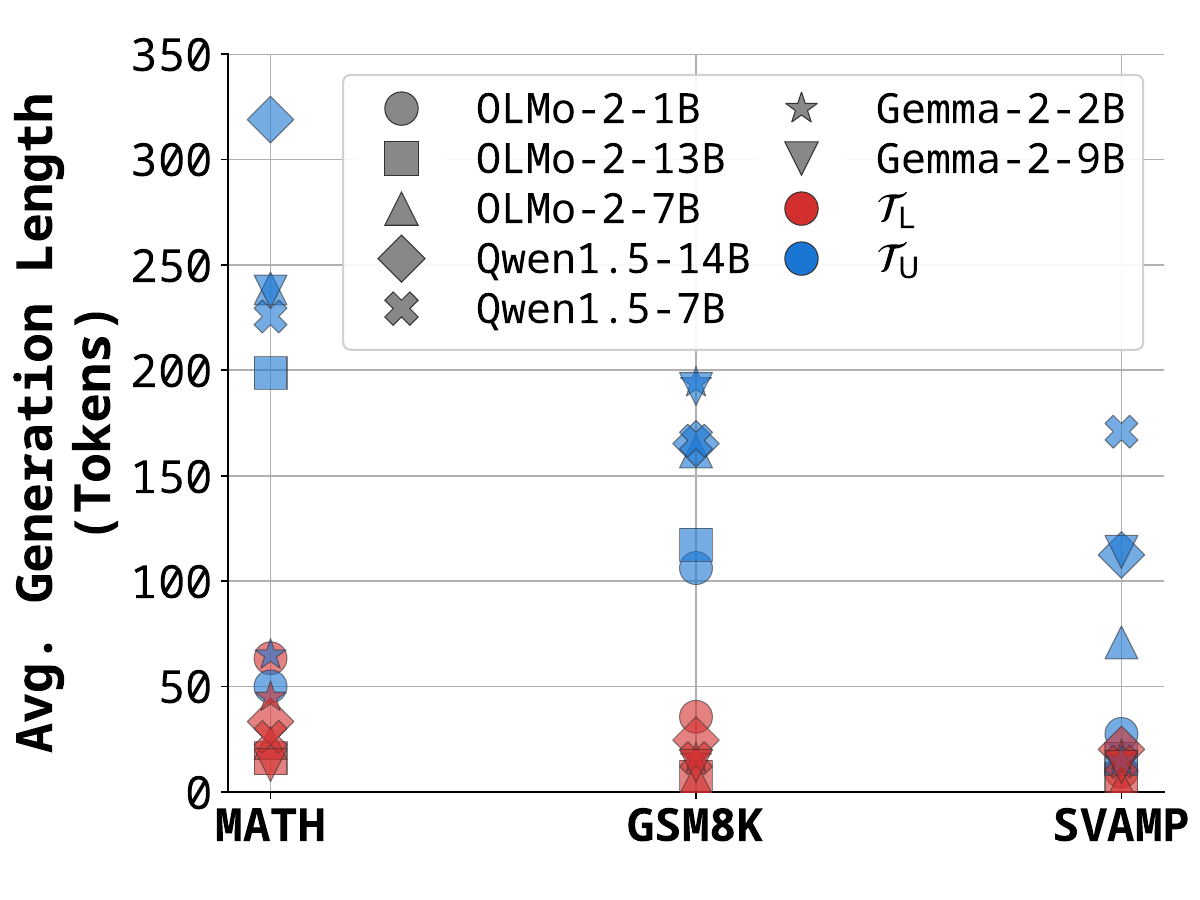}
    \caption{\textbf{Comparison in Generation Length between $\TL$ and $\TU$:} A consistent increase in generation length is observed post-transfer, representative of reasoning behavior.}
    \label{fig:generation-length}
\end{wrapfigure}

Figure~\ref{fig:generation-length} plots the average length of generated outputs for each model and dataset.
A consistent increase in generation length is observed across all model--dataset pairs, supporting the view that the performance gains stem from Chain-of-Thought elicitation rather than surface-level output changes.
We provide further analysis into the structure of the generated outputs and examples of step-by-step reasoning from $\TU$ in Appendix~\ref{sec:cot-appendix}.

\paragraph{Asymmetry in Transfer Direction:}
We observe a consistent directional asymmetry: small-to-large transfer typically produces larger gains than large-to-small transfer.
A plausible explanation is that larger models implement a functional superset of the mechanisms present in smaller models.

Under this view, a CoT direction transferred from a smaller model can activate latent circuitry already present in the larger model. The reverse, however, is capacity-limited: the smaller model's reduced representational capacity may be insufficient to support the more complex reasoning structure of the larger model.
This is illustrated clearly within the \texttt{gemma-2} family.
In the small-to-large direction, $\TU$ improves by an average of $44.4\%$ over $\TL$ and comes within $18.5\%$ of $\TPT$, while large-to-small transfer improves $\TU$ by only $1.6\%$ over $\TL$ and remains $32.3\%$ below $\TPT$.
Importantly, we observe a similar asymmetry when using CoT prompts: \texttt{gemma-2-2B} improves by 2\% while \texttt{gemma-2-9B} improves by 48.2\%.
These results suggest that, like prompting, \method improves with scale and cannot introduce capabilities that are absent from the model.

\paragraph{Transfer Effectiveness Depends On The Salience of The Capability in $\TL$:}
Within the \texttt{Qwen1.5} family, base and instruction-tuned variants exhibit similar performance under CoT prompting, suggesting the reasoning capability is largely introduced during pre-training and can be reliably elicited by prompting. Consequently, $\TU$ significantly outperforms $\TL$ and remains within 1\% $\TPT$.
In contrast, \texttt{gemma-2} models exhibit a substantial gap between their base and instruction-tuned versions with similar prompting, providing evidence that step-by-step reasoning is learned during the post-training process.
Here, \method consistently improves $\TU$ over $\TL$ but does not match $\TPT$, with particularly small gains for \texttt{gemma-2-2B}.
A similar but less pronounced trend is also observable in \texttt{OLMo-2}.
Similar trends across scales are also observed within a model family, as demonstrated by \texttt{Qwen2.5} (Appendix~\ref{sec:cot-appendix}).

These findings reveal that \ohl{\mbox{\method} is most effective when the target capability is already present, though dormant, in the Locked model i.e. when the capability is \emph{atomic}.}

\paragraph{Takeaways:} These results suggest that \method operates analogously to prompting: it can reliably elicit a capability that is \emph{present but dormant} in the model, but cannot introduce one that is absent.
The \masterkey{} thus acts as a mechanism for exposing and activating existing capabilities.
By contrast, when the capability is genuinely absent, \method is unable to induce it.
Introducing a fundamentally missing capability likely requires substantial modification of the model parameters and therefore a corresponding reorganization of the underlying representation space.

\section{Non-Atomic Capability Transfer}
\label{sec:math-main}

Section~\ref{sec:cot-main} established that \method and prompting play analogous roles in eliciting desired model behavior.
We now ask whether this analogy extends to complex \emph{non-atomic} capabilities that only emerge after significant post-training.
Post-training can be thought of as a mapping from a set of input prompts to target behaviors (e.g. placing the final answer within \emph{\textbackslash boxed\{\}}).
Through this process, the model learns to associate inputs and the required capabilities.

Motivated by \cite{huh2024platonic} (which states larger models converge towards a shared representation of the world), and \cite{wang20258020rulehighentropyminority, yue2025doesreinforcementlearningreally} (where the authors find that post-training methods such as RLVR sharpen the output distribution rather than introducing new knowledge), we aim to induce these post-training behaviors with \method.
Intuitively, if post-training merely evokes latent capabilities, and if these capabilities reside in a shared representation space, then transferring them across models becomes a natural next step.
Since these behaviors are not reliably observed in the base model through prompting alone, we ask: \emph{can latent interventions activate non-atomic capabilities that prompting alone cannot?}
We study this question through the lens of \emph{mathematical reasoning}, which is one of the main focuses of modern post-training methods.

Our experiments show that combining prompting with \method not only outperforms prompting alone, but can in some cases surpass post-training.
For instance transferring a mathematical reasoning direction from \texttt{Qwen3-4B} to \texttt{Qwen3-14B} improves the model from 61.1\% to 71.3\% on AGIEval-Math, surpassing the 67.8\% of the 14B instruction-tuned variant.
We further observe that \method sharpens the model's output distribution, concentrating it onto a smaller set of promising early trajectories.

\begin{table}
\normalsize
\caption{\textbf{Math Reasoning Transfer Results:}
Performance of \method across model families. For simplicity, we use $\mathcal{I}(x)$ to denote the instruction-tuned version of the corresponding model.
}
\label{tab:math-results}
\centering

\begin{tabular}{
    >{\centering\arraybackslash}m{1.2cm}
    >{\centering\arraybackslash}m{1.6cm}
    ccccccc
}
\toprule
\textbf{Model} & $\TL$ & $\SU \equiv \mathcal{I}(\SL)$ & \makecell{AGIEval\\Math} & \makecell{Deepmind\\Math} & \makecell{Minerva\\Math} & \makecell{Olympiad\\Bench} \\
\midrule

\multirow{9}{*}{\rotatebox[origin=c]{90}{\texttt{Qwen3}}}
& 4B  & --  & 52.3  & 71.3  & 27.5  & 19.7 \\
& 14B  & --  & 61.1  & 78.8  & 34.7  & 29.0 \\
\cmidrule(lr){2-7}
& $\mathcal{I}$(4B)  & --  & 75.6  & 88.4  & 31.5  & 39.8 \\
& $\mathcal{I}$(14B)  & --  & 67.8  & 80.1  & 27.9  & 37.8 \\
\cmidrule(lr){2-7}
& \cellcolor{cyan!15}4B  & \cellcolor{cyan!15}+\method$_\text{from 14B}$  & \cellcolor{cyan!15}58.9  & \cellcolor{cyan!15}75.8  & \cellcolor{cyan!15}27.0  & \cellcolor{cyan!15}26.4 \\
& \cellcolor{cyan!15}14B  & \cellcolor{cyan!15}+\method$_\text{from 4B}$  & \cellcolor{cyan!15}64.1  & \cellcolor{cyan!15}79.9  & \cellcolor{cyan!15}31.5  & \cellcolor{cyan!15}35.4 \\
\cmidrule(lr){2-7}
& \cellcolor{purple!15}4B  & \cellcolor{purple!15}+\method$_\text{from 14B}$  & \cellcolor{purple!15}49.5  & \cellcolor{purple!15}72.9  & \cellcolor{purple!15}25.7  & \cellcolor{purple!15}20.8 \\
& \cellcolor{purple!15}14B  & \cellcolor{purple!15}+\method$_\text{from 4B}$  & \cellcolor{purple!15}71.3  & \cellcolor{purple!15}82.4  & \cellcolor{purple!15}39.2  & \cellcolor{purple!15}36.3 \\

\midrule

\multirow{9}{*}{\rotatebox[origin=c]{90}{\texttt{Ministral-3}}}
& 3B  & --  & 46.9  & 65.3  & 26.1  & 19.0 \\
& 8B  & --  & 50.7  & 67.4  & 29.3  & 20.0 \\
\cmidrule(lr){2-7}
& $\mathcal{I}$(3B)  & --  & 68.7  & 84.2  & 26.6  & 33.9 \\
& $\mathcal{I}$(8B)  & --  & 70.6  & 87.2  & 29.3  & 37.0 \\
\cmidrule(lr){2-7}
& \cellcolor{cyan!15}3B  & \cellcolor{cyan!15}+\method$_\text{from 8B}$  & \cellcolor{cyan!15}53.4  & \cellcolor{cyan!15}66.2  & \cellcolor{cyan!15}27.5  & \cellcolor{cyan!15}21.0 \\
& \cellcolor{cyan!15}8B  & \cellcolor{cyan!15}+\method$_\text{from 3B}$  & \cellcolor{cyan!15}51.9  & \cellcolor{cyan!15}71.3  & \cellcolor{cyan!15}37.4  & \cellcolor{cyan!15}20.2 \\
\cmidrule(lr){2-7}
& \cellcolor{purple!15}3B  & \cellcolor{purple!15}+\method$_\text{from 8B}$  & \cellcolor{purple!15}49.9  & \cellcolor{purple!15}65.5  & \cellcolor{purple!15}27.5  & \cellcolor{purple!15}21.0 \\
& \cellcolor{purple!15}8B  & \cellcolor{purple!15}+\method$_\text{from 3B}$  & \cellcolor{purple!15}54.0  & \cellcolor{purple!15}70.7  & \cellcolor{purple!15}34.7  & \cellcolor{purple!15}21.1 \\

\bottomrule
\end{tabular}
\end{table}

\subsection{Experiment Setup}
We study two contrasting experimental settings:

\paragraph{{\sethlcolor{cyan!15}\hl{Task-Conditioned Transfer With Limited Data:}}}
The \masterkey, transformation, and hyperparameters are all estimated using \emph{few} examples from the \emph{same} task as evaluation.
This follows the standard practice in the steering vector literature, where the steering direction is computed on the target task to maximize alignment with the evaluation distribution.
Since the evaluation set consists of a limited number of examples, we carry out all pre-computation on a small disjoint development set.

\paragraph{{\sethlcolor{purple!15}\hl{Task-Agnostic Transfer With Abundant Data:}}}
Mirroring conventional post-training practices, the \masterkey and hyperparameters are estimated on a \emph{large} dataset from a \emph{different} math task and applied to all evaluation datasets without modification. 
This setting tests whether the learned intervention captures \emph{general} mathematical reasoning behavior that transfers across tasks.

These two settings expose a central tradeoff between the \emph{in-distribution} signal and the \emph{data volume} required.
In the \sethlcolor{cyan!15}\hl{task-conditioned} regime, we estimate the \masterkey and alignment using limited in-distribution data, which is directly aligned with the evaluation suite, but can yield a noisier and less stable direction/transformation.
In contrast, the \sethlcolor{purple!15}\hl{task-agnostic} regime leverages abundant out-of-distribution examples to learn a more robust \masterkey and mapping, at the cost of estimating them from a distribution-mismatched dataset.
We discuss this tradeoff further in Appendix~\ref{sec:hyperparams-cot}

\paragraph{Models and Datasets:}
We focus on language models with strong reasoning capabilities from four model families: \texttt{Qwen2.5}~\cite{qwen2025qwen25technicalreport}, \texttt{Qwen3}~\cite{qwen3technicalreport}, \texttt{Ministral-3}~\cite{liu2026ministral3}, and \texttt{gemma-3}~\cite{gemmateam2025gemma3technicalreport}.
Within each family, we use the instruction-tuned model as $\SU$ and the corresponding base model as $\SL$.
We evaluate our framework across four mathematical reasoning benchmarks: AGIEval-Math~\citep{zhong2023agievalhumancentricbenchmarkevaluating}, Deepmind-Math~\citep{saxton2019analysingmathematicalreasoningabilities}, Minerva-Math~\citep{lewkowycz2022solvingquantitativereasoningproblems}, and OlympiadBench~\citep{he2024olympiadbenchchallengingbenchmarkpromoting}.
We apply \emph{CoT} prompting to all models, and therefore the information encoded by the \masterkey arises from the additional post-training efforts on $\SU$.
By utilizing different models for $\SL$ and $\SU$, we design a \emph{model-induced} capability transfer setting.
We provide additional experimental details and results with domain-specific models in Appendix~\ref{sec:math-appendix}.

\subsection{Results \& Discussion}
Table~\ref{tab:math-results} reports results for \sethlcolor{cyan!15}\hl{\emph{task-conditioned transfer}} and \sethlcolor{purple!15}\hl{\emph{task-agnostic transfer}}. Our evaluations show consistent gains from \method, and further analysis shows that these gains arise from a convergence in output trajectories, providing evidence that the \masterkey acts as a distribution sharpening mechanism.

\paragraph{Unlocking Matches Gains From Post-Training:}
Consistent with the findings in Section~\ref{sec:cot-main}, the Unlocked model $\TU$ systematically outperforms the baseline $\TL$ and often achieves performance comparable to, or even exceeding, the post-trained counterpart $\TPT$.
For example, \transfer{\texttt{Qwen3-14B}}{4B} and \transfer{\texttt{Qwen3-4B}}{14B} yield average gains of $6.4\%$ and $4.3\%$ over $\TL$ respectively.
Importantly, this shows that while CoT prompting alone is unable to elicit the math reasoning abilities, \method is able to achieve significant gains across models and tasks.
\ohl{Although mathematical reasoning is non-atomic by \mbox{Definition~\ref{def:atomicity}} (as it is not elicited by prompting alone), we find that such capabilities can nonetheless be applied to the Target model as latent test-time interventions, suggesting that the Target model's latent space may be capable of representing them to some degree.}

\paragraph{Assymetry in Task Utilization:}
While both \sethlcolor{cyan!15}\hl{\emph{task-conditioned}} and \sethlcolor{purple!15}\hl{\emph{task-agnostic}} transfer improve $\TU$ over $\TL$, their relative effectiveness depends on the transfer direction.
In the large-to-small setting, we find that \sethlcolor{cyan!15}\hl{\emph{task-conditioned}} transfer is superior, outperforming the \sethlcolor{purple!15}\hl{\emph{task-agnostic}} approach in 69.5\% of the evaluated configurations.
Conversely, for small-to-large transfer, \sethlcolor{purple!15}\hl{\emph{task-agnostic}} transfer yields better results in 70\% of the settings.\footnote{We ignore settings where both methods are within 0.5\% of each other.}

Consistent with our findings from Section~\ref{sec:cot-main}, the most substantial performance gains are observed in the small-to-large transfer scenario.
These trends suggest that when transferring from larger to smaller models, a precise, task-aligned \masterkey is critical for overcoming mismatches in internal circuitry and abilities.
In contrast, because larger models likely contain a functional superset of the circuits and capabilities present in smaller models, small-to-large transfer benefits more from a generalizable \masterkey, and a stronger and more robust transformation.
In this regime, emphasizing general reasoning transfer is more effective than optimizing for task-specific alignment.
We leave further analysis into this mismatch and how capabilities arise with scale to future work.

\begin{figure*}[t]
    \centering
    \begin{subfigure}[t]{0.24\textwidth}
        \centering
        \includegraphics[width=\linewidth]{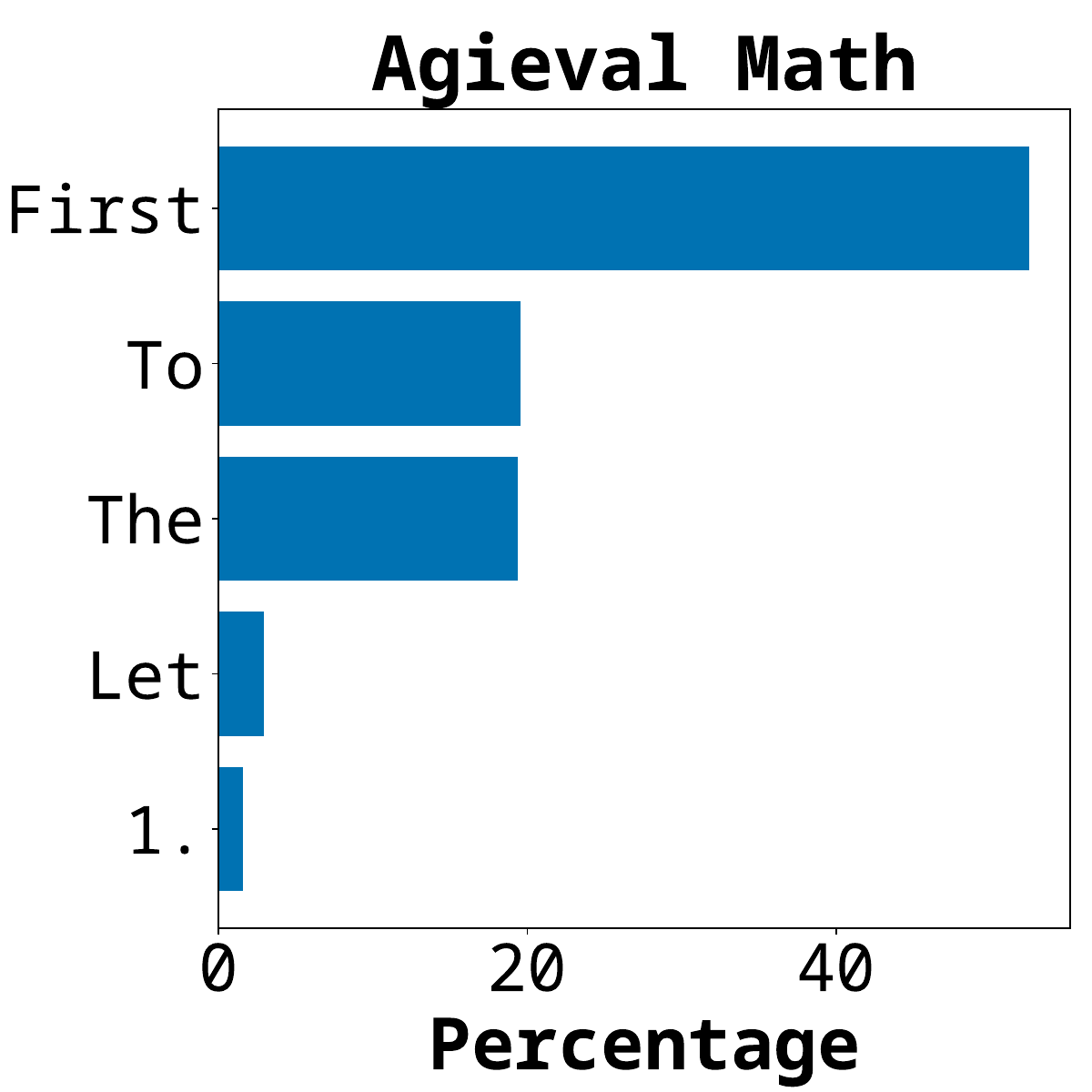}
        \caption{\centering\texttt{Ministral-3-8B}}
    \end{subfigure}
    \hfill
    \begin{subfigure}[t]{0.24\textwidth}
        \centering
        \includegraphics[width=\linewidth]{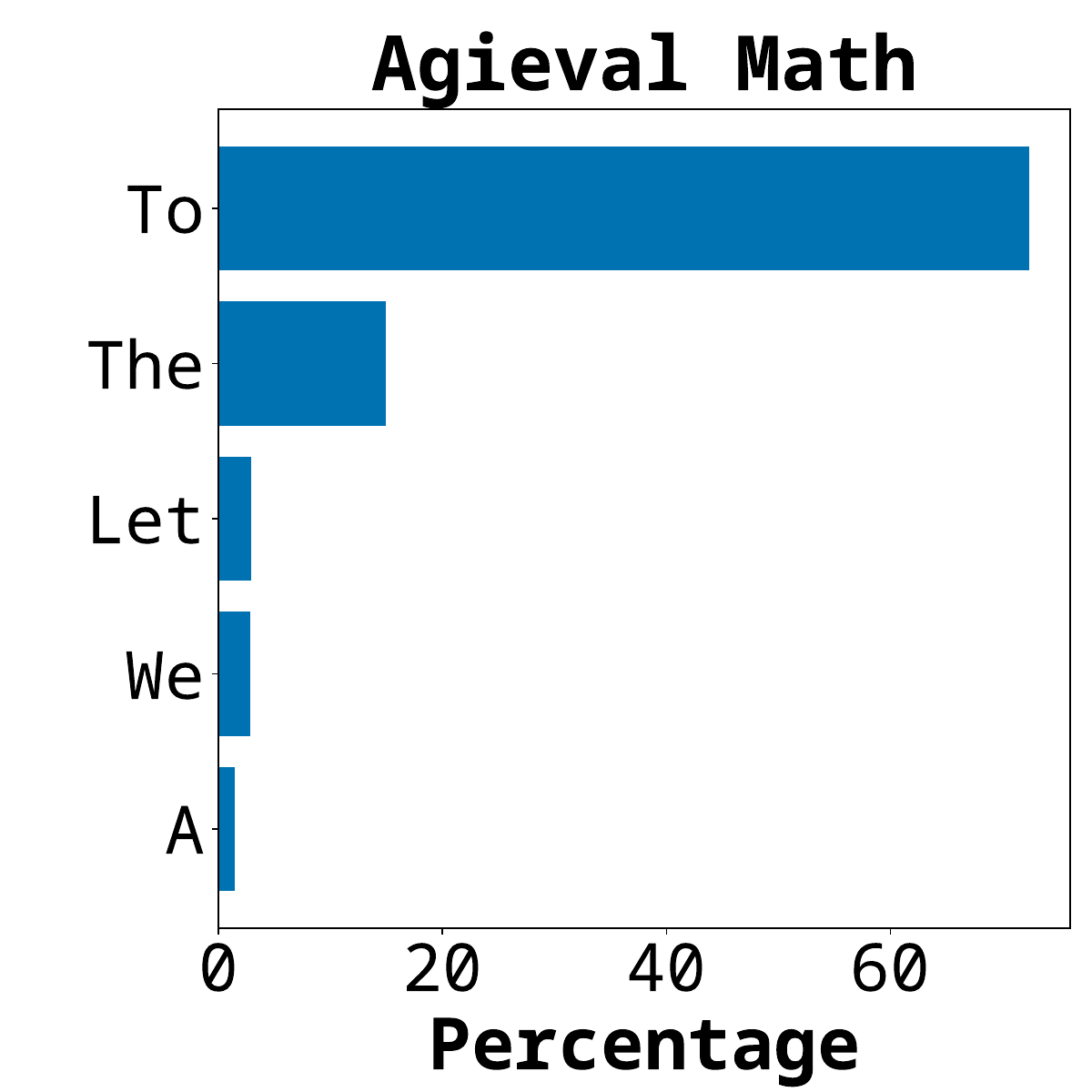}
        \caption{\centering\transfer{\texttt{Ministral-3-8B}}{3B}}
    \end{subfigure}
    \hfill
    \begin{subfigure}[t]{0.24\textwidth}
        \centering
        \includegraphics[width=\linewidth]{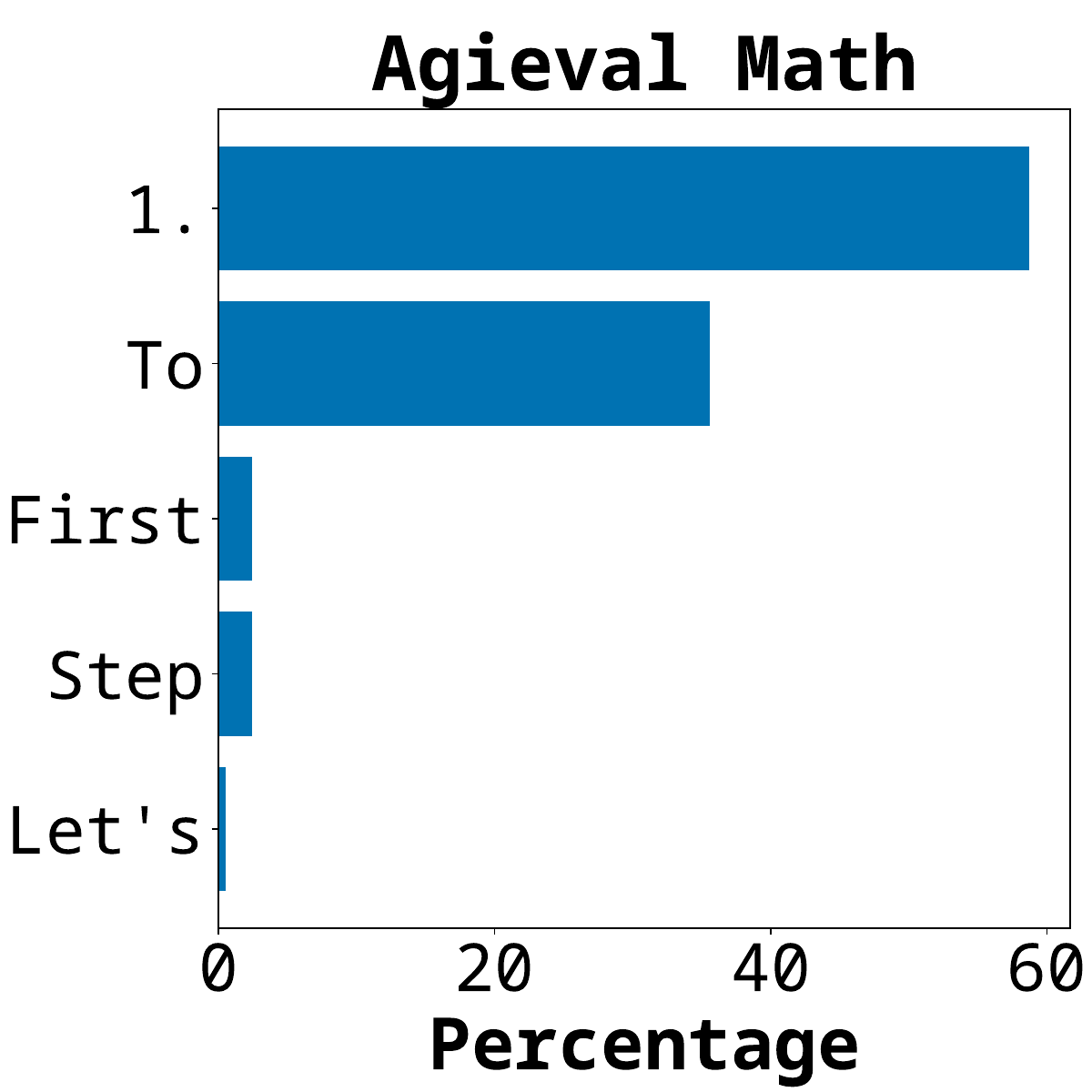}
        \caption{\centering\texttt{Qwen3-14B}}
        \label{fig:3}
    \end{subfigure}
    \hfill
    \begin{subfigure}[t]{0.24\textwidth}
        \centering
        \includegraphics[width=\linewidth]{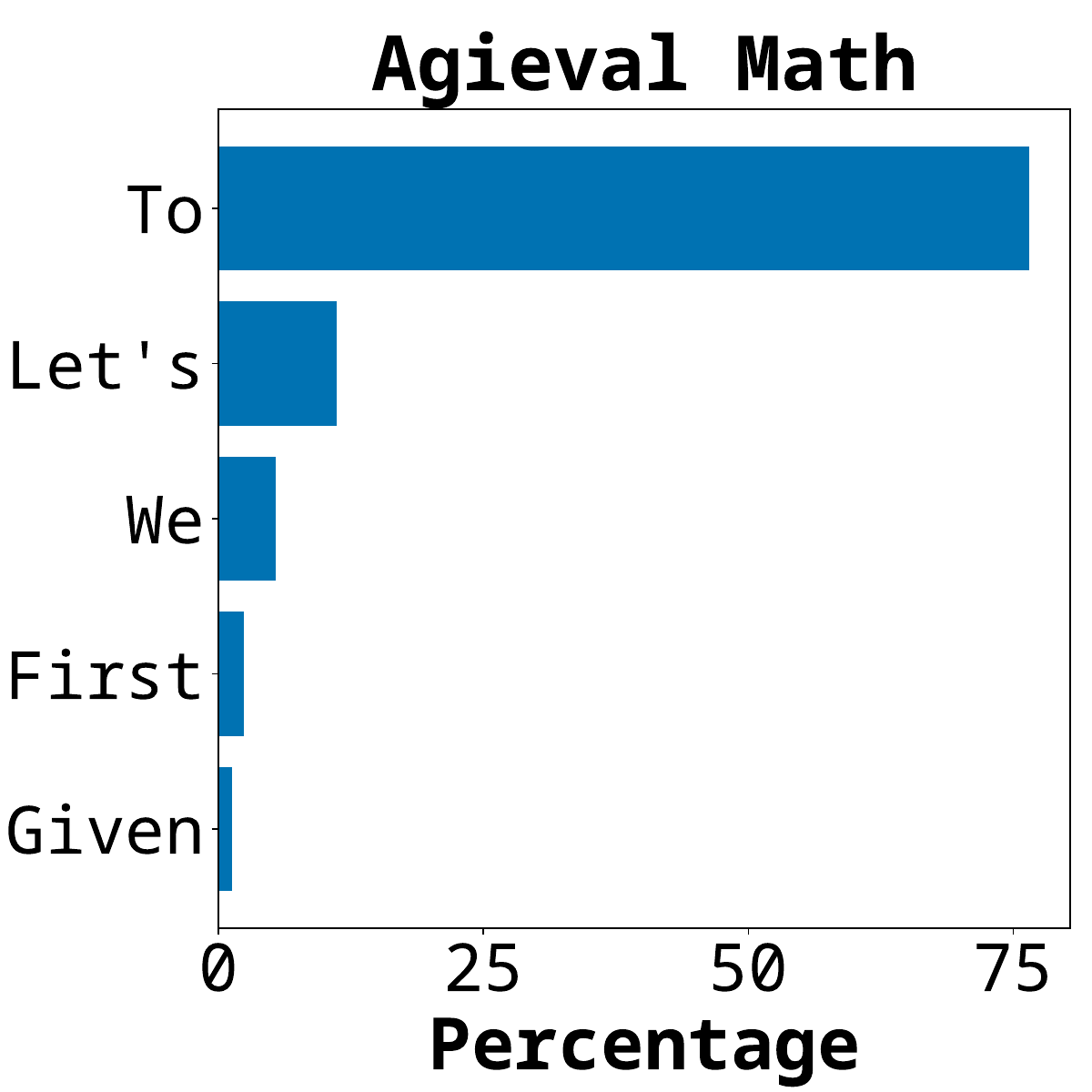}
        \caption{\centering\transfer{\texttt{Qwen3-14B}}{4B}}
    \end{subfigure}
    \caption{\textbf{Statistics of First Generated Word:} The output distribution is significantly skewed to a minimal set of starting traces post-steering.}
    \label{fig:vocab-sharp}
\end{figure*}

\subsection{Convergence of Reasoning Traces:}
To probe for the source of $\TU$'s gains, we analyze the structure of the generated reasoning traces and find that the Unlocked model displays a narrower set of opening trajectories.
We show the distribution of the first generated token of $\TL$ and $\TU$ in Figures~\ref{fig:vocab-sharp},~\ref{fig:first-word-dist-appendix}.
Across models and datasets, we find $\TU$ to converge in it's opening statements, while $\TL$ displays a more diffuse distribution.
We show examples of these changes, along with additional discussions in Appendix~\ref{sec:math-appendix}.
Combined with the improvement in downstream performance these patterns suggest that \method increases the likelihood of producing plausible reasoning traces by consolidating representations and reducing variability in early trajectory selection.
We thus arrive at a similar conclusion as \cite{wang20258020rulehighentropyminority, yue2025doesreinforcementlearningreally} where the authors show that RLVR methods push the model towards narrower responses by editing the probability of a minimal set of tokens. 
We conclude that this \ohl{output reshaping mechanism of post-training can be captured in low dimensional subspaces and applied onto a target model to elicit similar capabilities.}

\paragraph{Takeaways:} Post-training trains the model to map input prompts to desired outputs, a process that relies on eliciting the combination of capabilities required to produce them. 
However, these capabilities are often already present within the model and not introduced during post-training.
Without this learned mapping, prompting alone is insufficient to elicit them.
Instead, \method exploits the presence of the capabilities in latent space. We find that it is possible to isolate and transfer such capabilities as direct latent interventions, without any training.
Put together, these results corroborate our previous findings that \method is most effective when the desired capability is dormant in the model, and the \method primarily improves elicitation of the capability rather than injecting new behaviors or information into the model.
We leave a more thorough analysis of diversity and mode coverage under latent space capability transfer, and its similarity to other post-training methods to future work.

\section{The Master Key Hypothesis \& Implications}
\label{sec:hypothesis-sec}

We now synthesize our empirical findings into a working hypothesis. 
Our results show that: (i) latent interventions extracted from Source contrasts can improve downstream behavior in Target models; (ii) transfer is strongest when the Target model already appears to weakly express the relevant capability; and (iii) a low-rank linear alignment is often sufficient to enable this transfer in practice.
Taken together, these observations motivate the following operational form of the Master Key Hypothesis (MKH).

\clearpage
\begin{hypobox}{The Master Key Hypothesis (MKH)}{master-key-hypothesis}
\label{master-key-hypothesis}
Let $\mathcal{M}^{\vphantom{*}}_1$ be a model with hidden dimension $d^{\vphantom{*}}_1$, and let $l^{\vphantom{*}}_1$ denote some layer of $\mathcal{M}^{\vphantom{*}}_1$.
We hypothesize that there may exist a projection
\(
P^{\vphantom{*}}_1:\mathbb{R}^{d^{\vphantom{*}}_1}\rightarrow\mathbb{R}^{k}, \; k \ll d^{\vphantom{*}}_1,
\)
such that a capability $\psi$ is represented at layer $l_1$ by a direction
\(
\mkey{\mathcal{M}^{\vphantom{*}}_1}{l^{\vphantom{*}}_1} \in \mathbb{R}^{d^{\vphantom{*}}_1},
\)
whose capability-inducing effect is well-captured within the projected subspace $P^{\vphantom{*}}_1(\mathbb{R}^{d^{\vphantom{*}}_1})$.

For a second model $\mathcal{M}^{\vphantom{*}}_2$ with hidden dimension $d^{\vphantom{*}}_2$, and some corresponding layer $l^{\vphantom{*}}_2$, there may likewise exist a projection
\(
P^{\vphantom{*}}_2:\mathbb{R}^{d^{\vphantom{*}}_2}\rightarrow\mathbb{R}^{k},
\)
together with a linear alignment map
\(
f:\mathbb{R}^{k}\rightarrow\mathbb{R}^{k},
\)
such that the capability can approximately be transferred across models as
\[
P^{\vphantom{*}}_2\!\left(\mkey{\mathcal{M}^{\vphantom{*}}_2}{l^{\vphantom{*}}_2}\right)
\approx
f\!\left(P^{\vphantom{*}}_1\!\left(\mkey{\mathcal{M}^{\vphantom{*}}_1}{l^{\vphantom{*}}_1}\right)\right).
\]
where $f$ denotes the low-rank alignment between the corresponding latent subspaces.\\

In this view, the success of transfer depends on the representational capacity of the model.
A capability dormant but present can be isolated and transferred albeit noisily.
When a capability is absent in the Source model, there may be no direction to isolate; when it is absent in the Target model, the Target may lack a compatible representational basis to support the transferred intervention.
\end{hypobox}

Our experiments provide three lines of empirical evidence that are consistent with the MKH.
First, capability directions extracted from Source models reliably transfer to Target models across scales and multiple architectures, which demonstrates that such directions are not model-specific artifacts.
Second, our analysis of the \masterkey (Appendix~\ref{sec:hyperparams-cot}) suggests that the transferable intervention can often be well-approximated in a compact subspace, with effective rank substantially smaller than the hidden size of the model.
Moreover, we observe that these interventions stabilize as the number of examples used to estimate them increases, which is consistent with the view that the transferred signal is structured rather than arbitrary noise.
Third, transfer efficacy varies predictably with capability atomicity: transfer is strongest when the Target model already appears to contain a latent, though weak or dormant, form of the capability, and substantially weaker when that capability is largely absent.

We find that non-atomic capabilities are also transferrable, if they are well represented in the Source contrast, and the Target model possesses sufficient capacity to represent them in latent space.
While we define (non-)atomicity of a capability with respect to its post-training gains and stability across prompts, we find that this does not completely explain our results.
For example, a simple capability such as Chain-of-Thought is difficult to transfer perfectly in the \texttt{gemma-2} family, while complex math reasoning abilities can be transferred in the \texttt{Qwen-3} family.
Further, the fact that non-atomic capabilities are transferrable hints at the possibility that they could be represented as a combination of simpler abilities.
As such, we believe that Definitions~\ref{def:capability},~\ref{def:atomicity} are functionally incomplete --- the atomicity of a capability should be defined based on how well it can be isolated in latent space, and not by its stability or elicitation in input/output token space.
We believe this to be outside the line of this work and leave it to future research.

The Master Key Hypothesis builds on two lines of prior work.
The Linear Representation Hypothesis (LRH)~\cite{mikolov2013efficientestimationwordrepresentations,park2024linearrepresentationhypothesisgeometry} suggests that concepts can correspond to consistent directions in representation space \emph{within} a model.
The Platonic Representation Hypothesis (PRH)~\cite{huh2024platonic} suggests that latent representations may converge \emph{across} models.
The MKH unifies these findings at the level of capabilities, arguing that post-training behaviors can often be modeled as \emph{transferable} latent interventions \emph{across} model scales.
Our results are consistent with extending these ideas from concepts to behaviors: not only semantic features, but also some capability-inducing interventions, may admit compact and partially transferable latent structure across related models.
We emphasize, however, that our results provide empirical support for this view rather than a mechanistic proof of it.

The MKH also offers one possible interpretation of recent findings of~\cite{yue2025doesreinforcementlearningreally,wang20258020rulehighentropyminority,li2026rlvrtrainingllmsdoes}, which suggest that reinforcement-style post-training often sharpens or re-weights existing output trajectories rather than introducing entirely new knowledge.
In our setting, we find that the behavior associated with post-training can sometimes be partially reproduced by transferring a latent intervention (\masterkey).
This is consistent with the view that certain post-training effects such as mathematical reasoning operate by amplifying pre-existing latent tendencies rather than introducing new representational structure.
At the same time, our results also suggest clear limits: such transfer is much less effective for older or weaker models that appear to not possess the necessary representational basis for the desired behavior.

While our findings support the usefulness of the MKH as an empirical abstraction, they do not yet determine the precise mechanism by which capabilities are formed, represented, or interact with each other.
The MKH posits the existence of shared low-dimensional subspaces without specifying how they arise from pre-training dynamics or architectural constraints.
We therefore view MKH as a useful operational hypothesis that organizes the empirical patterns observed in this work and generates concrete predictions for future study.
Just as the Linear Representation Hypothesis motivated subsequent mechanistic work on how concepts are encoded, MKH motivates analogous investigation into how capabilities are learned, organized, and combined in representation space. We leave this to future work.

\section{Related Work}
\paragraph{Steering vectors:}
Steering vectors modulate model behavior by intervening on internal activations \citep{turner2024steeringlanguagemodelsactivation}, with early work emphasizing safety-relevant behaviors \citep{panickssery2024steeringllama2contrastive}.
A broad literature argues that many attributes are captured by low-dimensional directions \citep{gurnee2024languagemodelsrepresentspace,arditi2024refusallanguagemodelsmediated,konen2024stylevectorssteeringgenerative,vanderweij2024extendingactivationsteeringbroad, zou2025representationengineeringtopdownapproach}.
Steering has also been used to improve reasoning and downstream performance and to support mechanistic analysis \citep{liu2024incontextvectorsmakingcontext, tan2025analyzinggeneralizationreliabilitysteering, todd2024functionvectorslargelanguage, stoehr2024activationscalingsteeringinterpreting, ghandeharioun2024patchscopesunifyingframeworkinspecting, chen2024selfieselfinterpretationlargelanguage, hong2025reasoningmemorizationinterplaylanguagemodels, venhoff2025understandingreasoningthinkinglanguage, stolfo2025improvinginstructionfollowinglanguagemodels, venhoff2025basemodelsknowreason, zhong2025watchweightsunsupervisedmonitoring}.
Recently, there has been growing interest in distilling capabilities in language modes using steering vectors.
\cite{azizi2025activationsteeringchainofthoughtcompression} show that concise Chain-of-Though abilities can be isolated as a single vector within a language model.
Parallel to our work, \cite{COTeditLessisWrong} show that jail-breaking in language models can be simply performed by substituting or sampling for targeted words, to fool the model into generating coherent reasoning traces for unsafe questions.

\paragraph{Distinction from prior steering transfer work:}
Most cross-model steering transfer is demonstrated on safety, jailbreak, or style behaviors, where evaluation often relies on coarse proxies (e.g., refusal-string presence), the steering vectors are constructed from explicit positive/negative supervision, and applied to the same model/task. 
In contrast, we study \emph{capability transfer} across \emph{model sizes and families} and evaluate success using \emph{task-level correctness} on standard reasoning benchmarks.
We provide a unified formalization of (i) targeted shifts derived from \emph{prompt- or model-induced} representational differences and (ii) the cross-model alignment required to apply such shifts in a new model.

\paragraph{Capability transfer across models:}
Prior approaches define the transfer signal in (i) \textbf{weight space}, (ii) \textbf{output/probability space}, or (iii) \textbf{distillation-based training}.
Weight-space methods reuse parameter deltas as task directions \citep{ilharco2023editingmodelstaskarithmetic, huang2024chatvectorsimpleapproach, buzzega2025rethinkinglayerwisemodelmerging, wu2025shadowfttuninginstructmodel, wan2024fusechatknowledgefusionchat, zbeeb2025reasoningvectorstransferringchainofthought}, but typically do not carry across sizes or families.
Logit-space methods guide a student using stronger-model outputs \citep{ouyang2025rastreasoningactivationllms, fei2025nudginginferencetimealignmentllms, liu2024tuninglanguagemodelsproxy}, but require multi-model computation at inference.

\paragraph{Representational convergence and cross-model alignment:}
A growing line of work argues that different models learn compatible representations, enabling transfer through shared subspaces or simple maps \citep{kaushik2025universalweightsubspacehypothesis, berman2025saidneuralnetworksarent, huang2025crossmodeltransferabilitylargelanguage}.
We also acknowledge concurrent efforts that learn mappings across model sizes \citep{oozeer2025activationspaceinterventionstransferred, bello2025linearrepresentationtransferabilityhypothesis}.
Unlike prior work, we use a \emph{low-rank linear alignment} rather than non-linear autoencoders or full-dimensional psuedoinverse matrices, and we focus on improvements on quantifiable improvements on downstream tasks.

\paragraph{Knowledge distillation:}
Finally, classical distillation transfers capabilities by \emph{training} a student model to match a teacher distribution \citep{gu2025minillmknowledgedistillationlarge,wang2025lightreasonersmalllanguagemodels,gu2025minillmknowledgedistillationlarge,shen2025codicompressingchainofthoughtcontinuous,caccia2025trainingplugnplayknowledgemodules,qin2022knowledgeinheritancepretrainedlanguage}.
Unlike our setting, distillation typically incurs a nontrivial training cost and must be repeated per student model.
Concurrently with our work, others have explored self-distillation in language models \citep{shenfeld2026selfdistillationenablescontinuallearning,hubotter2026reinforcementlearningselfdistillation}, and claim that contextual knowledge and capabilities can be distilled into a model simply by training on it's logits along with additional feedback or examples.
While we take a training-free approach, these works provide further grounding and motivation by empirically proving that target abilities can be elicited simply by incorporating additional task-conditioned signals.

\section{Conclusion}
In this paper, we present a training-free approach for \emph{transferring capabilities} across models.
Our method extracts a \masterkey direction from prompt- /model-induced representational differences and transfers it to a new model via low-rank linear subspace alignment, avoiding gradient updates and requiring no architectural or tokenization correspondence between Source and Target pairs.
Empirical evaluations across multiple model families and benchmarks confirm the effectiveness of our approach. 
More broadly, our results support the \emph{Master Key Hypothesis}, suggesting that useful behaviors can be isolated as linearly transferrable latent directions in shared low-dimensional subspaces.

\section{Acknowledgments}
We thank Quyet Do, Thinh Phan, Nguyen Nguyen, Weiyuan Chen, Jing Chen, Yu-Min Tseng, Noah Provenzano, and Yeana Bond for valuable discussions and feedback.
Rishab, Pin-Jie, and Tu were supported by an award from the Amazon - Virginia Tech Initiative for Efficient and Robust Machine Learning.
We acknowledge Advanced Research Computing at Virginia Tech for providing computational resources and support.\footnote{https://arc.vt.edu/}

\newpage

\bibliographystyle{abbrvnat}
\bibliography{main}

\clearpage
\newpage
\appendix

\section{Additional Preliminaries}

\subsection{Comparison to Previous Approaches}
\label{sec:comp-prev-work}

We provide a comparison of our approach to prior work in Table~\ref{tab:method-comparison}.
We are amongst the first to demonstrate that high-level capability transfer is inherently low-rank.
Building on this insight, we perform the extensive evaluation of both large-to-small and small-to-large capability transfer using latent steering vectors.
Crucially, our approach is entirely \emph{training-free} and requires \emph{no labeled data}, distinguishing it from existing methods that rely on gradient updates or supervised signals.

\begin{table*}[!ht]
\caption{\textbf{Comparison of our method to other approaches:} \method is completely training-free and label-free, and shows improvements across model scales and architectures for extrinsic evaluation tasks.}
	\begin{center}
        \begin{small}
        \begin{sc}
            \begin{tabular}{cccccc}
                \toprule
                \textbf{Method} & \textbf{\makecell{Transfer\\Space}} & \textbf{\makecell{No Labeled\\Data}} & \textbf{\makecell{Fixed\\Compute}} & \textbf{\makecell{Transferrable\\Across Sizes}} & \textbf{\makecell{Extrinsic\\Evaluations}} \\

                \midrule
                \makecell{Task Vectors\\ \citep{ilharco2023editingmodelstaskarithmetic}} & Weight & \cmark & \cmark & \xmark & \cmark \\[7pt]

                \makecell{Knowledge Distillation\\ \citep{gu2025minillmknowledgedistillationlarge}} & Weight & \xmark & \cmark & \cmark & \cmark \\[7pt]

                \makecell{Proxy Tuning\\ \citep{liu2024tuninglanguagemodelsproxy}} & Logit & \cmark & \xmark & \cmark & \cmark \\[7pt]

                \makecell{Steering Vectors\\ \citep{panickssery2024steeringllama2contrastive}} & Latent & \xmark & \cmark & \xmark & \xmark \\[7pt]

                \makecell{Patchscopes\\ \citep{ghandeharioun2024patchscopesunifyingframeworkinspecting}} & Latent & \xmark & \cmark & \cmark & \xmark \\[7pt]

                \makecell{Activation Intervention\\ \citep{oozeer2025activationspaceinterventionstransferred}} & Latent & \xmark & \cmark & \cmark & \xmark \\[7pt]

                \midrule
                \method & Latent & \cmark & \cmark & \cmark & \cmark \\
 
                \bottomrule
            \end{tabular}
        \end{sc}
        \end{small}
	\end{center}
	\label{tab:method-comparison}
\end{table*}

\subsection{Prompts and Models Used}
\label{sec:prompts}

\newcommand{\atok}{\texttt{\textless atok\textgreater}}
\newcommand{\catok}{\texttt{\textless/atok\textgreater}}

\newtcolorbox{promptbox}[1]{%
	enhanced,
	colback=#1!18,
	colframe=black!35,
	boxrule=0.7pt,
	arc=10pt,
	left=10pt,
	right=10pt,
	top=8pt,
	bottom=8pt,
	width=\linewidth,
	boxsep=3pt
}

\begin{figure}[h!]
	\centering
	\begin{minipage}{\columnwidth}
		\small
		\setlength{\parskip}{4pt}
		\setlength{\parindent}{0pt}

		\begin{promptbox}{green}
			\centering
            \emph{Reason step by step and give a final answer to the following question.}
            Your response should always end with ``The final answer is \atok\ [answer]\ \catok.''
            where [answer] is the correct solution to the problem.
            
            \textbf{Question:}\\
            Natalia sold clips to 48 of her friends in April, and then she sold half as many clips in May.
            How many clips did Natalia sell altogether in April and May?
            
            \textbf{Answer:} \emph{Let's think step by step.}
            \normalfont
            
		\end{promptbox}

		\vspace{8pt}

		\begin{promptbox}{red}
			\centering
			Solve the following question and place the answer at the end.
			Your response should always end with ``The final answer is \atok\ [answer]\ \catok.''
			where [answer] is the correct solution to the problem.
            
			\textbf{Question:}\\
			Natalia sold clips to 48 of her friends in April, and then she sold half as many clips in May.
			How many clips did Natalia sell altogether in April and May?

			\textbf{Answer:}
            \normalfont
		\end{promptbox}
	\end{minipage}
	\caption{\textbf{Example Prompts:} Example \emph{CoT} (green) and \emph{Direct} (red) prompts which are used for all evaluations.}
	\label{fig:prompt}
\end{figure}

We show the \emph{Direct} and \emph{CoT} prompts that we used in Figure~\ref{fig:prompt}.
To avoid discrepancies in prompt templates across models, we use only the two prompt types shown for all experiments.
We observe that some post-trained models tend to format outputs to end with "\emph{\textbackslash boxed\{\textnormal{ans}}\}".
To prevent results from being skewed in favor of such models, we instead use a unified concluding token pattern, "\emph{<atok>\textnormal{ans}</atok>}", for the final answer.

\section{Additional Results for Unlocking Chain of Thought}
\label{sec:cot-appendix}

We provide comparisons of the Unlocked model $\TU$ to the post-trained model $\TPT$, along with additional experiments for \texttt{Qwen2.5} and \texttt{Qwen3} model families in Table~\ref{tab:cot-results-appendix}

\begin{figure}[!t]
	\centering
    \begin{subfigure}{0.48\linewidth}
		\centering
		\includegraphics[width=\linewidth]{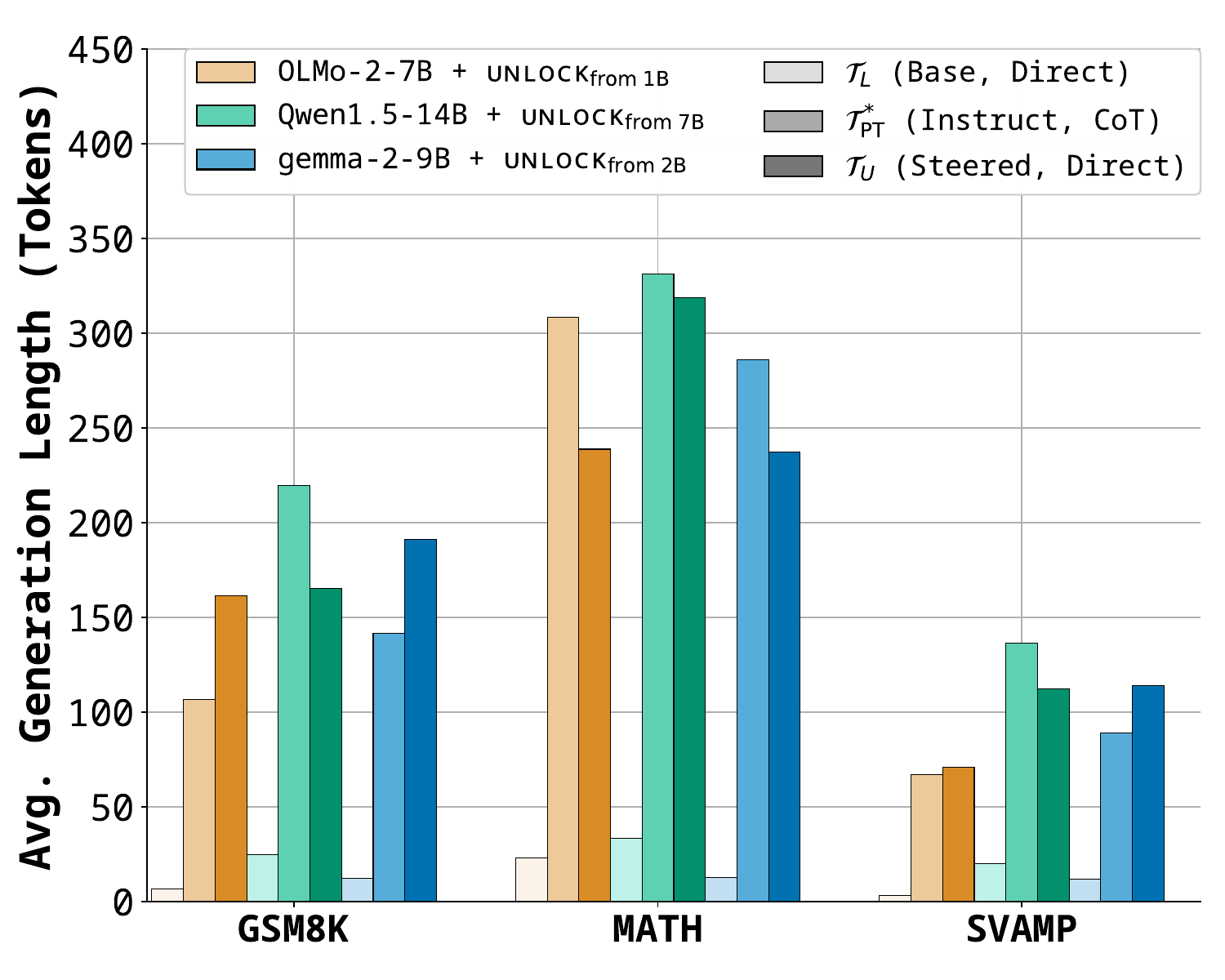}
		\caption{Small-to-Large}
		\label{fig:gen-len-small2large}
	\end{subfigure}
    \hfill
	\begin{subfigure}{0.48\linewidth}
		\centering
		\includegraphics[width=\linewidth]{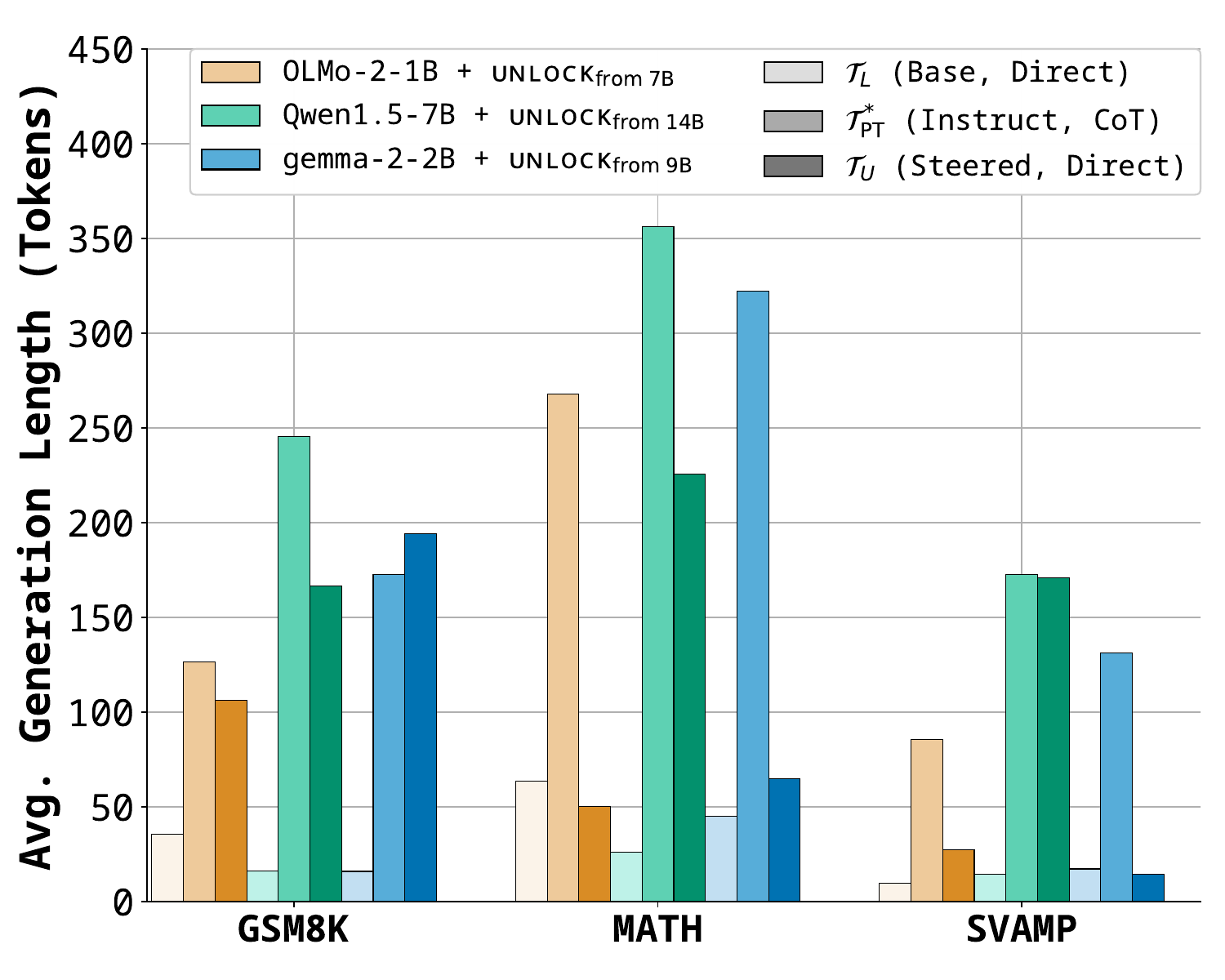}
		\caption{Large-to-Small}
		\label{fig:gen-len-large2small}
	\end{subfigure}	

	\caption{\textbf{Increased Generation Length of $\TU$:} \method leads to a clear increase in generation length over the base model with Direct prompting, matching the length of the instruction-tuned model with explicit CoT prompts.}
	\label{fig:gen-len}
\end{figure}

\subsection{Impact of Unlocking}
\label{sec:app-impact-unlock-cot}

\paragraph{Increased Generation Lengths and Task Performance:}
We find a significant increase in the length of generated answers across all evaluated models in Figure~\ref{fig:gen-len}.
While increased generation length is consistent with step-by-step reasoning, it may also be a result of unhelpful verbosity, such as repetition or hallucination.
To assess whether the additional text is task-relevant, we analyze correctness as a function of generation length.

\begin{figure*}[!hp]
	\centering

	\begin{subfigure}{\linewidth}
		\centering
		\includegraphics[width=\linewidth,height=0.45\textheight]{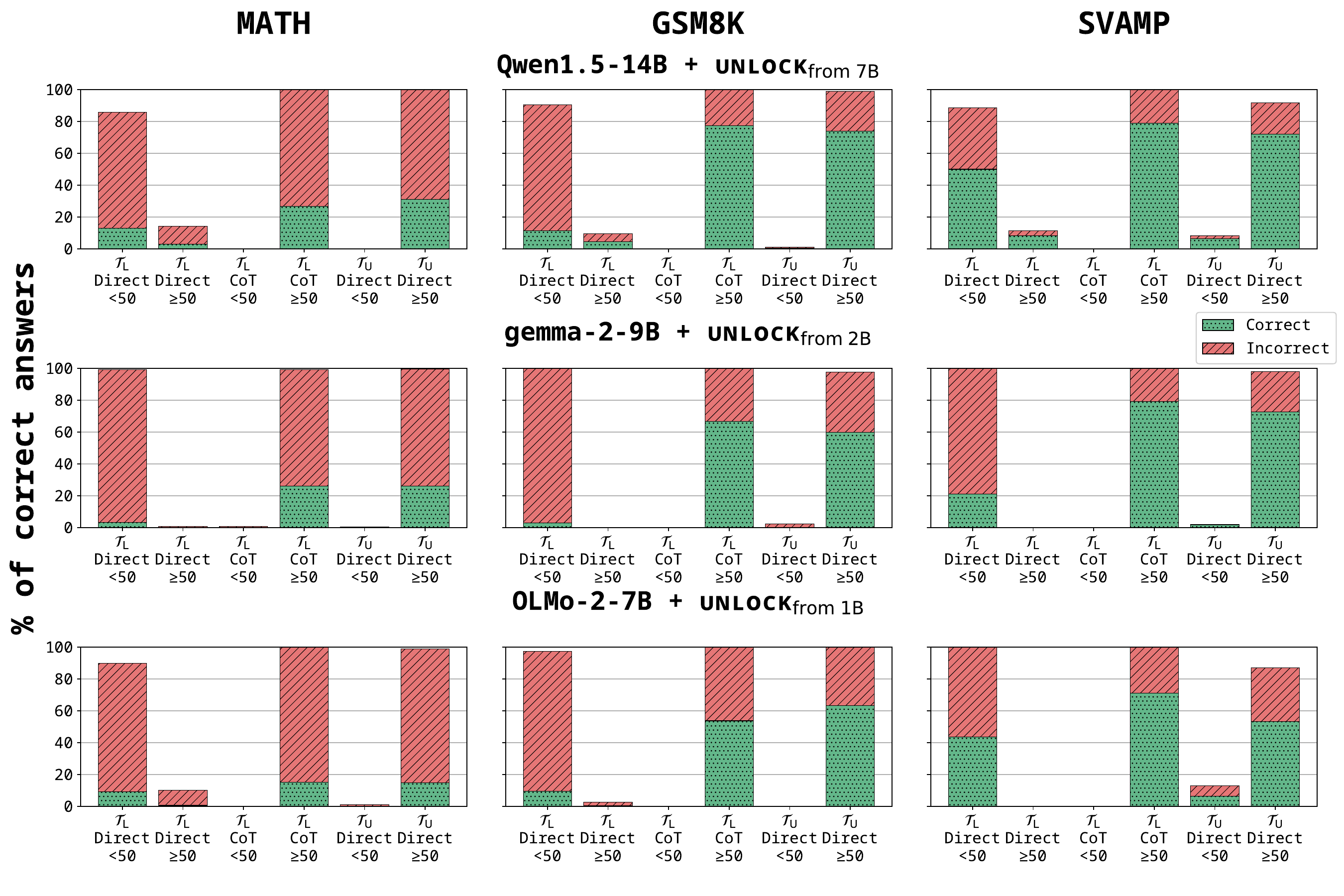}
		\caption{small-to-large}
	\end{subfigure}
	\hfill
	\begin{subfigure}{\linewidth}
		\centering
		\includegraphics[width=\linewidth,height=0.45\textheight]{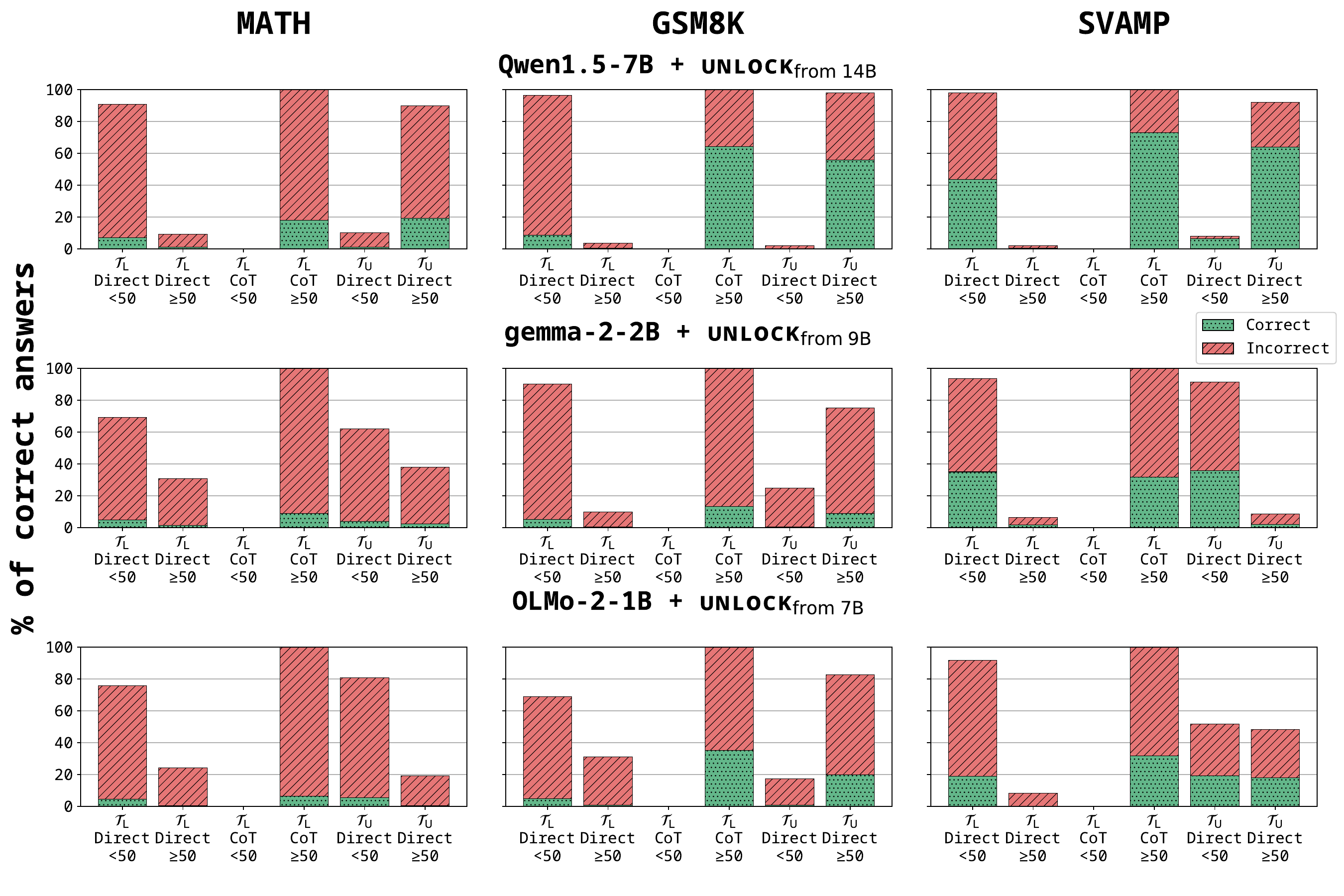}
		\caption{large-to-small}
	\end{subfigure}

	\caption{
        \textbf{Evidence for Improved Reasoning:} Generation length of the Unlocked model significantly increases over the Locked model, with a corresponding improvement in downstream performance.
    }
	\label{fig:gen-bins}
\end{figure*}

Specifically, we bin outputs based on the number of generated characters for the base model with Direct prompt (i.e. Locked model $\TL$), the base model with CoT prompt, and the Unlocked model $\TU$ with Direct prompt.
We choose a binning threshold of 50 characters, which corresponds to the length of our response template.
Figure~\ref{fig:gen-bins} displays the percentage of correct solutions in each bin.

We consistently observe an increase in generation under two conditions: (1) as we transition from direct to CoT prompting; and (2) when we move from $\TL$ to $\TU$ with Direct prompting.
Crucially, this increase in length is accompanied by a higher proportion of correct solutions.
This correlation indicates that \method does not merely append extraneous text but instead elicits meaningful intermediate content that improves downstream performance.
We provide qualitative examples illustrating these behavioral shifts in Examples~\ref{ex:cot-success-1}--\ref{ex:cot-success-4}.

\paragraph{\method is Non-Destructive \& Compliments Gains From Parameter Scaling:}
In the \texttt{Qwen2.5} family, effective CoT usage is present at 1.5B size but the model typically requires explicit CoT prompting to produce intermediate steps (supported by the significant gains from CoT prompts in Table~\ref{tab:cot-results-appendix}).
In contrast, the 7B model often produces intermediate steps even under direct prompting.
These findings are in line with \cite{wei2023chainofthoughtpromptingelicitsreasoning}, who show that Chain-of-Thought reasoning emerges with scale.

Across both sizes, we observe that $\TU$ displays strong reasoning capabilities and performs competitively with ($\pm 2\%$ of) $\TPT$.
We observe a similar trend in \texttt{Qwen-3}, supporting the view that \method is \emph{non-destructive}: it does not inhibit performance in models where the behavior is reliably displayed, while it reliably elicits the capability when it is present but unused.

\begin{table}
\normalsize
\caption{\textbf{Chain-of-thought Capability Transfer Results.}
Performance of \method across model families.
\sethlcolor{gray!15}\hl{Accuracies of the base model with Direct prompting (Locked model), base model with CoT prompting, and instruction-tuned model with CoT prompting shown in gray}.}
\label{tab:cot-results-appendix}
\centering

\begin{tabular}{
    >{\centering\arraybackslash}m{1.4cm}
    >{\centering\arraybackslash}m{1.6cm}
    cccccc
}
\toprule
\textbf{Model} & \textbf{Prompt} & $\TL$ & $\SU \equiv \SL$ & GSM8K & MATH & SVAMP \\
\midrule

\multirow{9}{*}{\rotatebox[origin=c]{90}{\texttt{Qwen1.5}}}
& \multirow{2}{*}{\graybox{Direct}}
& \cellcolor{gray!15}{7B}  & \cellcolor{gray!15}--  & \cellcolor{gray!15}9.2  & \cellcolor{gray!15}8.0  & \cellcolor{gray!15}44.0 \\
& 
& \cellcolor{gray!15}{14B} & \cellcolor{gray!15}--  & \cellcolor{gray!15}16.0 & \cellcolor{gray!15}16.0 & \cellcolor{gray!15}58.3 \\

\cmidrule(lr){2-7}

& \multirow{2}{*}{\graybox{CoT}}
& \cellcolor{gray!15}{7B}  & \cellcolor{gray!15}--  & \cellcolor{gray!15}64.4  & \cellcolor{gray!15}17.9  & \cellcolor{gray!15}73.0 \\
& 
& \cellcolor{gray!15}{14B} & \cellcolor{gray!15}--  & \cellcolor{gray!15}77.3 & \cellcolor{gray!15}26.8 & \cellcolor{gray!15}79.0 \\

\cmidrule(lr){2-7}

& \multirow{2}{*}{\graybox{CoT}}
& \cellcolor{gray!15}{7B-Chat}  & \cellcolor{gray!15}--  & \cellcolor{gray!15}58.1  & \cellcolor{gray!15}18.2  & \cellcolor{gray!15}69.7 \\
& 
& \cellcolor{gray!15}{14B-Chat} & \cellcolor{gray!15}--  & \cellcolor{gray!15}74.8 & \cellcolor{gray!15}30.2 & \cellcolor{gray!15}82.0 \\

\cmidrule(lr){2-7}

& \multirow{2}{*}{\plainbox{Direct}}
& 7B  & +\method$_{\text{from {14B}}}$  & 56.0  & 20.1  & 70.3 \\
& 
& 14B & +\method$_{\text{from {7B}}}$  & 74.4 & 31.2 & 78.3 \\

\midrule

\multirow{14}{*}{\rotatebox[origin=c]{90}{\texttt{OLMo-2}}}
& \multirow{3}{*}{\graybox[3.3em][1.4cm]{Direct}}
& \cellcolor{gray!15}{1B} & \cellcolor{gray!15}--  & \cellcolor{gray!15}5.5 & \cellcolor{gray!15}4.9 & \cellcolor{gray!15}19.0 \\
& 
& \cellcolor{gray!15}{7B}  & \cellcolor{gray!15}--  & \cellcolor{gray!15}10.0  & \cellcolor{gray!15}9.7  & \cellcolor{gray!15}43.7 \\
& 
& \cellcolor{gray!15}{13B}  & \cellcolor{gray!15}--  & \cellcolor{gray!15}18.7  & \cellcolor{gray!15}13.8  & \cellcolor{gray!15}65.7 \\

\cmidrule(lr){2-7}

& \multirow{1}{*}{\graybox[3.3em][1.4cm]{CoT}}
& \cellcolor{gray!15}{1B}  & \cellcolor{gray!15}--  & \cellcolor{gray!15}35.0  & \cellcolor{gray!15}6.4  & \cellcolor{gray!15}31.7 \\
&
& \cellcolor{gray!15}{7B}  & \cellcolor{gray!15}--  & \cellcolor{gray!15}53.8  & \cellcolor{gray!15}15.3  & \cellcolor{gray!15}71.0 \\
&
& \cellcolor{gray!15}{13B}  & \cellcolor{gray!15}--  & \cellcolor{gray!15}67.1  & \cellcolor{gray!15}20.2  & \cellcolor{gray!15}75.7 \\

\cmidrule(lr){2-7}

& \multirow{1}{*}{\graybox[3.3em][1.4cm]{CoT}}
& \cellcolor{gray!15}{1B-Instruct}  & \cellcolor{gray!15}--  & \cellcolor{gray!15}63.7 & \cellcolor{gray!15}16.1 & \cellcolor{gray!15}64.0 \\
&
& \cellcolor{gray!15}{7B-Instruct}  & \cellcolor{gray!15}--  & \cellcolor{gray!15}79.4  & \cellcolor{gray!15}24.9  & \cellcolor{gray!15}78.7 \\
&
& \cellcolor{gray!15}{13B-Instruct}  & \cellcolor{gray!15}--  & \cellcolor{gray!15}80.6  & \cellcolor{gray!15}33.8  & \cellcolor{gray!15}75.0 \\

\cmidrule(lr){2-7}

& \multirow{4}{*}{\plainbox{Direct}}
& {1B}  & +\method$_{\text{from {7B}}}$  & 20.5  & 5.9  & 37.3 \\
&
& {7B}  & +\method$_{\text{from {1B}}}$  & 63.4  & 15.1  & 59.7 \\
& 
& {7B} & +\method$_{\text{from {13B}}}$  & 36.1 & 14.3 & 58.7 \\
& 
& {13B} & +\method$_{\text{from {7B}}}$  & 45.8 & 16.0 & 67.3 \\

\midrule

\multirow{9}{*}{\rotatebox[origin=c]{90}{\texttt{gemma-2}}}
& \multirow{2}{*}{\graybox{Direct}}
& \cellcolor{gray!15}{2B} & \cellcolor{gray!15}--  & \cellcolor{gray!15}5.8 & \cellcolor{gray!15}6.2 & \cellcolor{gray!15}36.7 \\
&
& \cellcolor{gray!15}{9B}  & \cellcolor{gray!15}--  & \cellcolor{gray!15}3.0  & \cellcolor{gray!15}3.5  & \cellcolor{gray!15}21.0 \\ 

\cmidrule(lr){2-7}

& \multirow{2}{*}{\graybox{CoT}}
& \cellcolor{gray!15}{2B}  & \cellcolor{gray!15}--  & \cellcolor{gray!15}13.3  & \cellcolor{gray!15}8.9  & \cellcolor{gray!15}31.7 \\
& 
& \cellcolor{gray!15}{9B} & \cellcolor{gray!15}--  & \cellcolor{gray!15}66.6 & \cellcolor{gray!15}26.4 & \cellcolor{gray!15}79.3 \\

\cmidrule(lr){2-7}

& \multirow{2}{*}{\graybox{CoT}}
& \cellcolor{gray!15}{2B-it}  & \cellcolor{gray!15}--  & \cellcolor{gray!15}60.3  & \cellcolor{gray!15}22.7  & \cellcolor{gray!15}67.3 \\
& 
& \cellcolor{gray!15}{9B-it} & \cellcolor{gray!15}--  & \cellcolor{gray!15}87.6 & \cellcolor{gray!15}43.5 & \cellcolor{gray!15}85.3 \\

\cmidrule(lr){2-7}

& \multirow{2}{*}{\plainbox{Direct}}
& {2B} & +\method$_{\text{from {9B}}}$  & 9.5 & 6.4 & 37.7 \\
& 
& {9B}  & +\method$_{\text{from {2B}}}$  & 60.1  & 26.4  & 74.3 \\

\midrule

\multirow{9}{*}{\rotatebox[origin=c]{90}{\texttt{Qwen2.5}}}
& \multirow{2}{*}{\graybox{Direct}}
& \cellcolor{gray!15}{1.5B}  & \cellcolor{gray!15}--  & \cellcolor{gray!15}11.1  & \cellcolor{gray!15}13.5  & \cellcolor{gray!15}48.0 \\
& 
& \cellcolor{gray!15}{7B} & \cellcolor{gray!15}--  & \cellcolor{gray!15}85.2 & \cellcolor{gray!15}46.1 & \cellcolor{gray!15}90.3 \\

\cmidrule(lr){2-7}

& \multirow{2}{*}{\graybox{CoT}}
& \cellcolor{gray!15}{1.5B}  & \cellcolor{gray!15}--  & \cellcolor{gray!15}67.5  & \cellcolor{gray!15}30.8  & \cellcolor{gray!15}76.3 \\
& 
& \cellcolor{gray!15}{7B} & \cellcolor{gray!15}--  & \cellcolor{gray!15}87.0 & \cellcolor{gray!15}48.8 & \cellcolor{gray!15}85.0 \\

\cmidrule(lr){2-7}

& \multirow{2}{*}{\graybox{CoT}}
& \cellcolor{gray!15}{1.5B-Instruct}  & \cellcolor{gray!15}--  & \cellcolor{gray!15}65.0  & \cellcolor{gray!15}26.7  & \cellcolor{gray!15}74.7 \\
& 
& \cellcolor{gray!15}7B-Instruct & \cellcolor{gray!15}--  & \cellcolor{gray!15}90.4 & \cellcolor{gray!15}46.1 & \cellcolor{gray!15}91.7 \\

\cmidrule(lr){2-7}

& \multirow{2}{*}{\plainbox{Direct}}
& 1.5B  & +\method$_{\text{from {7B}}}$  & 59.7  & 31.2  & 78.0 \\
& 
& 7B & +\method$_{\text{from {1.5B}}}$  & 85.3 & 46.5 & 89.3 \\

\midrule

\multirow{9}{*}{\rotatebox[origin=c]{90}{\texttt{Qwen3}}}
& \multirow{2}{*}{\graybox{Direct}}
& \cellcolor{gray!15}{4B-Base}  & \cellcolor{gray!15}--  & \cellcolor{gray!15}89.6  & \cellcolor{gray!15}51.8  & \cellcolor{gray!15}89.7 \\
& 
& \cellcolor{gray!15}{8B-Base} & \cellcolor{gray!15}--  & \cellcolor{gray!15}85.3 & \cellcolor{gray!15}50.5 & \cellcolor{gray!15}93.3 \\

\cmidrule(lr){2-7}

& \multirow{2}{*}{\graybox{CoT}}
& \cellcolor{gray!15}{4B-Base}  & \cellcolor{gray!15}--  & \cellcolor{gray!15}84.9  & \cellcolor{gray!15}50.5  & \cellcolor{gray!15}83.0 \\
& 
& \cellcolor{gray!15}{8B-Base} & \cellcolor{gray!15}--  & \cellcolor{gray!15}89.4 & \cellcolor{gray!15}51.6 & \cellcolor{gray!15}86.7 \\

\cmidrule(lr){2-7}

& \multirow{2}{*}{\graybox{CoT}}
& \cellcolor{gray!15}{4B}  & \cellcolor{gray!15}--  & \cellcolor{gray!15}91.1  & \cellcolor{gray!15}51.9  & \cellcolor{gray!15}92.3 \\
& 
& \cellcolor{gray!15}{8B} & \cellcolor{gray!15}--  & \cellcolor{gray!15}81.6 & \cellcolor{gray!15}53.4 & \cellcolor{gray!15}86.7 \\

\cmidrule(lr){2-7}

& \multirow{2}{*}{\plainbox{Direct}}
& {4B}  & +\method$_{\text{from {8B}}}$  &  89.7 & 52.2  & 90.7 \\
& 
& {8B} & +\method$_{\text{from {4B}}}$  & 92.4 & 52.3 & 93.0 \\

\bottomrule
\end{tabular}
\end{table}

\clearpage
\clearpage
\subsection{Hyperparameter Search}
\label{sec:hyperparams-cot}

\subsubsection{Impact of Number of Examples on the Master Key}

We now investigate the impact of the number of examples $n$ used in computing the \masterkey.
Using the same shared prompt $p=P^{\vphantom{*}}_{(\SL,\SU)}$ and set of queries $\mathcal{D} = \{q_i\}_{i=1}^{n}$ we stack the final-token hidden states of $\SL$ and $\SU$ across queries at a fixed layer $l$:
\begin{equation*}
\begin{aligned}
X_{\SL} &=
\begin{bmatrix}
\big(\mathbf{h}_{\SL}^{(l)}(p \oplus q^{\vphantom{*}}_1)\big)^\top;\\
\big(\mathbf{h}_{\SL}^{(l)}(p \oplus q^{\vphantom{*}}_2)\big)^\top;\\
\vdots\\
\big(\mathbf{h}_{\SL}^{(l)}(p \oplus q^{\vphantom{*}}_n)\big)^\top\\
\end{bmatrix}
\in\mathbb{R}^{n\times d^{\vphantom{*}}_{\mathcal{S}}},
\qquad
X_{\SU} &=
\begin{bmatrix}
\big(\mathbf{h}_{\SU}^{(l)}(p \oplus q^{\vphantom{*}}_1)\big)^\top;\\
\big(\mathbf{h}_{\SU}^{(l)}(p \oplus q^{\vphantom{*}}_2)\big)^\top;\\
\vdots\\
\big(\mathbf{h}_{\SU}^{(l)}(p \oplus q^{\vphantom{*}}_n)\big)^\top\\
\end{bmatrix}
\in\mathbb{R}^{n\times d^{\vphantom{*}}_{\mathcal{S}}},
\end{aligned}
\end{equation*}
where $d^{\vphantom{*}}_{\mathcal{S}}$ represents the hidden size of the \emph{Source} models.
We define the difference matrix \mbox{\(X=X^{\vphantom{*}}_{\mkern-2mu\SU}-X^{\vphantom{*}}_{\mkern-2mu\SL}\)}, where each row represents a per-example steering vector.
The corresponding covariance matrix is computed as
\[\Sigma = X^{\top}X \in \mathbb{R}^{d^{\vphantom{*}}_{\mathcal{S}} \times d^{\vphantom{*}}_{\mathcal{S}}}.\]
Let $\lambda^{\vphantom{*}}_1 \geq \lambda^{\vphantom{*}}_2 \geq \cdots \geq \lambda^{\vphantom{*}}_r$ be the eigenvalues of $\Sigma$, where $r = \min(n, d^{\vphantom{*}}_\mathcal{S})$ denotes the maximum possible rank. 
Following \citet{skean2025layerlayeruncoveringhidden}, we define the normalized eigenvalues as
\begin{equation}
\tilde{\lambda}^{\vphantom{*}}_i = \frac{\lambda^{\vphantom{*}}_i}{\sum_{j=1}^{r}\lambda^{\vphantom{*}}_j},
\end{equation}
and the spectral entropy as
\begin{equation}
H(\Sigma) = -\sum_{i=1}^{r} \tilde{\lambda}^{\vphantom{*}}_i \log \tilde{\lambda}^{\vphantom{*}}_i.
\end{equation}

The spectral entropy serves as a measure of the distributional compression of the steering vectors within the latent space.
A lower entropy indicates a more \emph{compressed} representation, where a small number of dominant eigenvalues capture the majority of the variance.
Conversely, a higher entropy reflects a more \emph{diffuse} representation, where the \masterkey is distributed more broadly across multiple orthogonal directions.

\begin{figure*}[!hp]
	\centering

	\begin{subfigure}{\linewidth}
		\centering
		\includegraphics[width=\linewidth]{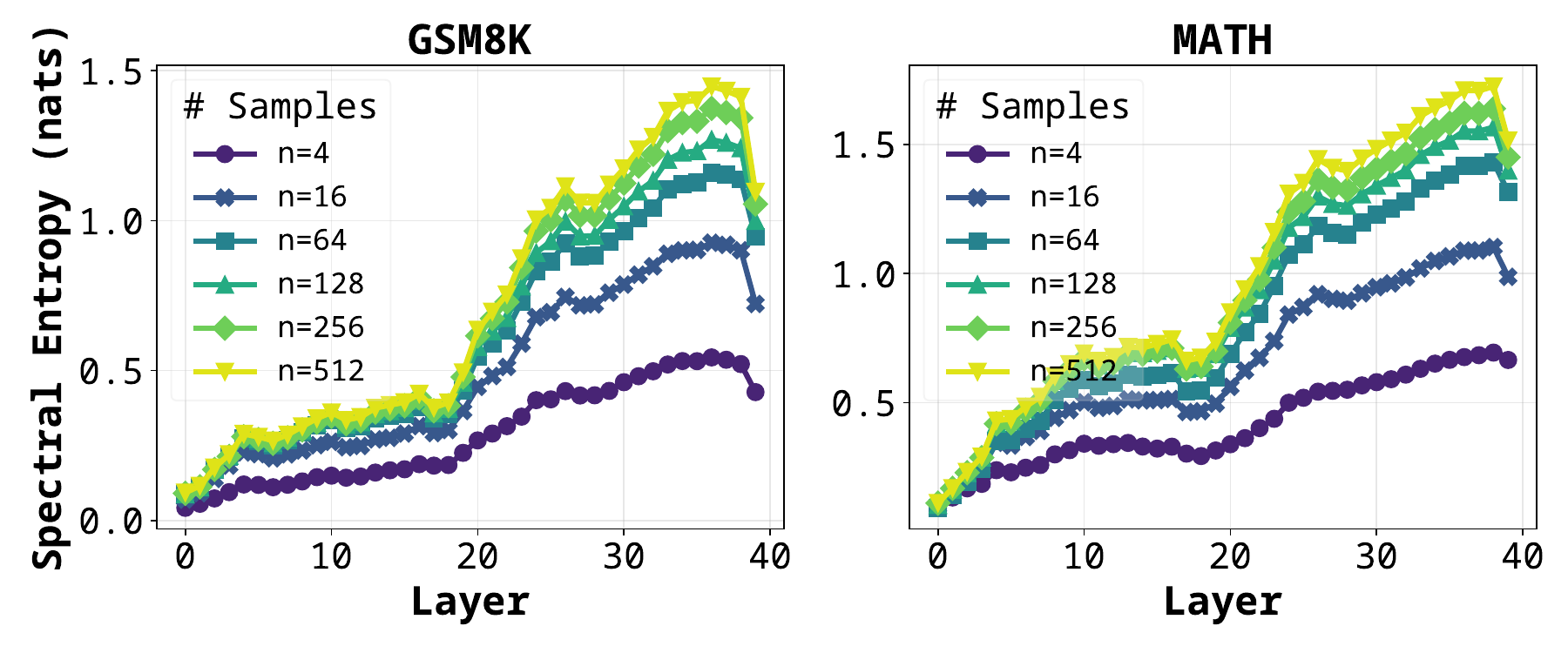}
        \vspace{-7mm}
        \caption{\texttt{Qwen1.5-14B}}
	\end{subfigure}
    \\
    \begin{subfigure}{\linewidth}
		\centering
		\includegraphics[width=\linewidth]{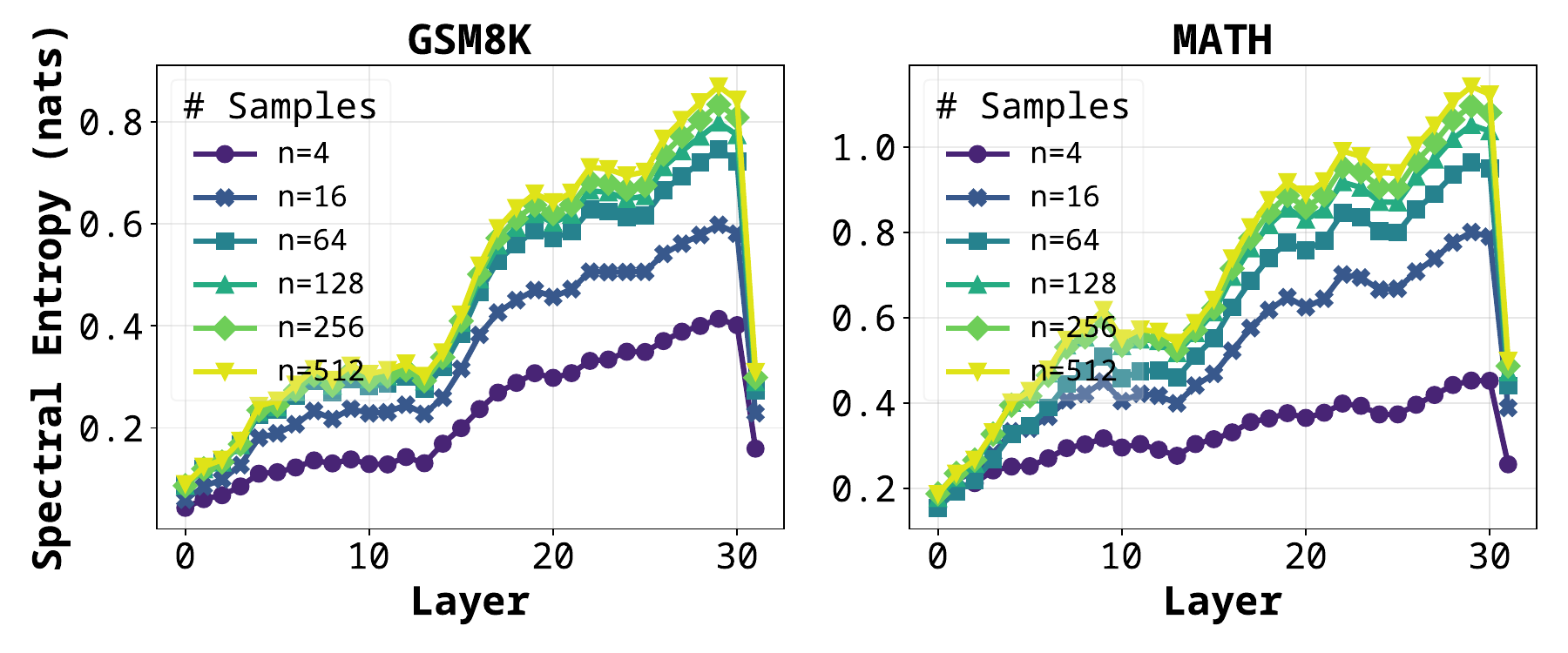}
        \vspace{-7mm}
        \caption{\texttt{Qwen1.5-7B}}
	\end{subfigure}
    \\
    \begin{subfigure}{\linewidth}
		\centering
		\includegraphics[width=\linewidth]{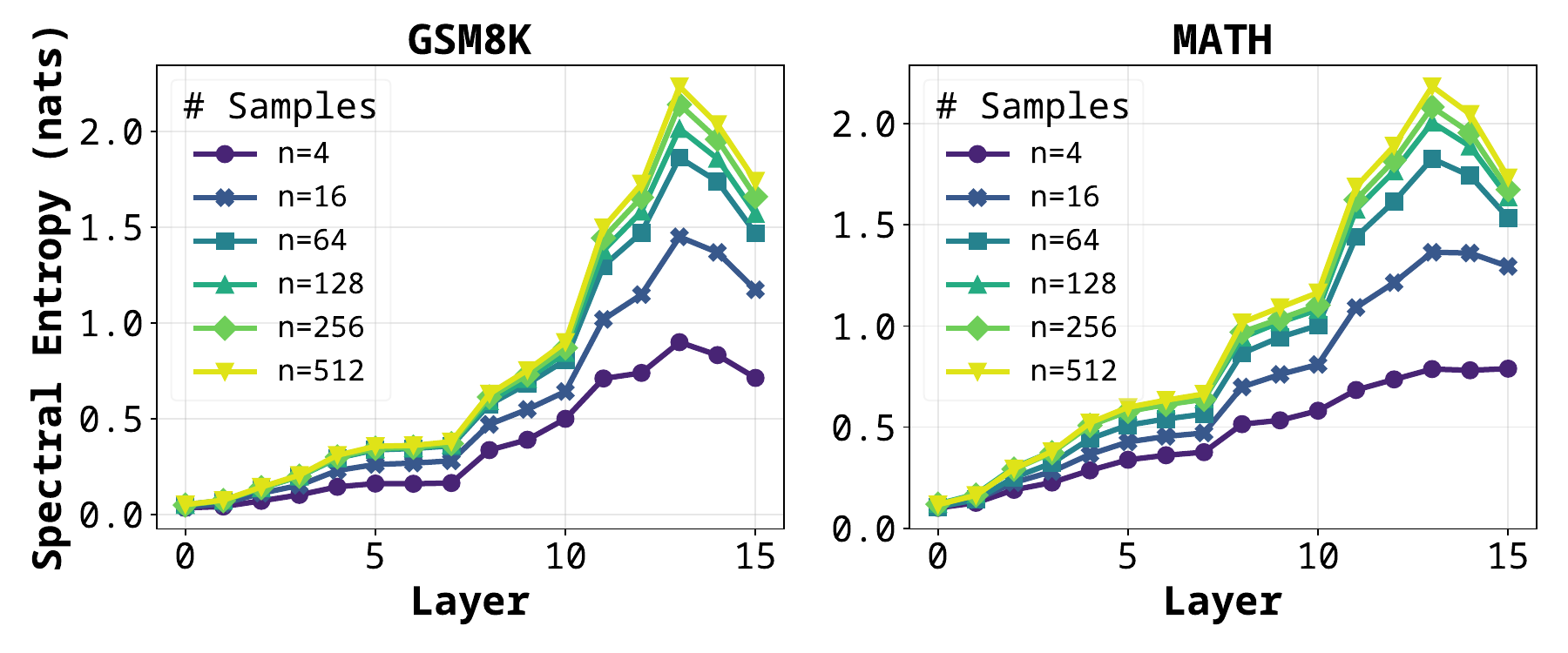}
        \vspace{-7mm}
        \caption{\texttt{OLMo-2-1B}}
	\end{subfigure}
	\caption{
        \textbf{Spectral Entropy of the Covariance Matrix:}  Increasing the number of examples leads to a corresponding increase in entropy --- providing evidence that the \masterkey captures more information with additional examples.
    }
	\label{fig:spectral-entropy}
\end{figure*}

Figure~\ref{fig:spectral-entropy} illustrates how spectral entropy evolves as a function of the number of examples $n$.
Empirically, we find that spectral entropy plateaus between approximately 1.4 and 2.5 nats across all evaluated datasets.
This corresponds to an effective rank in the range 4-12, (since $e^{1.4} \approx 4$ and $e^{2.5} \approx 12$), which is negligible relative to the model's latent dimensionality ($d^{\vphantom{*}}_{\mathcal{S}} \geq 1024$ for all models used in this work).

Notably, this extreme compression persists even as $n$ increases, providing strong evidence that the isolated capability resides in a stable, low-dimensional subspace.
We further observe that the rate of entropy growth begins to saturate across models and datasets as $n$ increases from 256 to 512, indicating diminishing returns in characterizing the \masterkey with sample sizes.
However, given the pronounced increase in entropy between $n=16$ and $n=64$, we assume that at least $64$ examples are required for an accurate and sufficiently complete estimate of the Master Key.

\subsubsection{Effect of Rank $k$ and Number of examples $n$ on The Linear Transformation}

\begin{figure*}[!hp]
	\centering

	\begin{subfigure}{\linewidth}
		\centering
		\includegraphics[width=\linewidth]{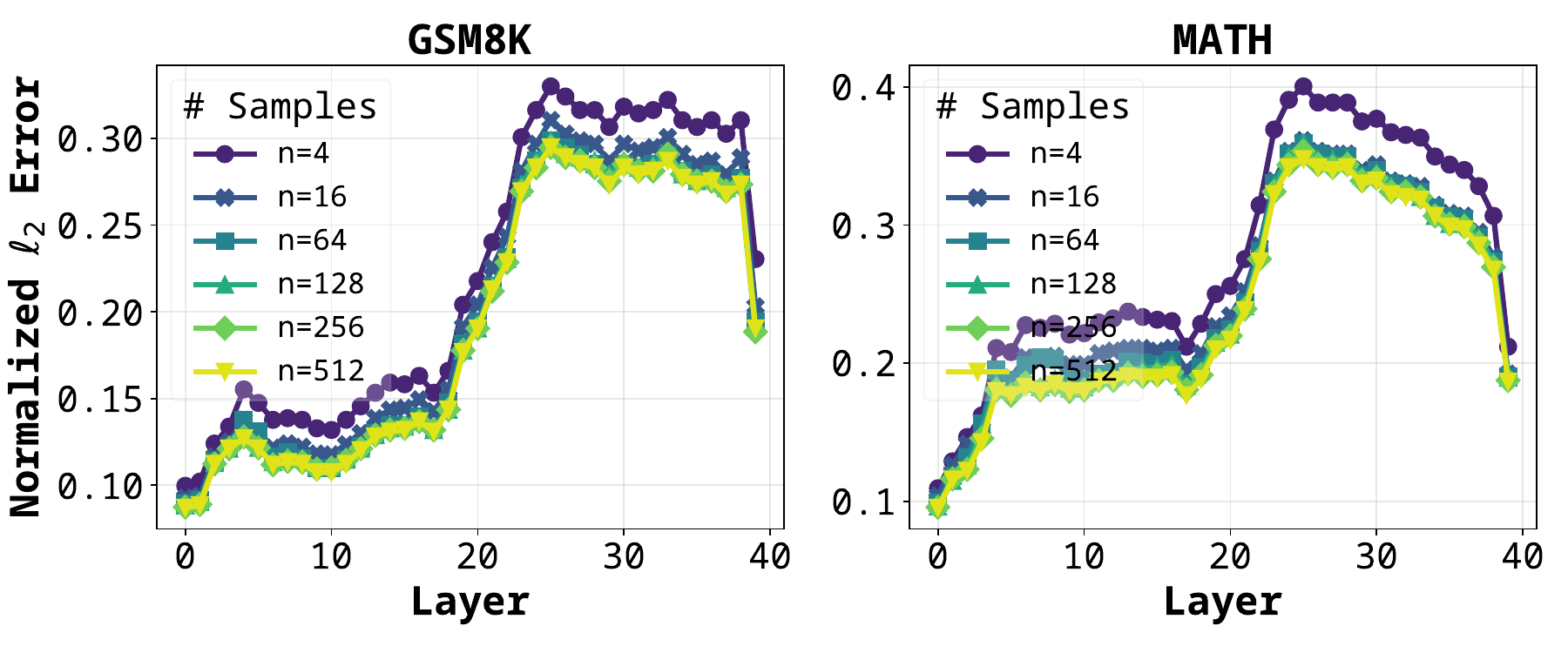}
        \vspace{-7mm}
        \caption{\transfer{\texttt{Qwen1.5-14B}}{7B}}
	\end{subfigure}
    \\
	\begin{subfigure}{\linewidth}
		\centering
		\includegraphics[width=\linewidth]{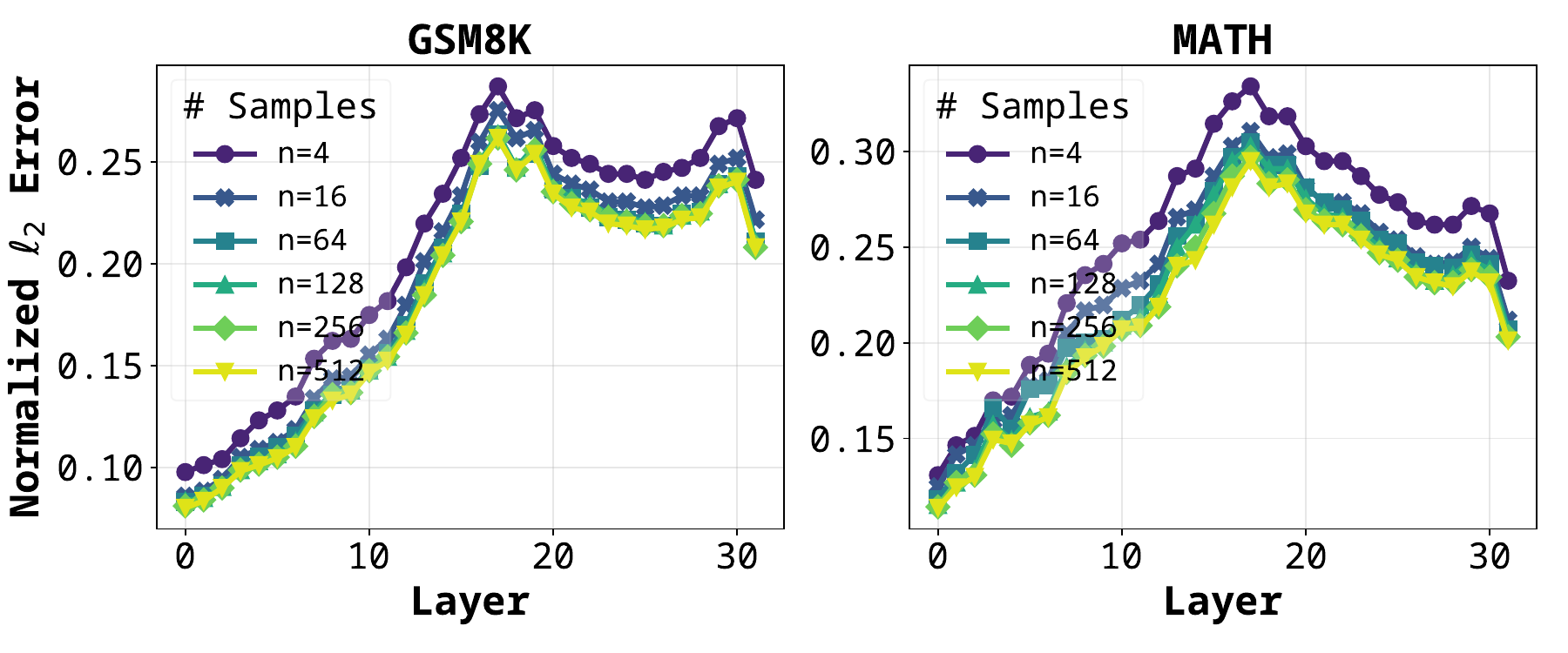}
        \vspace{-7mm}
        \caption{\transfer{\texttt{Qwen1.5-7B}}
        {14B}}
	\end{subfigure}
    \\
	\begin{subfigure}{\linewidth}
		\centering
		\includegraphics[width=\linewidth]{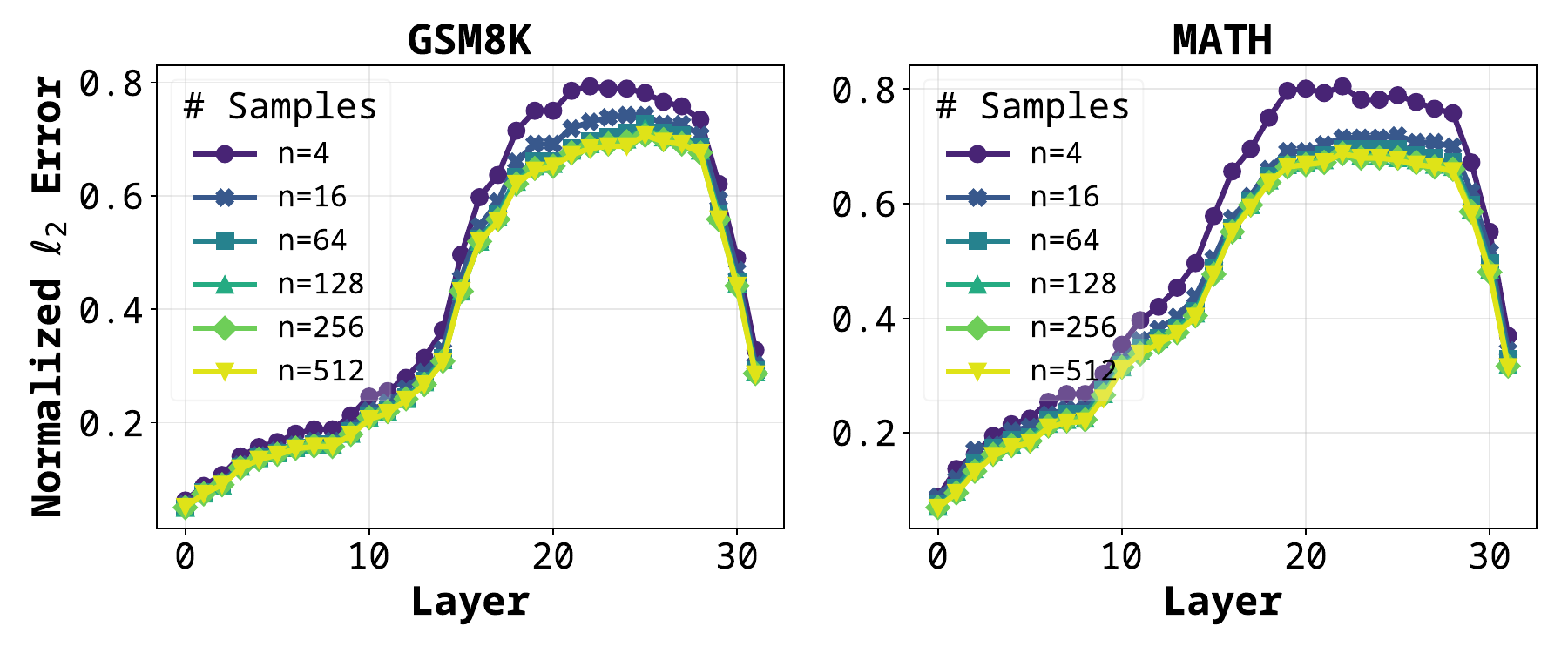}
        \vspace{-7mm}
        \caption{\transfer{\texttt{OLMo-2-7B}}{1B}}
	\end{subfigure}
	\caption{
        \textbf{Convergence in performance of the linear transformation at low ranks:} The normalized $\ell^{\vphantom{*}}_2$ error of the linear mapping as a function of number of samples $n$ with rank $k=4$ shows the diminishing impact of additional examples in rank-constrained settings.
    }
	\label{fig:transformation-error-num-ex}
\end{figure*}

\begin{figure*}[!hp]
	\centering

	\begin{subfigure}{\linewidth}
		\centering
		\includegraphics[width=\linewidth]{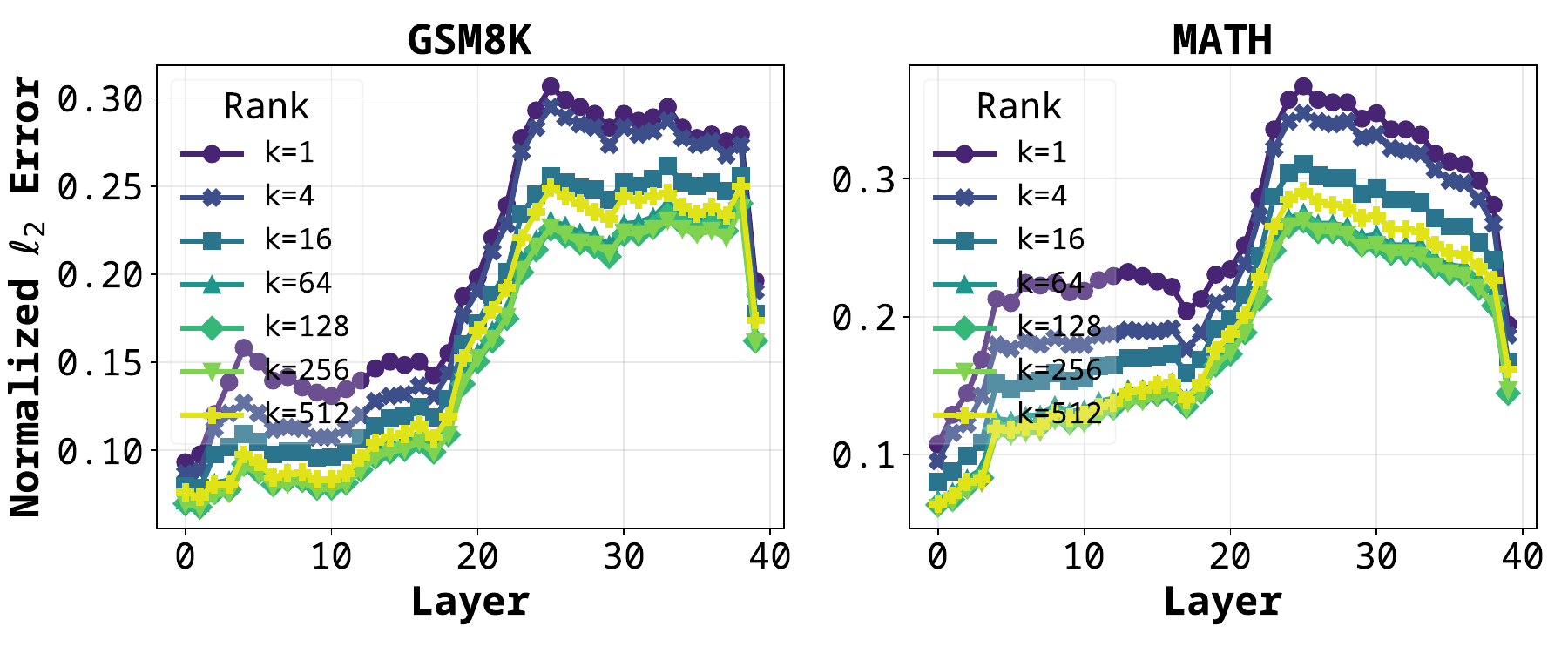}
        \vspace{-7mm}
		\caption{\transfer{\texttt{Qwen1.5-14B}}{7B}}
	\end{subfigure}
	\\
	\begin{subfigure}{\linewidth}
		\centering
		\includegraphics[width=\linewidth]{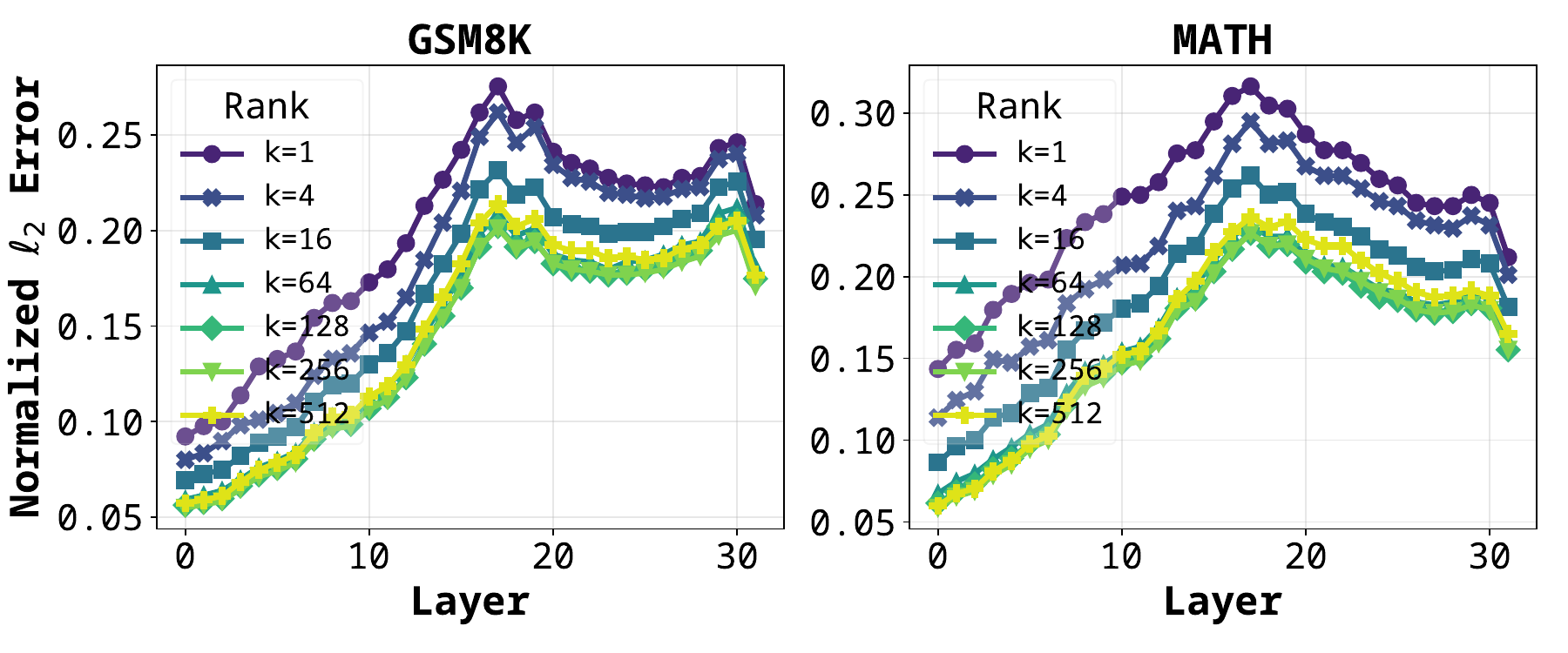}
        \vspace{-7mm}
		\caption{\transfer{\texttt{Qwen1.5-7B}}{14B}}
	\end{subfigure}
    \\
	\begin{subfigure}{\linewidth}
		\centering
		\includegraphics[width=\linewidth]{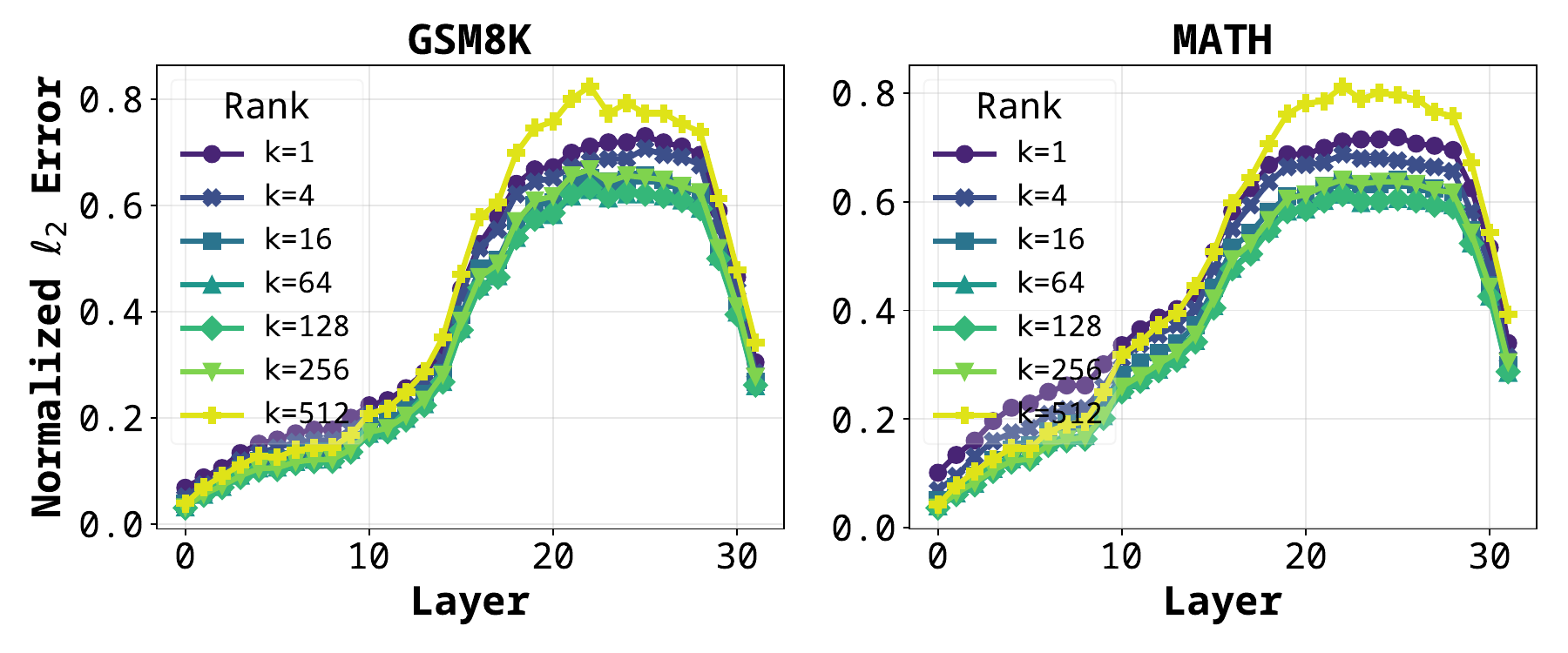}
        \vspace{-7mm}
		\caption{\transfer{\texttt{OLMo-2-7B}}{1B}}
	\end{subfigure}
	\caption{
        \textbf{Overfitting of the linear transformation at high ranks:} The normalized $\ell^{\vphantom{*}}_2$ error of the linear mapping as a function of rank $k$ with number of samples $n=512$ shows the overfitting of the transformation at high ranks.
    }
	\label{fig:transformation-error-rank}
\end{figure*}

Next, we evaluate the fidelity of the cross-model alignment by measuring its reconstruction error.
Concretely, we run the same set of $n$ queries through the \emph{Source Locked} model $\SL$ and the \emph{Target Locked} model $\TL$, extract final-token hidden states at layers $(l^{\vphantom{*}}_s,l^{\vphantom{*}}_t)$, and fit the low-rank mapping described in Section~\ref{sec:low_dim_match}.
For each query, we map the \emph{Source} hidden state into the \emph{Target} space and compute the $\ell^{\vphantom{*}}_2$ distance to the corresponding ground-truth \emph{Target} hidden state; we report the mean error over the $n$ examples.
Figures~\ref{fig:transformation-error-num-ex} and~\ref{fig:transformation-error-rank} show this mapping error as a function of the number of examples $n$ and the transformation rank $k$, respectively.

Recall that the rank $k$ controls the expressivity of the projection: larger $k$ allows the mapping to preserve and align more directions of variation, whereas smaller $k$ forces the alignment to concentrate on the most prominent structures shared across the two models.
Accordingly, higher rank can, in principle, encode more complex correspondences between latent features, but at the cost of increased sensitivity and a greater risk of overfitting.
In contrast, lower rank constrains the mapping to capture only the most dominant and robust shared structure, while prone to underfitting.

Figure~\ref{fig:transformation-error-num-ex} shows that in very low-rank regimes (e.g., $k \in \{1, 4\}$), the benefit of increasing the number of examples $n$ rapidly saturates. Specifically, while reconstruction error improves initially, it plateaus as early as $n \approx 64$. 
Consequently, for highly constrained projections, additional examples do not yield further gains because the mapping lacks sufficient capacity to represent finer structural correspondences; in this regime, the bottleneck is rank rather than sample size.

In contrast, even with an abundance of examples, we find that increasing the rank $k$ does not lead to a monotonic improvement in accuracy. 
While moderate ranks can reduce reconstruction error effectively, pushing $k$ beyond a threshold consistently degrades performance across models, with this effect becoming pronounced beyond $k \approx 128$ in our experiments (shown in Figure~\ref{fig:transformation-error-rank}).
This behavior is characteristic of overfitting: high-rank projections begin to align superficial, example-specific artifacts hindering generalization. 
These findings provide strong evidence that \emph{capabilities are better captured through low-rank transformations} because they effectively filter out spurious information, and highlights a critical limitation in previous approaches such as \cite{bello2025linearrepresentationtransferabilityhypothesis, oozeer2025activationspaceinterventionstransferred}, which utilize full-rank transformations that are both computationally intensive and prone to capturing noise.

Qualitatively, we observe complementary failure modes at the two extremes.
Examples~\ref{ex:cot-fail-low-rank-1}--\ref{ex:cot-fail-low-rank-3} illustrate cases where we scale $n$ while keeping $k$ highly constrained.
In these instances, although CoT-like behavior is occasionally elicited, it remains fragmented or poorly structured. 
Conversely, Examples~\ref{ex:cot-fail-high-rank-1},~\ref{ex:cot-fail-high-rank-2} demonstrate the emergence of unintended behaviors at high rank; for example, while CoT is induced, it may manifest in an undesired language (e.g., Chinese instead of English).

This tension between the \masterkey (which benefits from additional examples) and transformation (which overfits with too many examples) motivates the two regimes for mathematical reasoning transfer introduced in Section~\ref{sec:math-main}: the \sethlcolor{cyan!15}\hl{\emph{task-conditioned}} setting, which prioritizes in-distribution signals for estimating the \masterkey and mapping under limited data, and the \sethlcolor{purple!15}\hl{\emph{task-agnostic}} setting, which leverages abundant (but distribution-mismatched) data to fit a more stable alignment.

\subsubsection{Latent Space Geometry and Sensitivity}
\begin{figure*}[hp!]
	\centering

	\begin{subfigure}{0.32\linewidth}
		\centering
		\includegraphics[width=\linewidth]{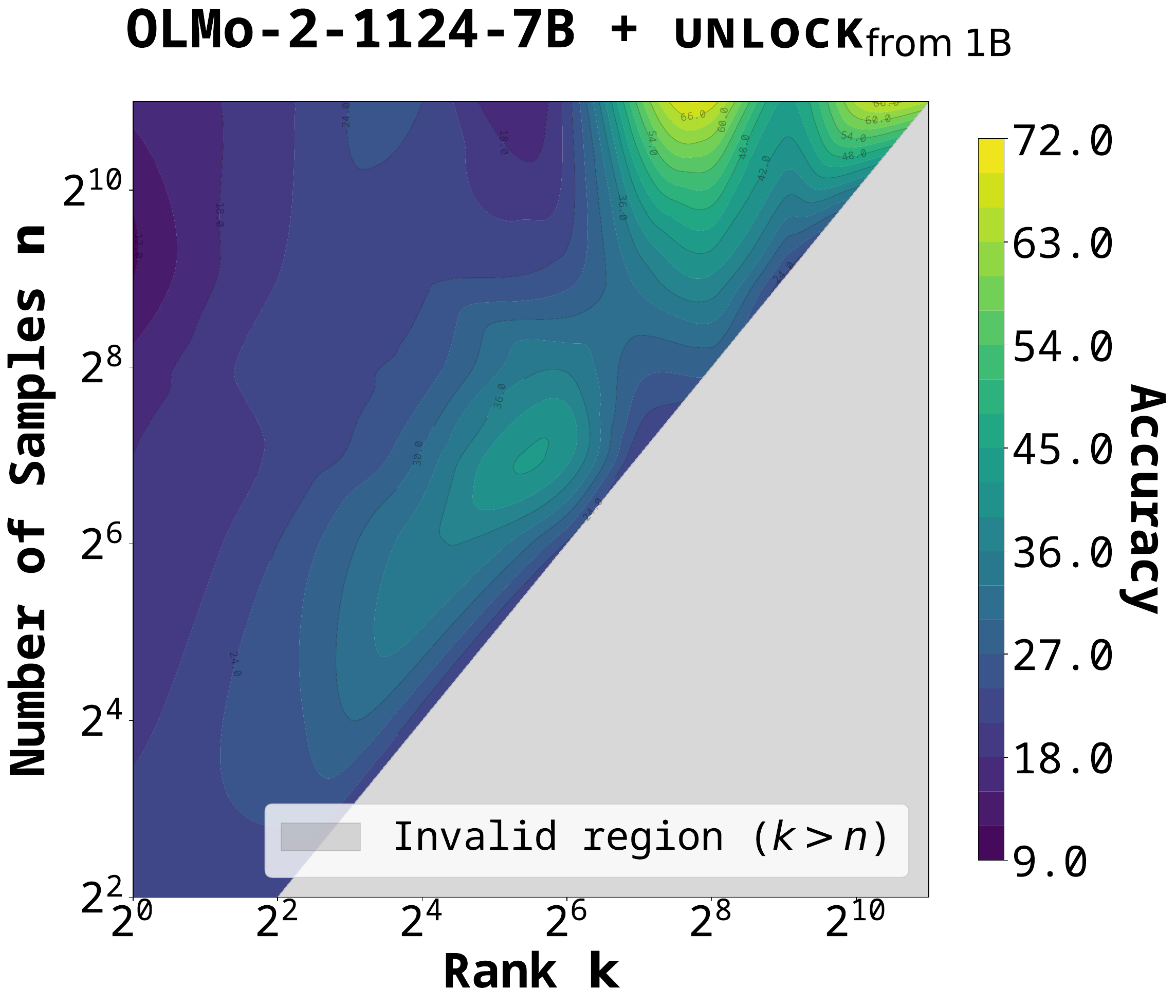}
		\caption{$\mathcal{D}$=GSM8K}
	\end{subfigure}
	\hfill
	\begin{subfigure}{0.32\linewidth}
		\centering
		\includegraphics[width=\linewidth]{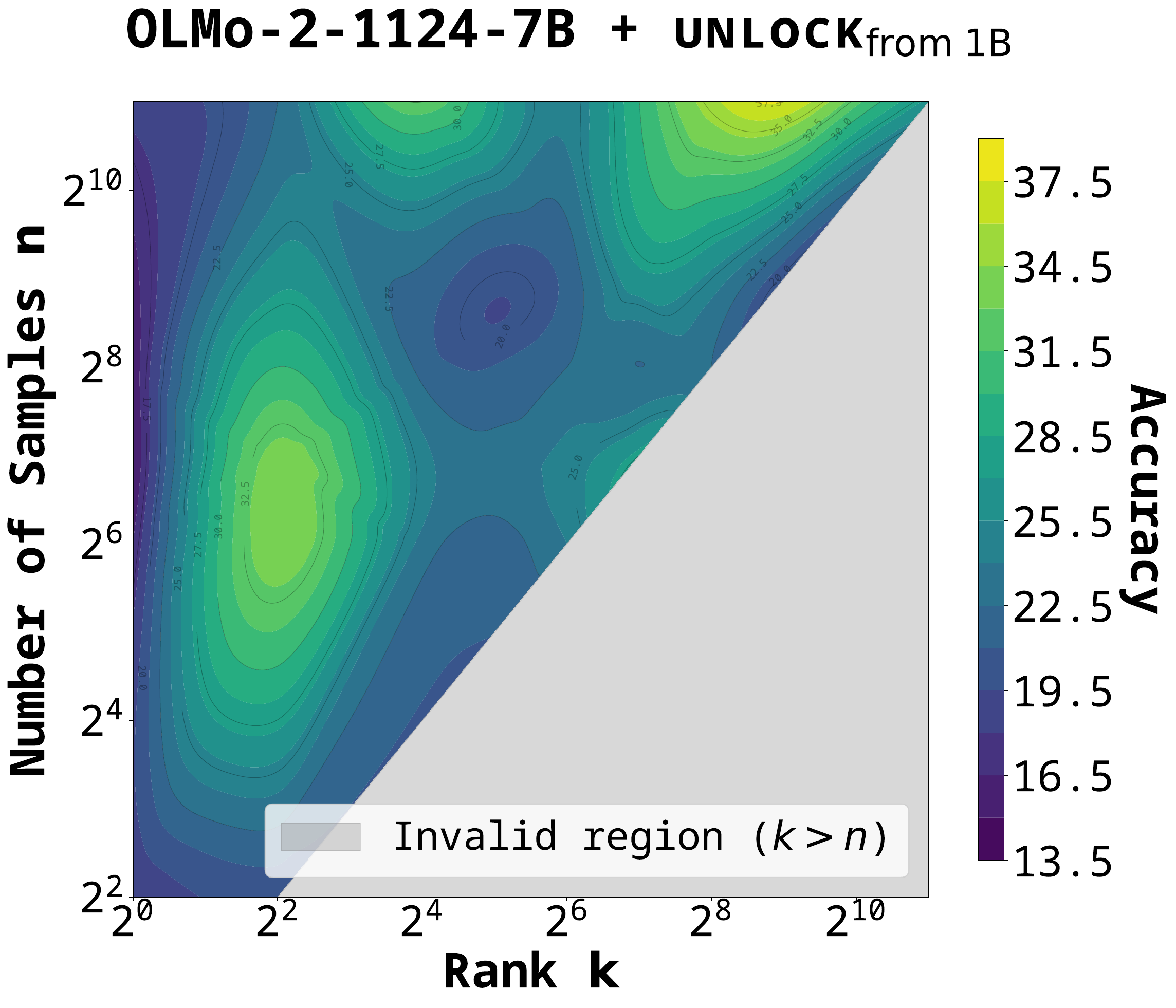}
		\caption{$\mathcal{D}$=MATH}
	\end{subfigure}
    \hfill
	\begin{subfigure}{0.32\linewidth}
		\centering
		\includegraphics[width=\linewidth]{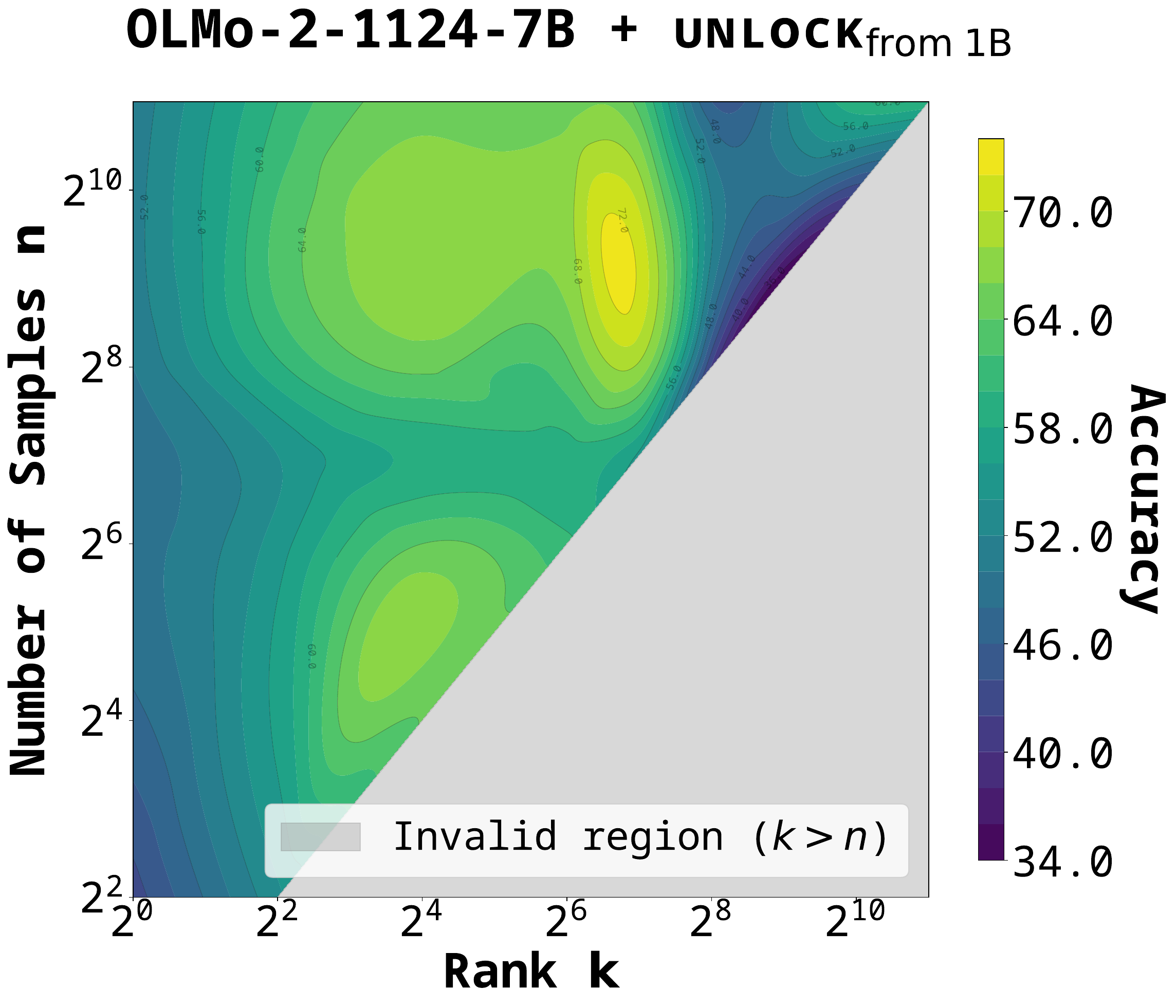}
		\caption{$\mathcal{D}$=SVAMP}
	\end{subfigure}
    \\[0.5em]
    \begin{subfigure}{0.32\linewidth}
		\centering
		\includegraphics[width=\linewidth]{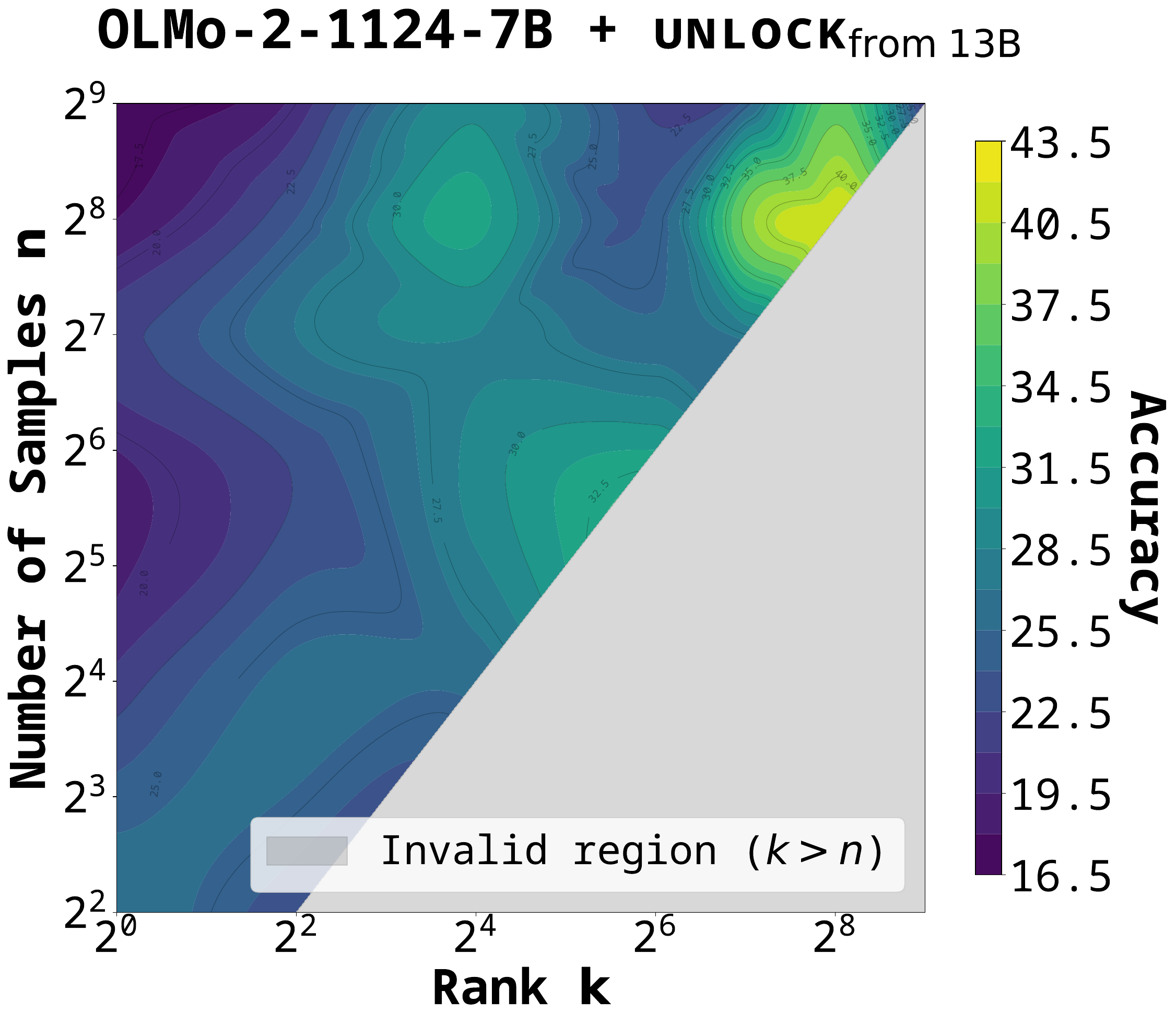}
		\caption{$\mathcal{D}$=GSM8K}
	\end{subfigure}
	\hfill
	\begin{subfigure}{0.32\linewidth}
		\centering
		\includegraphics[width=\linewidth]{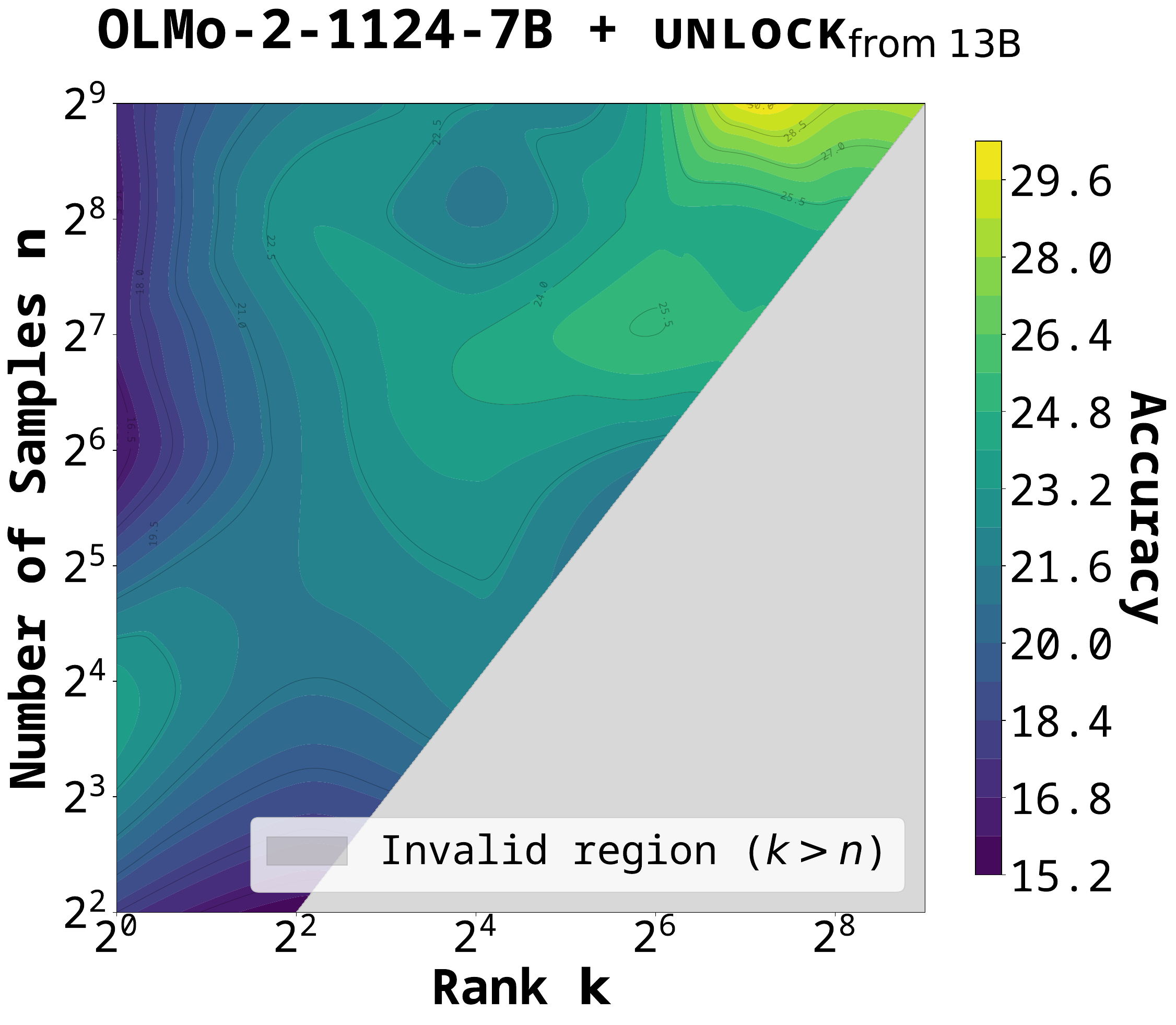}
		\caption{$\mathcal{D}$=MATH}
	\end{subfigure}
    \hfill
	\begin{subfigure}{0.32\linewidth}
		\centering
		\includegraphics[width=\linewidth]{figures/avg/topological_map_OLMo-2-1124-7B_+_OLMo-2-1124-13B_-_OLMo-2-1124-13B_math.pdf}
		\caption{$\mathcal{D}$=SVAMP}
	\end{subfigure}

    \vspace{-0.5em}
    \caption{\textbf{Representation Space with $\mathbf{\Phi}$=Avg:} Performance of \transfer{\texttt{OLMo-2-7B}}{1B} (top) and \transfer{\texttt{OLMo-2-7B}}{13B} (bottom) with the mean aggregator.}
	\label{fig:feature-maps-avg}
\end{figure*}

\begin{figure*}[hp!]
	\centering

	\begin{subfigure}{0.32\linewidth}
		\centering
		\includegraphics[width=\linewidth]{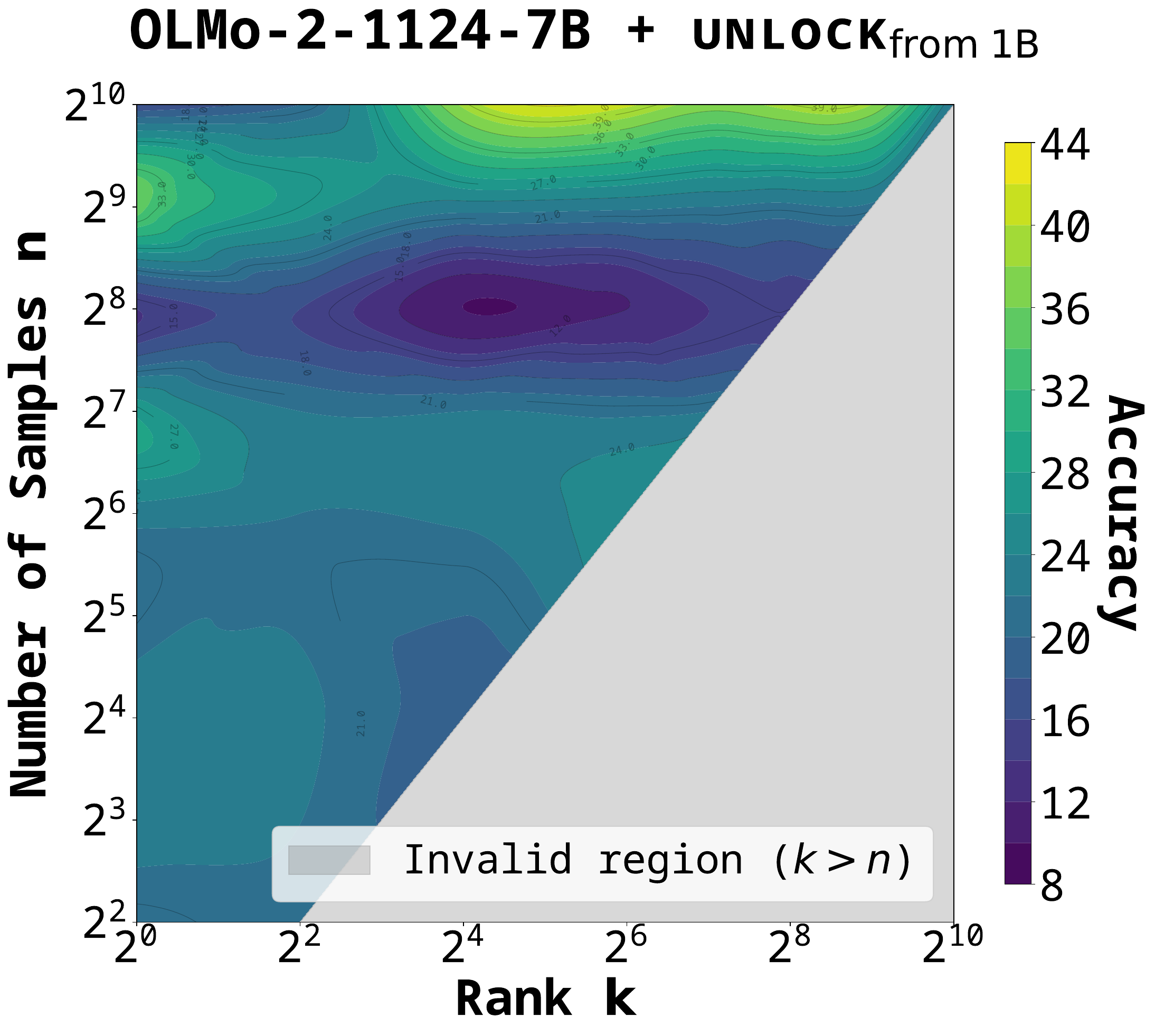}
		\caption{$\mathcal{D}$=GSM8K}
	\end{subfigure}
	\hfill
	\begin{subfigure}{0.32\linewidth}
		\centering
		\includegraphics[width=\linewidth]{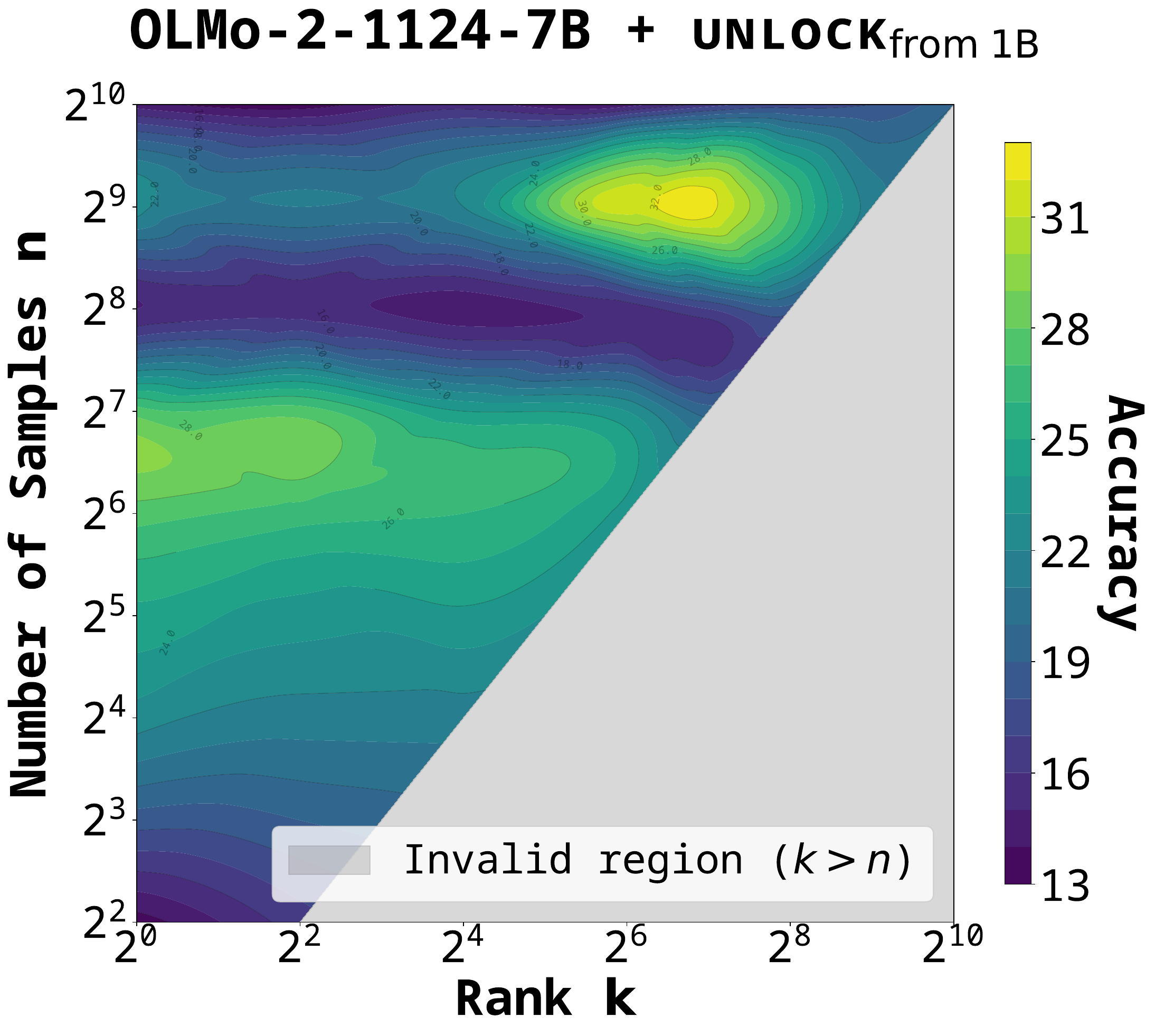}
		\caption{$\mathcal{D}$=MATH}
	\end{subfigure}
    \hfill
	\begin{subfigure}{0.32\linewidth}
		\centering
		\includegraphics[width=\linewidth]{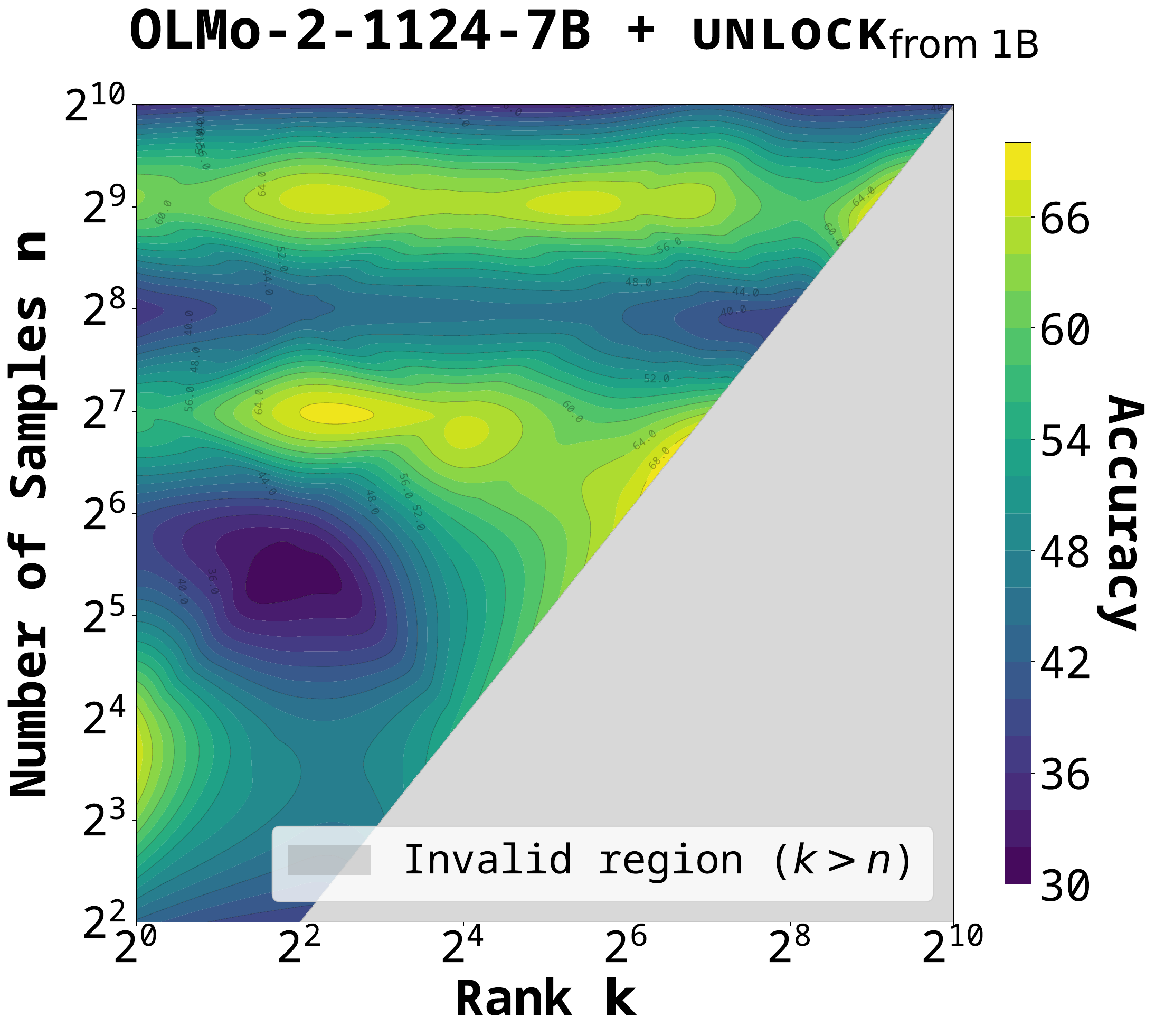}
		\caption{$\mathcal{D}$=SVAMP}
	\end{subfigure}
    \\[0.5em]
    \begin{subfigure}{0.32\linewidth}
		\centering
		\includegraphics[width=\linewidth]{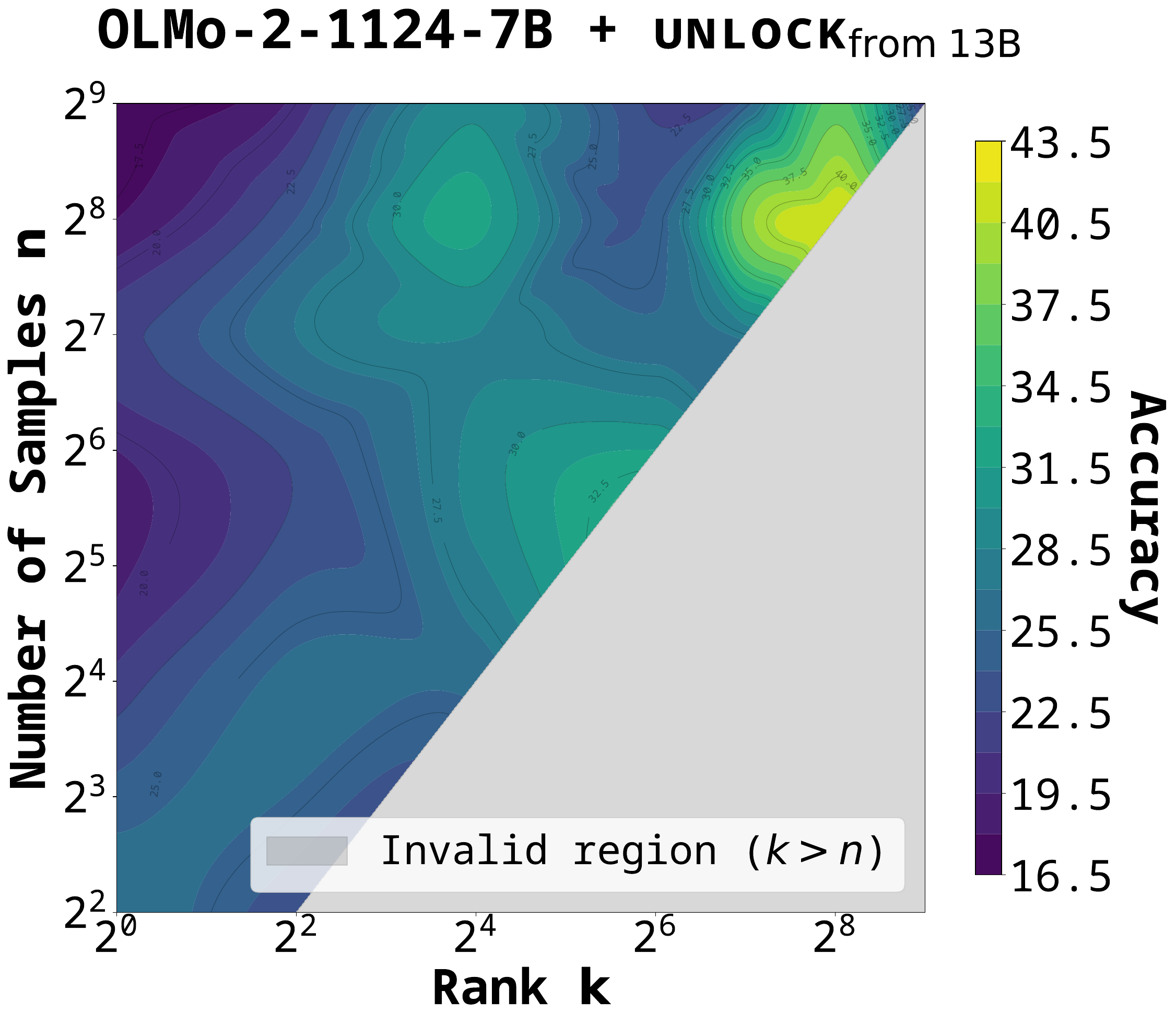}
		\caption{$\mathcal{D}$=GSM8K}
	\end{subfigure}
	\hfill
	\begin{subfigure}{0.32\linewidth}
		\centering
		\includegraphics[width=\linewidth]{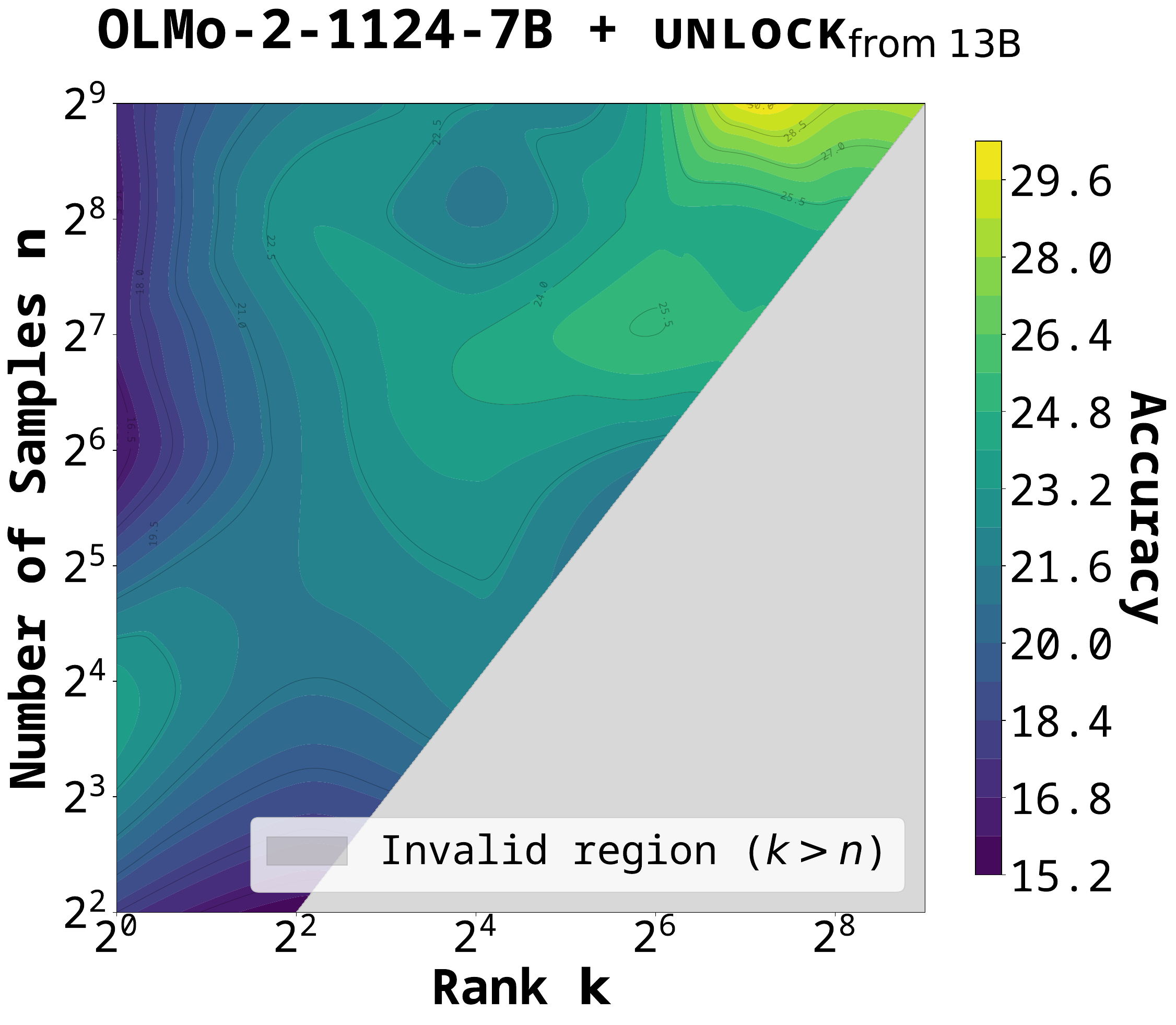}
		\caption{$\mathcal{D}$=MATH}
	\end{subfigure}
    \hfill
	\begin{subfigure}{0.32\linewidth}
		\centering
		\includegraphics[width=\linewidth]{figures/pca/topological_map_OLMo-2-1124-7B_+_OLMo-2-1124-13B_-_OLMo-2-1124-13B_math.pdf}
		\caption{$\mathcal{D}$=SVAMP}
	\end{subfigure}

    \vspace{-0.5em}
    \caption{\textbf{Representation Space with $\mathbf{\Phi}$=PCA:} Performance of \transfer{\texttt{OLMo-2-7B}}{1B} (top) and \transfer{\texttt{OLMo-2-7B}}{13B} (bottom) with the principal component aggregator.}
	\label{fig:feature-maps-pca}
\end{figure*}

Finally, we present topological visualizations of the feature space for \texttt{OLMo-2-7B} in Figure~\ref{fig:feature-maps-avg},\ref{fig:feature-maps-pca}. We find that successful capability transfer typically occurs within localized ``pockets'' of the latent manifold. This localization highlights the necessity of precise hyperparameter calibration.

In comparing different extraction strategies, we find that neither the principal component aggregator nor the mean aggregator provides a definitive advantage.
Across our benchmarks, the superior method is split approximately evenly, with neither consistently outperforming the other.
Ultimately, while subspace matching exhibits sensitivity to the chosen configuration, it yields substantial performance gains when the low-rank projection is well-optimized.
We leave a deeper exploration of this hyperparameter landscape to future work.

\begin{table*}[!ht]
\caption{\textbf{Optimal Hyperparameters for CoT Transfer:} The absence of a universal optimal hyperparameter combination across models suggests misalignment in the underlying representation spaces.}
\centering

\begin{tabularx}{\textwidth}{
	>{\centering\arraybackslash}p{5cm}
	|
	>{\centering\arraybackslash}X
	>{\centering\arraybackslash}X
	>{\centering\arraybackslash}X
    >{\centering\arraybackslash}X
    >{\centering\arraybackslash}X
}

\toprule
{$\TU$} &
{Dataset} &
{Agg. Method} &
{n} &
{k} &
{$\alpha$} \\

\midrule
\makecell{\transfer{\texttt{Qwen1.5-7B}}{\texttt{14B}}} &
\makecell{GSM8K\\MATH\\SVAMP} & 
\makecell{Avg\\Avg\\PCA} &
\makecell{512\\64\\64} & 
\makecell{64\\64\\16} &
\makecell{0.1\\0.05\\0.1} \\

\midrule
\makecell{\transfer{\texttt{Qwen1.5-14B}}{\texttt{7B}}} &
\makecell{GSM8K\\MATH\\SVAMP} & 
\makecell{PCA\\PCA\\PCA} &
\makecell{128\\512\\128} & 
\makecell{4\\512\\128} &
\makecell{0.1\\0.1\\0.1} \\

\midrule
\makecell{\transfer{\texttt{OLMo-2-1B}}{\texttt{7B}}} &
\makecell{GSM8K\\MATH\\SVAMP} & 
\makecell{Avg\\Avg\\Avg} &
\makecell{512\\512\\64} & 
\makecell{128\\256\\64} &
\makecell{0.2\\0.05\\0.2} \\

\midrule
\makecell{\transfer{\texttt{OLMo-2-7B}}{\texttt{1B}}} &
\makecell{GSM8K\\MATH\\SVAMP} & 
\makecell{Avg\\Avg\\Avg} &
\makecell{1024\\128\\256} & 
\makecell{1024\\4\\128} &
\makecell{0.5\\0.2\\0.2} \\

\midrule
\makecell{\transfer{\texttt{OLMo-2-7B}}{\texttt{13B}}} &
\makecell{GSM8K\\MATH\\SVAMP} & 
\makecell{PCA\\Avg\\PCA} &
\makecell{512\\512\\16} & 
\makecell{1\\128\\16} &
\makecell{0.1\\0.1\\0.1} \\

\midrule
\makecell{\transfer{\texttt{OLMo-2-13B}}{\texttt{7B}}} &
\makecell{GSM8K\\MATH\\SVAMP} & 
\makecell{Avg\\PCA\\Avg} &
\makecell{64\\64\\256} & 
\makecell{64\\4\\16} &
\makecell{0.1\\0.1\\0.05} \\

\midrule
\makecell{\transfer{\texttt{gemma-2-2B}}{\texttt{9B}}} &
\makecell{GSM8K\\MATH\\SVAMP} & 
\makecell{PCA\\PCA\\PCA} &
\makecell{16\\64\\512} & 
\makecell{1\\64\\4} &
\makecell{0.2\\0.05\\0.2} \\

\midrule
\makecell{\transfer{\texttt{gemma-2-9B}}{\texttt{2B}}} &
\makecell{GSM8K\\MATH\\SVAMP} & 
\makecell{Avg\\Avg\\PCA} &
\makecell{128\\512\\4} & 
\makecell{64\\64\\1} &
\makecell{0.1\\0.1\\0.2} \\

\bottomrule
\end{tabularx}
\label{tab:cot-hyperparams}
\end{table*}

\clearpage
\onecolumn

\begin{center}

\begin{tcolorbox}[
	exbox,
	colback=blue!10,
	colframe=blue!100,
	title=\centering\textbf{Question}
]

James decides to run 3 sprints 3 times a week.  He runs 60 meters each sprint. How many total meters does he run a week?

\end{tcolorbox}

\begin{tcolorbox}[
	exbox,
	colback=red!10,
	colframe=red!70!black!90,
	title=\centering\textbf{Locked Model $\TL$}
]

The final answer is <atok> 180 </atok>

\end{tcolorbox}

\begin{tcolorbox}[
	exbox,
	colback=green!10,
	colframe=green!40!black!90,
	title=\centering\textbf{Unlocked Model $\TU$}
]

Let's break down the problem step by step:\\\\
1. James runs 3 sprints 3 times a week.\\
2. He runs 60 meters each sprint.\\\\
Now, let's calculate the total meters he runs in a week:\\\\
Total meters = (Number of sprints) $\times$ (Number of times per week) $\times$ (Distance per sprint)\\\\
Total meters = 3 sprints $\times$ 3 times per week $\times$ 60 meters per sprint\\\\
Total meters = 9 $\times$ 60\\\\
Total meters = 540\\\\
The final answer is <atok> 540 </atok>

\end{tcolorbox}

\captionof{example}{\textbf{Successful CoT Transfer} for \transfer{\texttt{Qwen-1.5-14B}}{7B}}.
\label{ex:cot-success-1}

\end{center}

\vspace{10mm}

\begin{center}

\begin{tcolorbox}[
	breakable,
    exbox,
	colback=blue!10,
	colframe=blue!100,
	title=\centering\textbf{Question}
]

Every day, Wendi feeds each of her chickens three cups of mixed chicken feed, containing seeds, mealworms and vegetables to help keep them healthy.  She gives the chickens their feed in three separate meals. In the morning, she gives her flock of chickens 15 cups of feed.  In the afternoon, she gives her chickens another 25 cups of feed.  How many cups of feed does she need to give her chickens in the final meal of the day if the size of Wendi's flock is 20 chickens?

\end{tcolorbox}

\begin{tcolorbox}[
	breakable,
    exbox,
	colback=red!10,
	colframe=red!70!black!90,
	title=\centering\textbf{Locked Model $\TL$}
]

The final answer is <atok> 10 cups of feed </atok>

\end{tcolorbox}

\begin{tcolorbox}[
	breakable,
    exbox,
	colback=green!10,
	colframe=green!40!black!90,
	title=\centering\textbf{Unlocked Model $\TU$}
]

If Wendi feeds each of her chickens three cups of mixed chicken feed every day, and she has 20 chickens, then she needs to feed a total of 20 chickens $\times$ 3 cups of feed = 60 cups of feed each day.\\\\
In the morning, she gives her flock of chickens 15 cups of feed, and in the afternoon, she gives them another 25 cups of feed. So far, she has given her chickens a total of 15 cups + 25 cups = 40 cups of feed.\\\\
To find out how many cups of feed she needs to give her chickens in the final meal of the day, we need to subtract the amount of feed she has already given from the total amount of feed she needs to give each day.\\\\
The total amount of feed she needs to give each day is 60 cups, and she has already given 40 cups, so she needs to give her chickens 60 cups - 40 cups = 20 cups of feed in the final meal of the day.\\\\
The final answer is 20 cups of feed

\end{tcolorbox}

\captionof{example}{\textbf{Successful CoT Transfer} for \transfer{\texttt{Qwen-1.5-14B}}{7B}}
\label{ex:cot-success-2}

\end{center}

\vspace{10mm}

\begin{center}

\begin{tcolorbox}[
	breakable,
    exbox,
	colback=blue!10,
	colframe=blue!100,
	title=\centering\textbf{Question}
]

Every day, Wendi feeds each of her chickens three cups of mixed chicken feed, containing seeds, mealworms and vegetables to help keep them healthy.  She gives the chickens their feed in three separate meals. In the morning, she gives her flock of chickens 15 cups of feed.  In the afternoon, she gives her chickens another 25 cups of feed.  How many cups of feed does she need to give her chickens in the final meal of the day if the size of Wendi's flock is 20 chickens?

\end{tcolorbox}

\begin{tcolorbox}[
	breakable,
    exbox,
	colback=red!10,
	colframe=red!70!black!90,
	title=\centering\textbf{Locked Model $\TL$}
]

The final answer is <atok> 10 cups of feed </atok>

\end{tcolorbox}

\begin{tcolorbox}[
	breakable,
    exbox,
	colback=green!10,
	colframe=green!40!black!90,
	title=\centering\textbf{Unlocked Model $\TU$}
]

If Wendi feeds each of her chickens three cups of mixed chicken feed every day, and she has 20 chickens, then she needs to feed a total of 20 chickens $\times$ 3 cups of feed = 60 cups of feed each day.\\\\
In the morning, she gives her flock of chickens 15 cups of feed, and in the afternoon, she gives them another 25 cups of feed. So far, she has given her chickens a total of 15 cups + 25 cups = 40 cups of feed.\\\\
To find out how many cups of feed she needs to give her chickens in the final meal of the day, we need to subtract the amount of feed she has already given from the total amount of feed she needs to give each day.\\\\
The total amount of feed she needs to give each day is 60 cups, and she has already given 40 cups, so she needs to give her chickens 60 cups - 40 cups = 20 cups of feed in the final meal of the day.\\\\
The final answer is 20 cups of feed

\end{tcolorbox}

\captionof{example}{\textbf{Successful CoT Transfer} for \transfer{\texttt{Qwen-1.5-14B}}{7B}}
\label{ex:cot-success-3}

\end{center}
\clearpage

\begin{center}

\begin{tcolorbox}[
	breakable,
    exbox,
	colback=blue!10,
	colframe=blue!100,
	title=\centering\textbf{Question}
]

Robin bought a four-scoop ice cream cone having a scoop each of vanilla, chocolate, strawberry and cherry. In how many orders can the four scoops be stacked on the cone if they are stacked one on top of the other and chocolate cannot be on the bottom?

\end{tcolorbox}

\begin{tcolorbox}[
	breakable,
    exbox,
	colback=red!10,
	colframe=red!70!black!90,
	title=\centering\textbf{Locked Model $\TL$}
]

The final answer is <atok>12</atok>

\end{tcolorbox}

\begin{tcolorbox}[
	breakable,
    exbox,
	colback=green!10,
	colframe=green!40!black!90,
	title=\centering\textbf{Unlocked Model $\TU$}
]

To solve this problem, we need to find the number of ways to stack the four scoops on the cone, with the condition that chocolate cannot be on the bottom.\\\\\\\\
First, let's consider the number of ways to stack the scoops without any restrictions. There are 4 choices for the bottom scoop, 3 choices for the second scoop, 2 choices for the third scoop, and 1 choice for the top scoop. So, the total number of ways to stack the scoops without any restrictions is 4 * 3 * 2 * 1 = 24.\\\\\\\\
Now, let's consider the number of ways to stack the scoops with the condition that chocolate cannot be on the bottom. Since chocolate cannot be on the bottom, there are only 3 choices for the bottom scoop (vanilla, strawberry, or cherry). For the second scoop, there are 3 choices (any of the remaining 3 flavors). For the third scoop, there are 2 choices (any of the remaining 2 flavors). And for the top scoop, there is 1 choice (the remaining flavor).\\\\\\\\
So, the total number of ways to stack the scoops with the condition that chocolate cannot be on the bottom is 3 * 3 * 2 * 1 = 18.\\\\\\\\
Therefore, the number of ways to stack the four scoops on the cone with the condition that chocolate cannot be on the bottom is 18.\\\\\\\\
The final answer is <atok>18</atok>

\end{tcolorbox}

\captionof{example}{\textbf{Successful CoT Transfer} for \transfer{\texttt{Qwen-1.5-14B}}{7B}}
\label{ex:cot-success-4}

\end{center}
\clearpage

\begin{center}

\begin{tcolorbox}[
	breakable,
    exbox,
	colback=blue!10,
	colframe=blue!100,
	title=\centering\textbf{Question}
]

The profit from a business transaction is shared among 2 business partners, Mike and Johnson in the ratio 2:5 respectively. If Johnson got \$2500, how much will Mike have after spending some of his share on a shirt that costs \$200?

\end{tcolorbox}

\begin{tcolorbox}[
	breakable,
    exbox,
	colback=red!10,
	colframe=red!70!black!90,
	title=\centering\textbf{Locked Model $\TL$}
]

The final answer is <atok> \$1600 </atok>

\end{tcolorbox}

\begin{tcolorbox}[
	breakable,
    exbox,
	colback=green!10,
	colframe=green!40!black!90,
	title=\centering\textbf{Unlocked Model $\TU$}
]

\begin{CJK*}{UTF8}{gbsn}
每份利润为2500/5=500美元。因此，Mike的份额为500*2=1000美元。如果他花了200美元买了一件衬衫，那么他将剩下1000-200=800美元。The final answer is 800.
\end{CJK*}

\end{tcolorbox}

\captionof{example}{\textbf{Failed CoT Transfer} at $n$=512, $k$=512 for \transfer{\texttt{Qwen-1.5-14B}}{7B}}
\label{ex:cot-fail-high-rank-1}

\end{center}

\vspace{10mm}

\begin{center}
\begin{tcolorbox}[
	breakable,
    exbox,
	colback=blue!10,
	colframe=blue!100,
	title=\centering\textbf{Question}
]

Alexis is applying for a new job and bought a new set of business clothes to wear to the interview. She went to a department store with a budget of \$200 and spent \$30 on a button-up shirt, \$46 on suit pants, \$38 on a suit coat, \$11 on socks, and \$18 on a belt. She also purchased a pair of shoes, but lost the receipt for them. She has \$16 left from her budget. How much did Alexis pay for the shoes?

\end{tcolorbox}

\begin{tcolorbox}[
	breakable,
    exbox,
	colback=red!10,
	colframe=red!70!black!90,
	title=\centering\textbf{Locked Model $\TL$}
]

The final answer is <atok>\$115</atok>

\end{tcolorbox}

\begin{tcolorbox}[
	breakable,
    exbox,
	colback=green!10,
	colframe=green!40!black!90,
	title=\centering\textbf{Unlocked Model $\TU$}
]

\begin{CJK*}{UTF8}{gbsn}
每件商品的價格如下：\\\\
- 衬衫：\$30\\
- 裤子：\$46\\
- 外套：\$38\\
- 袜子：\$11\\
- 腰帶：\$18\\
Alexis的預算為 \$200，她已經花了 \$30 + \$46 + \$38 + \$11 + \$18 = \$143。她還剩下 \$200 - \$143 = \$57。她還剩下 \$16，所以她花了 \$57 - \$16 = \$41 買鞋子。\\\\
The final answer is \$41
\end{CJK*}

\end{tcolorbox}

\captionof{example}{\textbf{Failed CoT Transfer} at $n=256$, $k=256$ for \transfer{\texttt{Qwen-1.5-14B}}{7B}}
\label{ex:cot-fail-high-rank-2}

\end{center}

\begin{center}

\begin{tcolorbox}[
	breakable,
    exbox,
	colback=blue!10,
	colframe=blue!100,
	title=\centering\textbf{Question}
]

Julie is reading a 120-page book. Yesterday, she was able to read 12 pages and today, she read twice as many pages as yesterday. If she wants to read half of the remaining pages tomorrow, how many pages should she read?

\end{tcolorbox}

\begin{tcolorbox}[
	breakable,
    exbox,
	colback=red!10,
	colframe=red!70!black!90,
	title=\centering\textbf{Locked Model $\TL$}
]

The final answer is <atok> \$20 </atok>

\end{tcolorbox}

\begin{tcolorbox}[
	breakable,
    exbox,
	colback=green!10,
	colframe=green!40!black!90,
	title=\centering\textbf{Unlocked Model $\TU$}
]

$120-12-24=84$ pages. Half of $84$ is $42$ pages. So, Julie should read $42$ pages tomorrow. The final answer is <atok>42</atok>

\end{tcolorbox}
\captionof{example}{\textbf{Weak CoT Transfer} at $n=4$, $k=1$ for \transfer{\texttt{Qwen-1.5-14B}}{7B}}
\label{ex:cot-fail-low-rank-1}

\end{center}

\vspace{10mm}

\begin{center}
\begin{tcolorbox}[
	breakable,
    exbox,
	colback=blue!10,
	colframe=blue!100,
	title=\centering\textbf{Question}
]

Natalia sold clips to 48 of her friends in April, and then she sold half as many clips in May. How many clips did Natalia sell altogether in April and May?

\end{tcolorbox}

\begin{tcolorbox}[
	breakable,
    exbox,
	colback=red!10,
	colframe=red!70!black!90,
	title=\centering\textbf{Locked Model $\TL$}
]

The final answer is <atok> 72 </atok>

\end{tcolorbox}

\begin{tcolorbox}[
	breakable,
    exbox,
	colback=green!10,
	colframe=green!40!black!90,
	title=\centering\textbf{Unlocked Model $\TU$}
]

$48 + \frac{48}{2} = 48 + 24 = 72$ clips\\\\
The final answer is 72.

\end{tcolorbox}
\captionof{example}{\textbf{Weak CoT Transfer} at $n=4$, $k=4$ for \transfer{\texttt{Qwen-1.5-14B}}{7B}}
\label{ex:cot-fail-low-rank-2}

\end{center}

\vspace{10mm}

\begin{center}
\begin{tcolorbox}[
	breakable,
    exbox,
	colback=blue!10,
	colframe=blue!100,
	title=\centering\textbf{Question}
]

James writes a 3-page letter to 2 different friends twice a week.  How many pages does he write a year?

\end{tcolorbox}

\begin{tcolorbox}[
	breakable,
    exbox,
	colback=red!10,
	colframe=red!70!black!90,
	title=\centering\textbf{Locked Model $\TL$}
]

The final answer is <atok>260</atok>

\end{tcolorbox}

\begin{tcolorbox}[
	breakable,
    exbox,
	colback=green!10,
	colframe=green!40!black!90,
	title=\centering\textbf{Unlocked Model $\TU$}
]

$24 \times 2 \times 52 = 2496$ pages\\\\
The final answer is 2496 </atok>

\end{tcolorbox}

\captionof{example}{\textbf{Weak CoT Transfer}  at $n=4$, $k=4$ for \transfer{\texttt{Qwen-1.5-14B}}{7B}}
\label{ex:cot-fail-low-rank-3}

\end{center}

\clearpage
\section{Additional Results for Unlocking Mathematical Reasoning}
\label{sec:math-appendix}

Our test suite consists of four mathematical reasoning benchmarks: AGIEval-Math \citep{zhong2023agievalhumancentricbenchmarkevaluating}, Deepmind-Math \citep{saxton2019analysingmathematicalreasoningabilities}, Minerva-Math \citep{lewkowycz2022solvingquantitativereasoningproblems}, and OlympiadBench \citep{he2024olympiadbenchchallengingbenchmarkpromoting}.
We withhold 32 examples from each dataset to use as the dev set for \sethlcolor{cyan!15}\hl{\emph{task-conditioned transfer}}.
We exclude these examples from the test sets across all settings.
For \sethlcolor{purple!15}\hl{\emph{task-agnostic transfer}}, we compute the \masterkey and linear transformation using data from MATH~\cite{hendrycks2021measuring}, and verify the robustness of \method on Gaokao2023En \citep{liao2024mario} and AMC23\footnote{\texttt{https://huggingface.co/datasets/AI-MO/\\aimo-validation-amc}}.
The best performing hyperparameters are used for evaluating on the test suite. 

We investigate four distinct model families: \texttt{Qwen2.5}~\cite{qwen2025qwen25technicalreport}, \texttt{Qwen3}~\cite{qwen3technicalreport}, \texttt{Ministral-3}~\cite{liu2026ministral3}, and \texttt{gemma-3}~\cite{gemmateam2025gemma3technicalreport}. 
For each family, the base model serves as the Locked variants $\TL$ and $\SL$, while a stronger post-trained model is selected as the Unlocked Source model $\SU$. 
We categorize these Unlocked models into two classes: 
\begin{enumerate}
    \item \emph{Instruction-tuned models}, optimized for general instruction following and trained with a combination of math, coding, and safety datasets; 
    \item \emph{Math-specific models}, specialized for math reasoning.
\end{enumerate}

We utilize the corresponding \texttt{-Instruct} or \texttt{-Chat} checkpoints publicly available on Hugging Face\footnote{\url{https://huggingface.co/models}} for the instruction-tuned models.
For the math-specific models, we employ \texttt{NVIDIA-OpenReasoning-Nemotron} ~\cite{ahmad2025opencodereasoningiisimpletesttime} and \texttt{NVIDIA-DLER-R1}~\cite{liu2025dler} for \texttt{Qwen2.5}, and \texttt{NVIDIA-Nemotron-Cascade}~\cite{wang2025nemotroncascadescalingcascadedreinforcement} and \texttt{Qwen3-Thinking}~\citep{qwen3technicalreport} for \texttt{Qwen3}. 
We omit \texttt{gemma-3} from the math-specific setting as no comparably strong math-oriented post-trained variants were identified for this family.

All models are prompted with the same \emph{CoT} prompt.
To reduce model- and dataset-specific variance, we do not apply chat templates or in-context demonstrations.
We evaluate with greedy decoding and a maximum generation length of 4096 tokens.
We report the results when using instruction-tuned Unlocked models in Table~\ref{tab:math-results-appendix} and math-specific models in Table~\ref{tab:math-specific-model-results-appendix}.

\subsection{Results \& Discussion:}

\subsubsection{Understanding the Impact of Unlocking}

\paragraph{Dependence on Capabilities present in $\TL$:}
We find that the gain of $\TU$ over $\TL$ depends not only on the strength of the Source contrast, measured by how much $\SU$ improves over $\SL$, but also the baseline competence of $\TL$.
For instance, \texttt{gemma-3} is the weakest-performing family in our experiments and underperforms its instruction-tuned counterpart by a wide margin, with average gaps of 32.37\% for \texttt{gemma-3-4B} and 31.65\% for \texttt{gemma-3-12B}, leaving limited scope for \method to recover post-training gains.
Accordingly, we observe modest improvements in this setting, and $\TU$ typically falls well short of $\TPT$.
Taken together, these results reinforce the interpretation that our method does not introduce new knowledge, but instead \emph{elicits and amplifies} capabilities already present but latent in the Target model.

\paragraph{What is Encoded in the Master Key?} We find that gains in accuracy typically arise from three types of changes:

\emph{(I.) Coherent reasoning traces:}
$\TL$ frequently fails to produce explicit step-by-step reasoning, or instead generates reasoning that is fragmented, inefficient, or prematurely terminated.
In contrast, $\TU$ more consistently produces coherent intermediate steps that connect the problem statement to the final answer.
Examples~\ref{ex:math-consistent-cot} and~\ref{ex:math-correct-reasoning} illustrate this effect.

\begin{figure*}[!hp]
	\centering

	\begin{subfigure}[t]{0.24\linewidth}
		\centering
		\includegraphics[width=\linewidth]{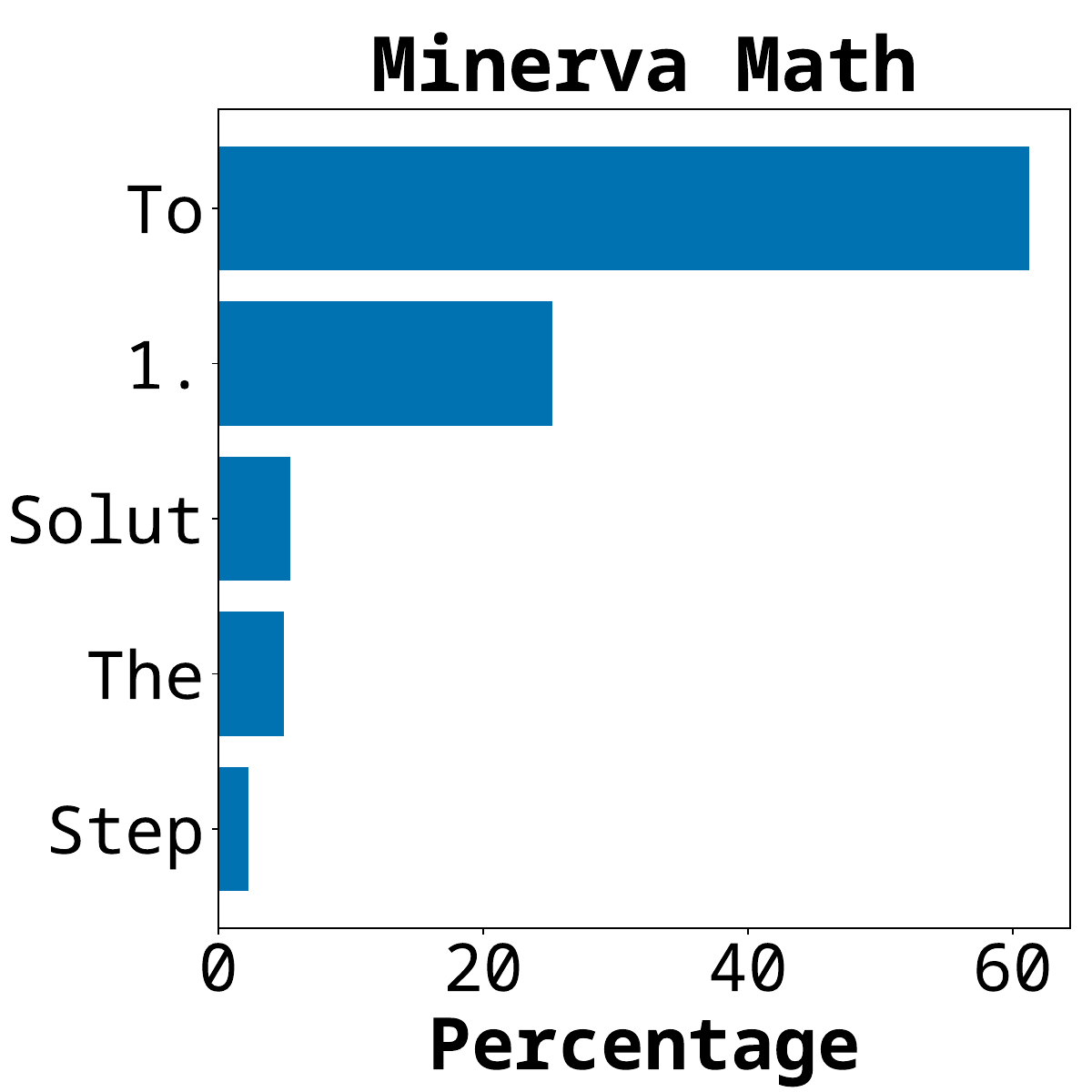}
		\caption{\centering\texttt{Qwen3-14B}}
	\end{subfigure}
	\hfill
	\begin{subfigure}[t]{0.24\linewidth}
		\centering
		\includegraphics[width=\linewidth]{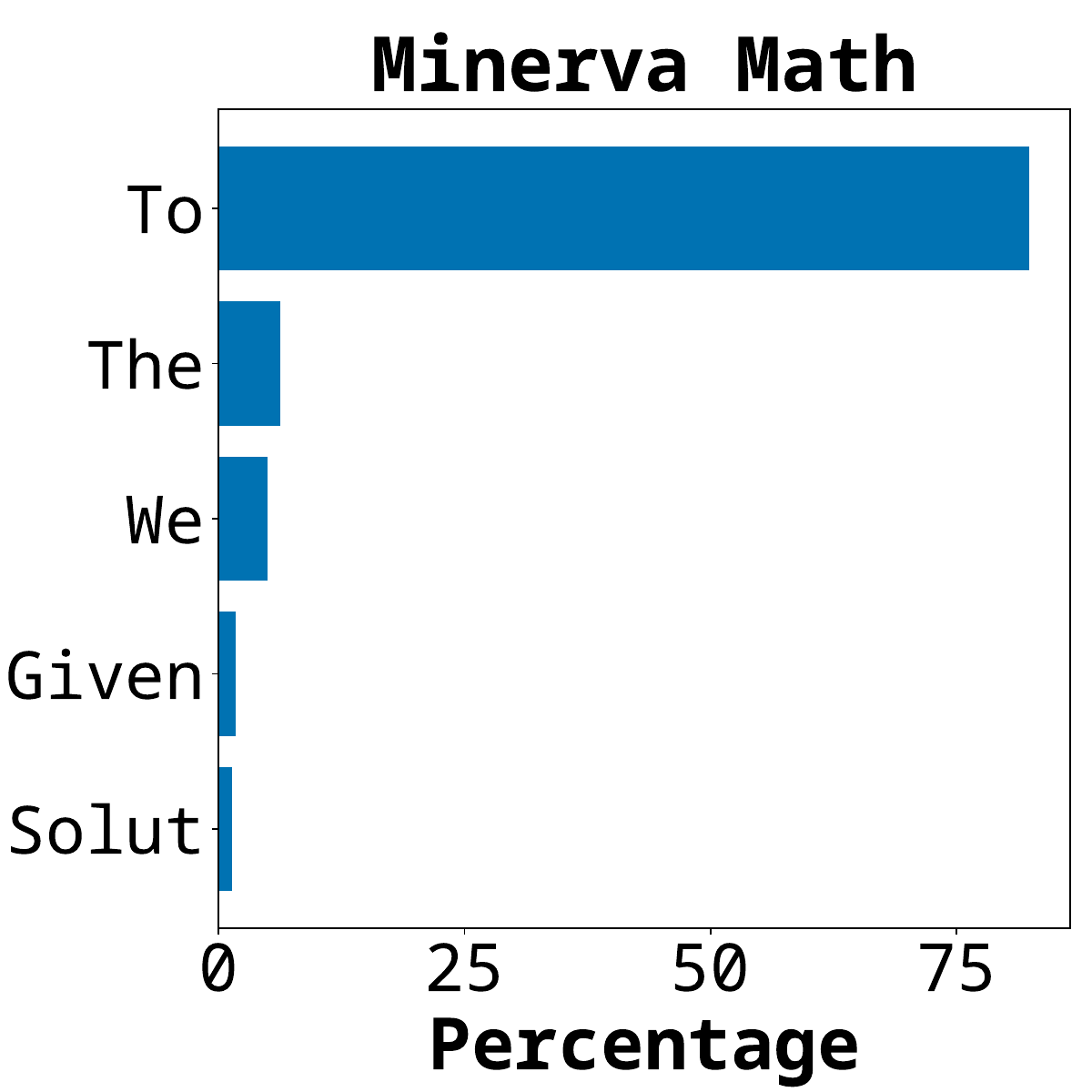}
		\caption{\centering\transfer{\texttt{Qwen3-14B}}{4B}}
	\end{subfigure}
    \hfill
    \begin{subfigure}[t]{0.24\linewidth}
		\centering
		\includegraphics[width=\linewidth]{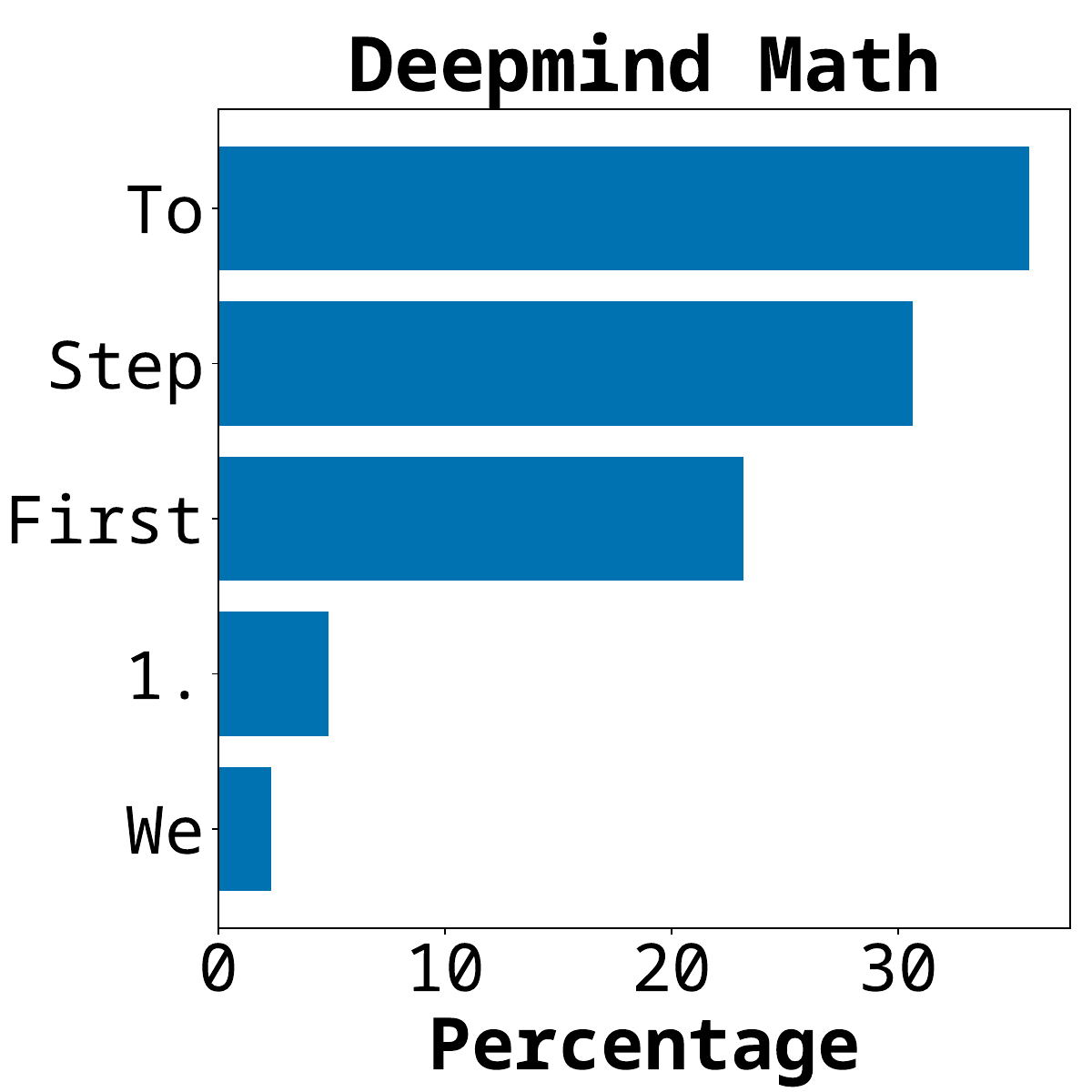}
		\caption{\centering\texttt{Qwen3-8B}}
	\end{subfigure}
	\hfill
	\begin{subfigure}[t]{0.24\linewidth}
		\centering
		\includegraphics[width=\linewidth]{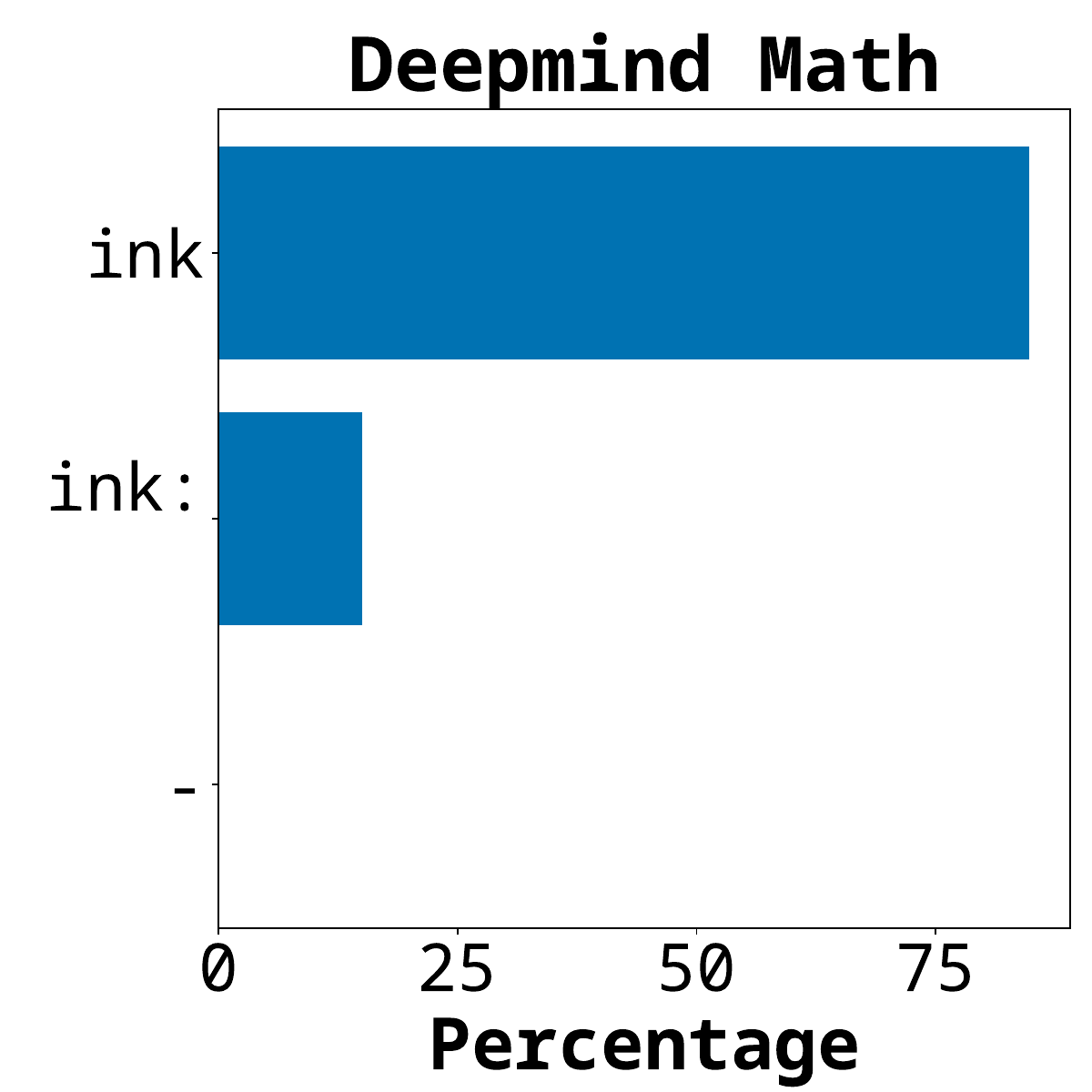}
		\caption{\centering\transfer{\texttt{Qwen3-8B}}{4B}}
	\end{subfigure}
    \\[0.7em]
    \begin{subfigure}[t]{0.24\linewidth}
		\centering
		\includegraphics[width=\linewidth]{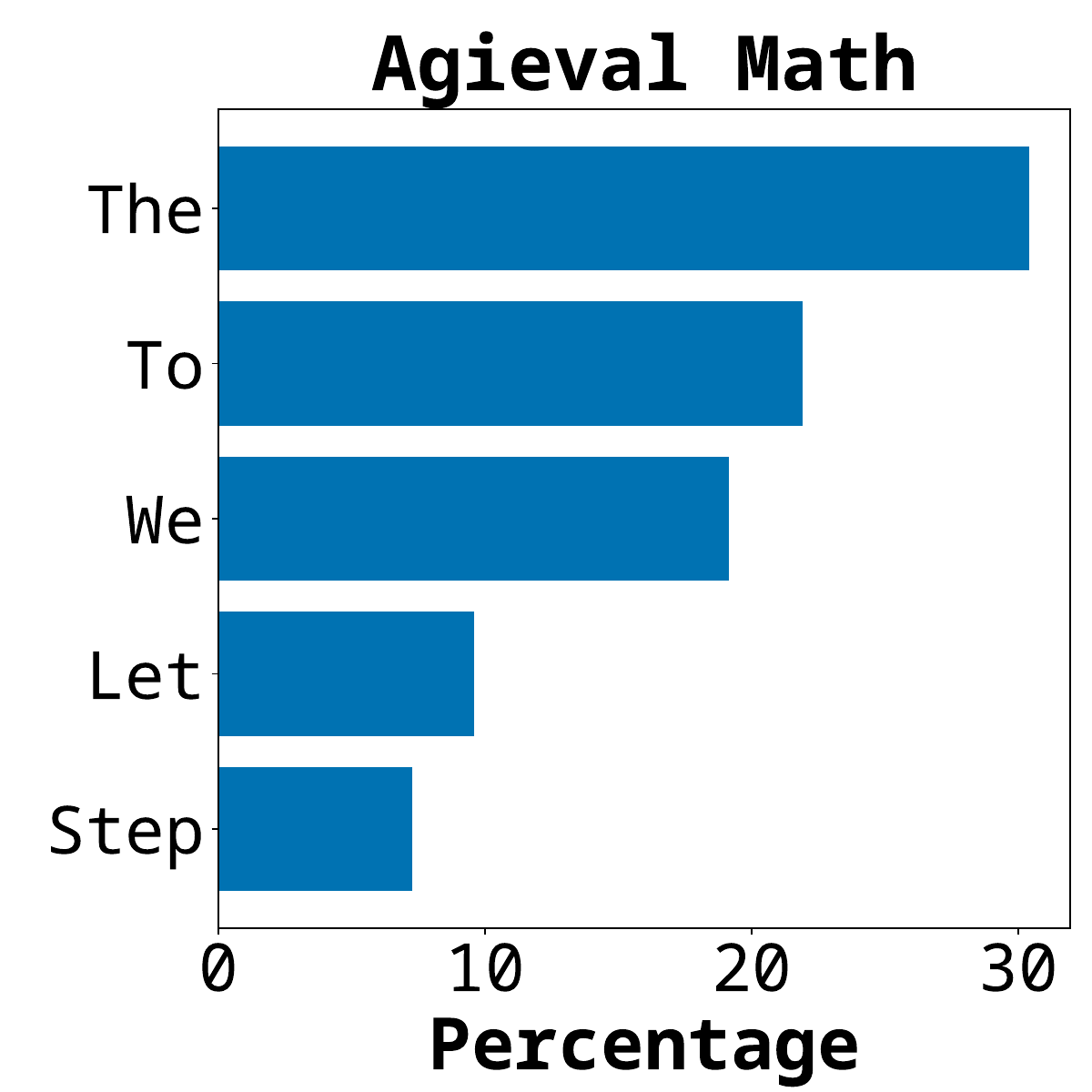}
		\caption{\centering\texttt{Qwen3-4B}}
	\end{subfigure}
	\hfill
	\begin{subfigure}[t]{0.24\linewidth}
		\centering
		\includegraphics[width=\linewidth]{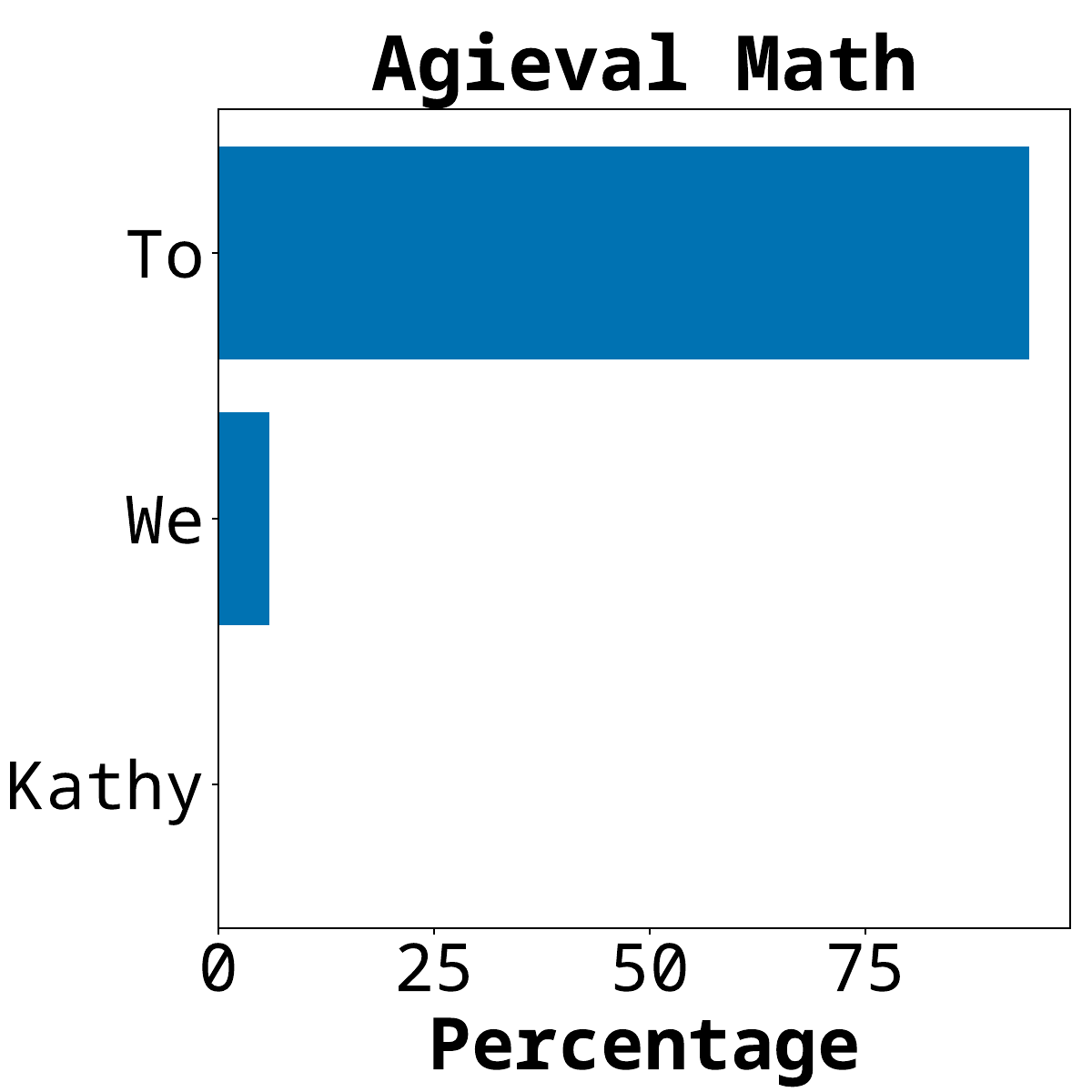}
		\caption{\centering\transfer{\texttt{Qwen3-4B}}{14B}}
	\end{subfigure}
    \hfill
    \begin{subfigure}[t]{0.24\linewidth}
		\centering
		\includegraphics[width=\linewidth]{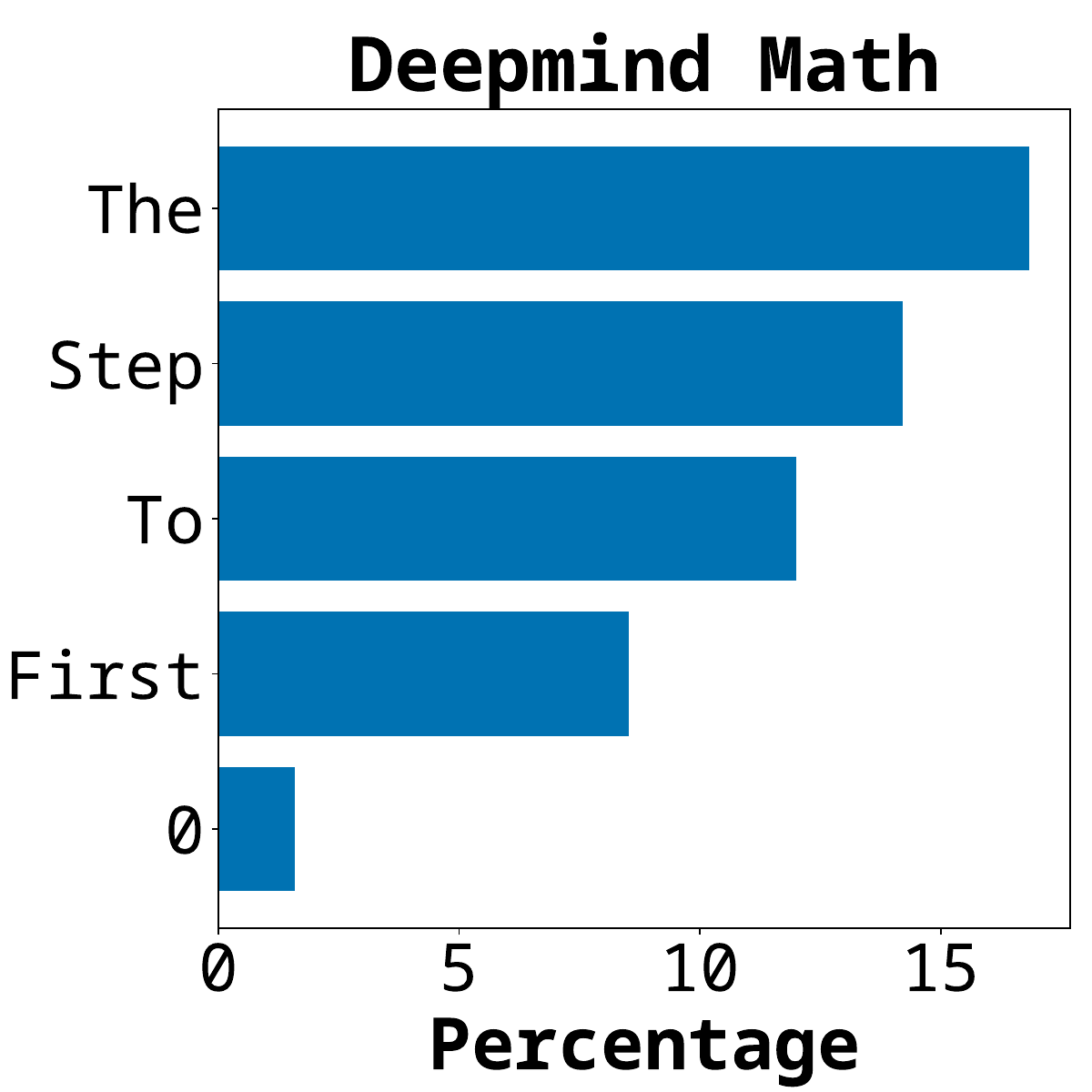}
		\caption{\centering\texttt{Qwen3-4B}}
	\end{subfigure}
	\hfill
	\begin{subfigure}[t]{0.24\linewidth}
		\centering
		\includegraphics[width=\linewidth]{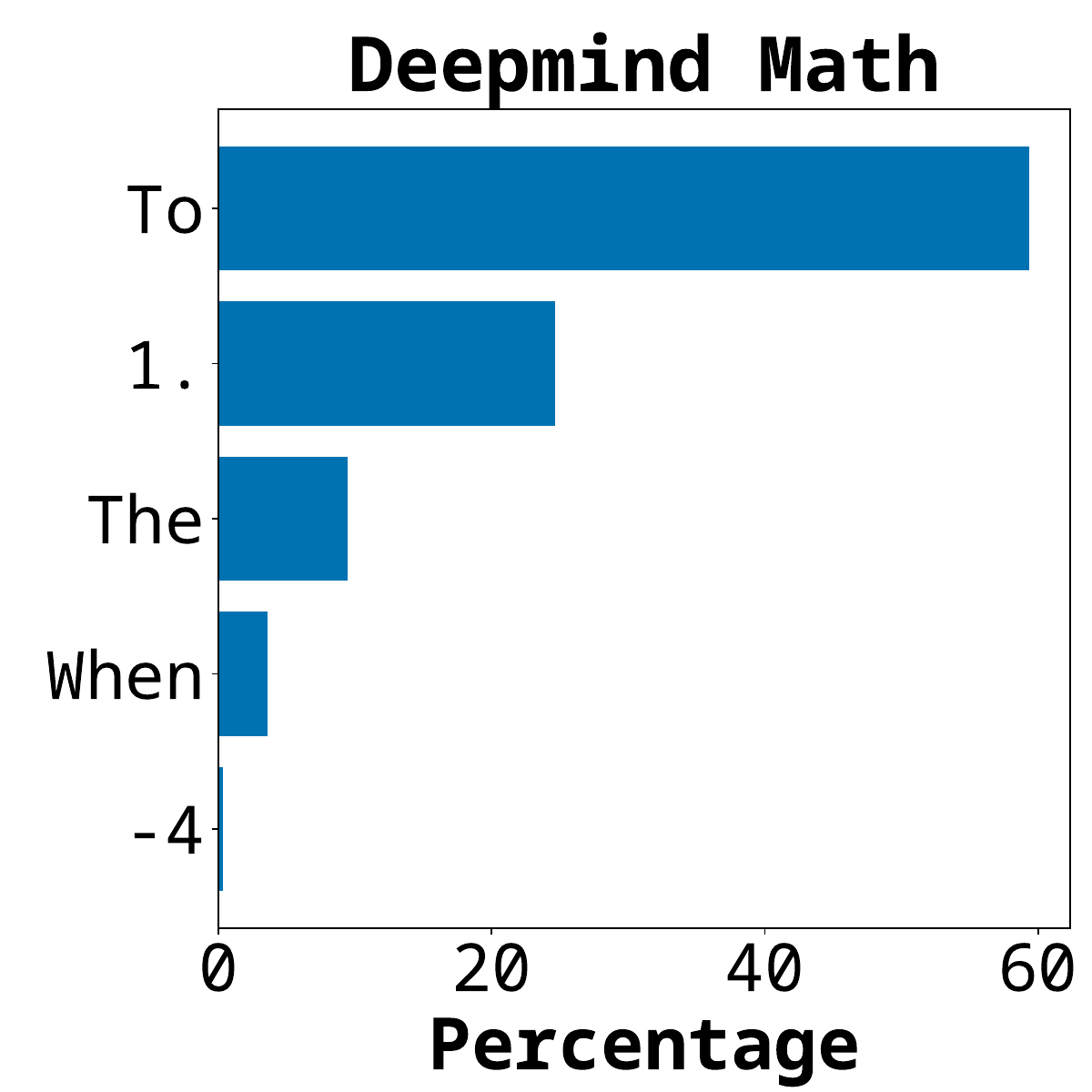}
		\caption{\centering\transfer{\texttt{Qwen3-4B}}{14B}}
	\end{subfigure}
    \\[0.7em]
    \begin{subfigure}[t]{0.24\linewidth}
		\centering
		\includegraphics[width=\linewidth]{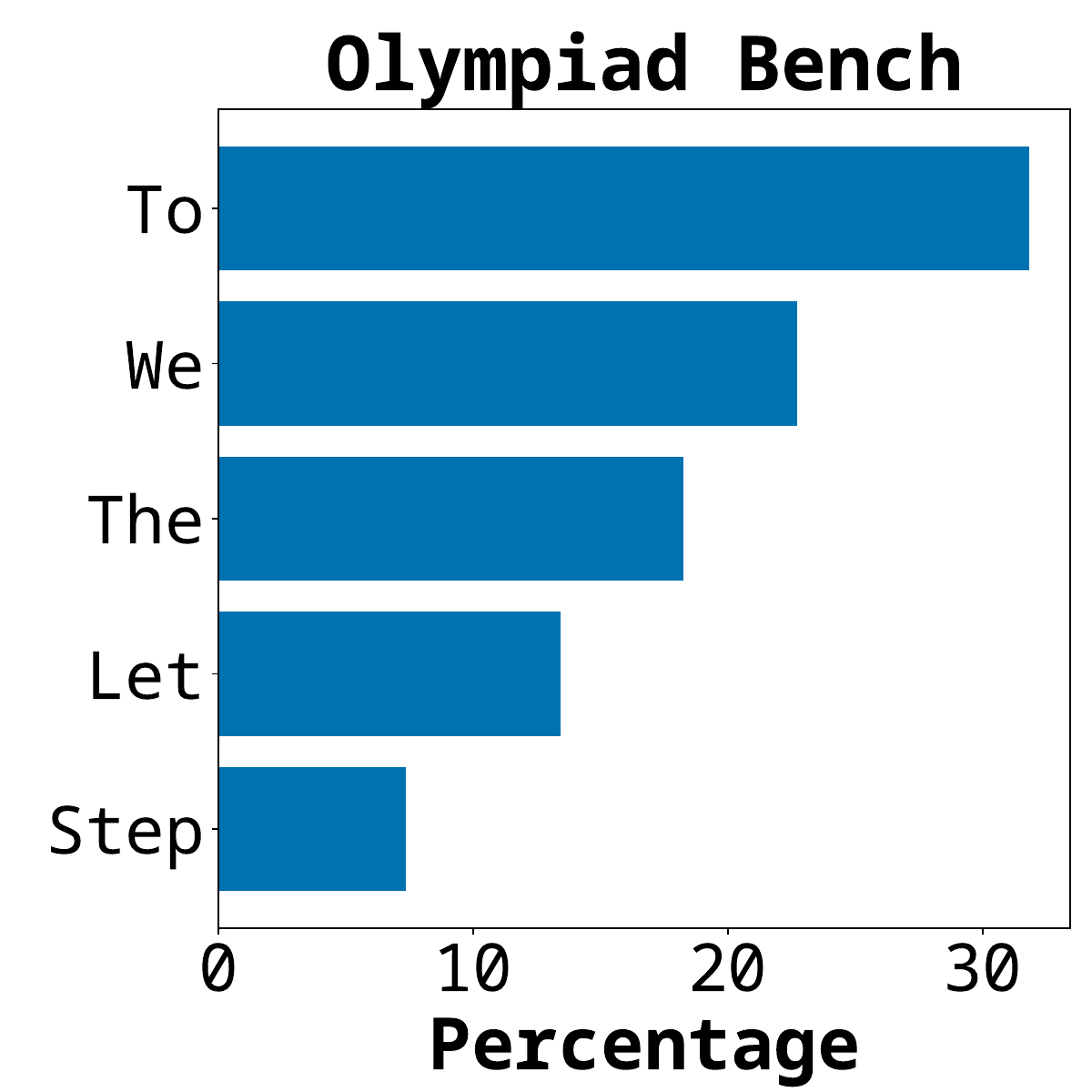}
		\caption{\centering\texttt{Qwen3-4B}}
	\end{subfigure}
	\hfill
	\begin{subfigure}[t]{0.24\linewidth}
		\centering
		\includegraphics[width=\linewidth]{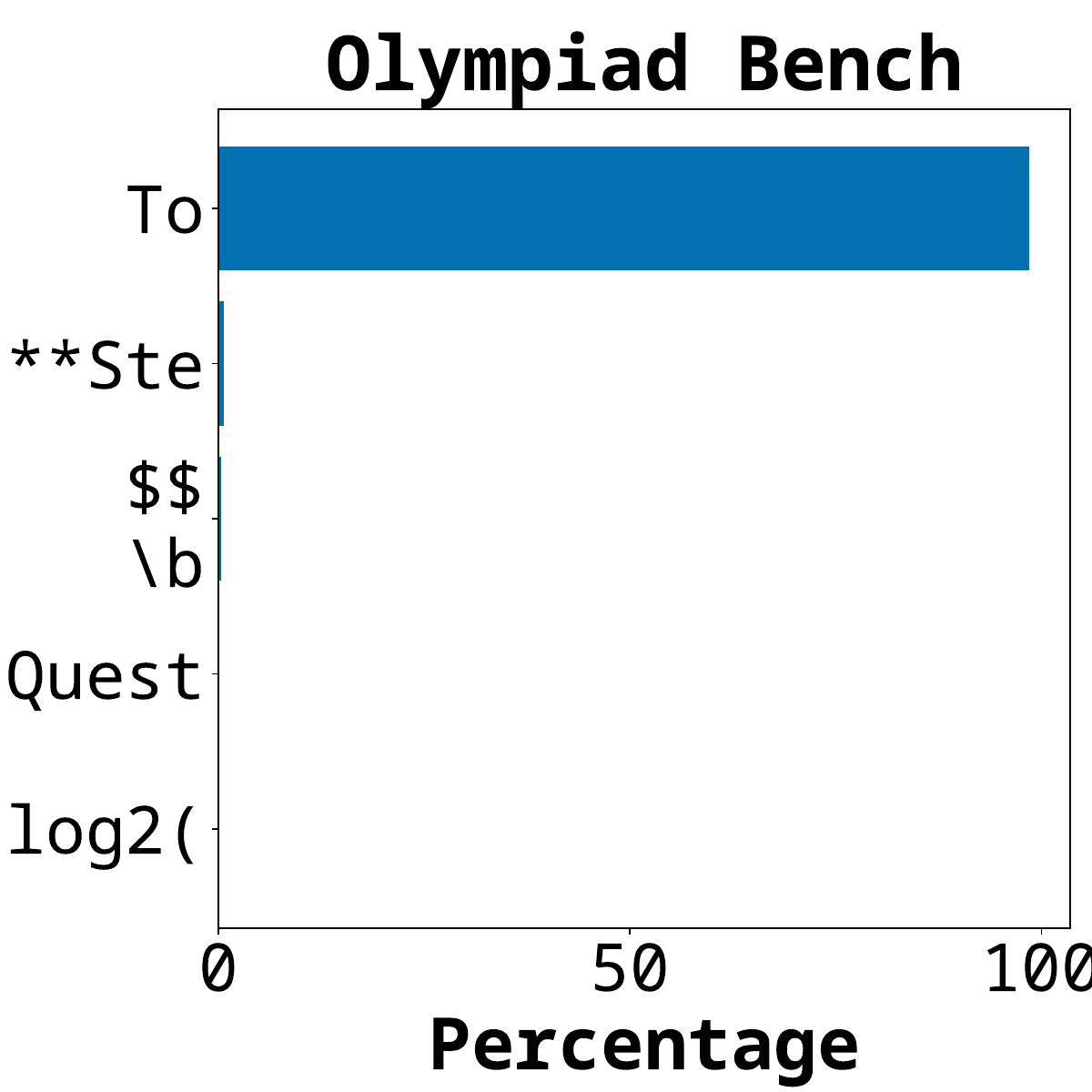}
		\caption{\centering\transfer{\texttt{Qwen3-4B}}{8B}}
	\end{subfigure}
    \begin{subfigure}[t]{0.24\linewidth}
		\centering
		\includegraphics[width=\linewidth]{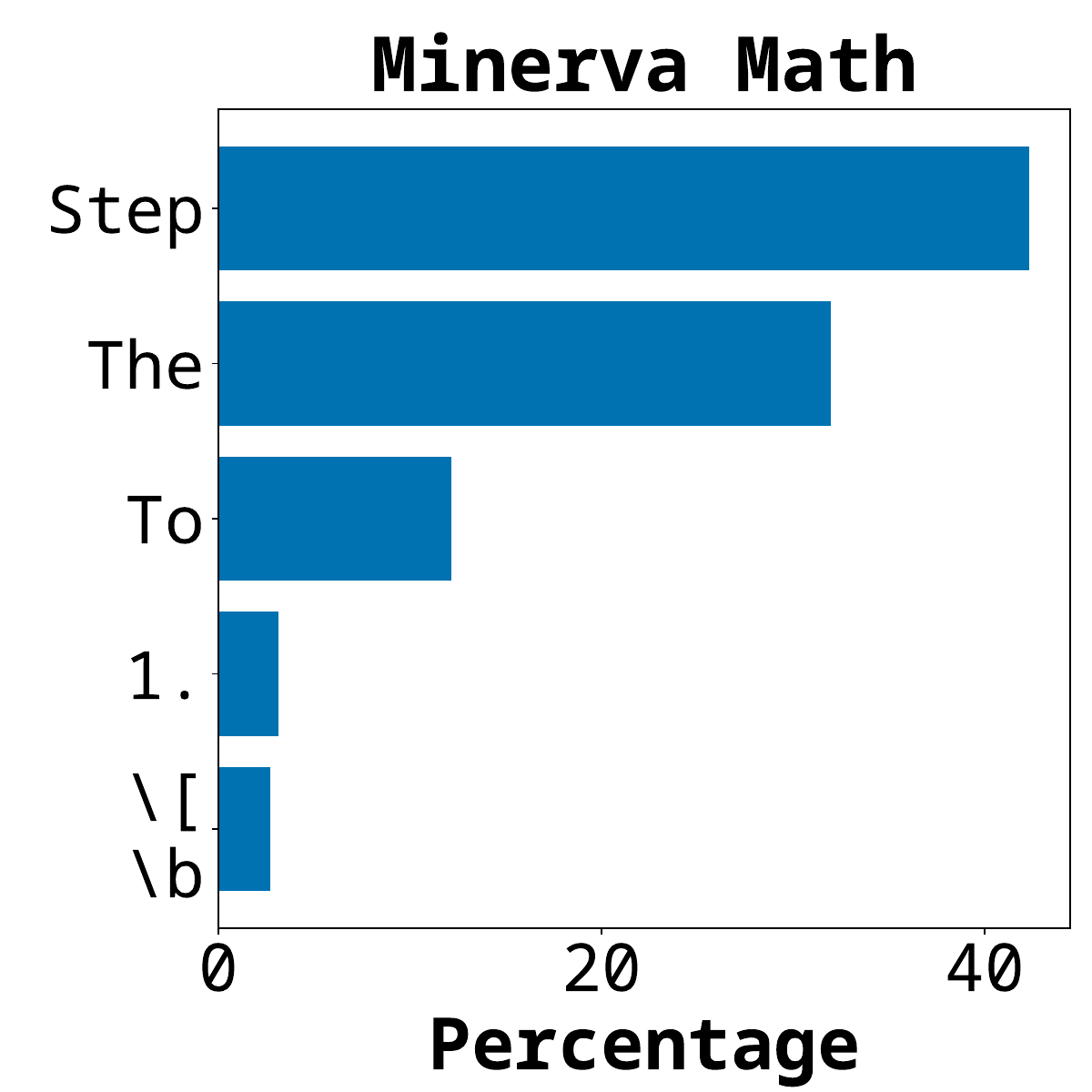}
		\caption{\centering\texttt{Qwen2.5-14B}}
	\end{subfigure}
	\hfill
	\begin{subfigure}[t]{0.24\linewidth}
		\centering
		\includegraphics[width=\linewidth]{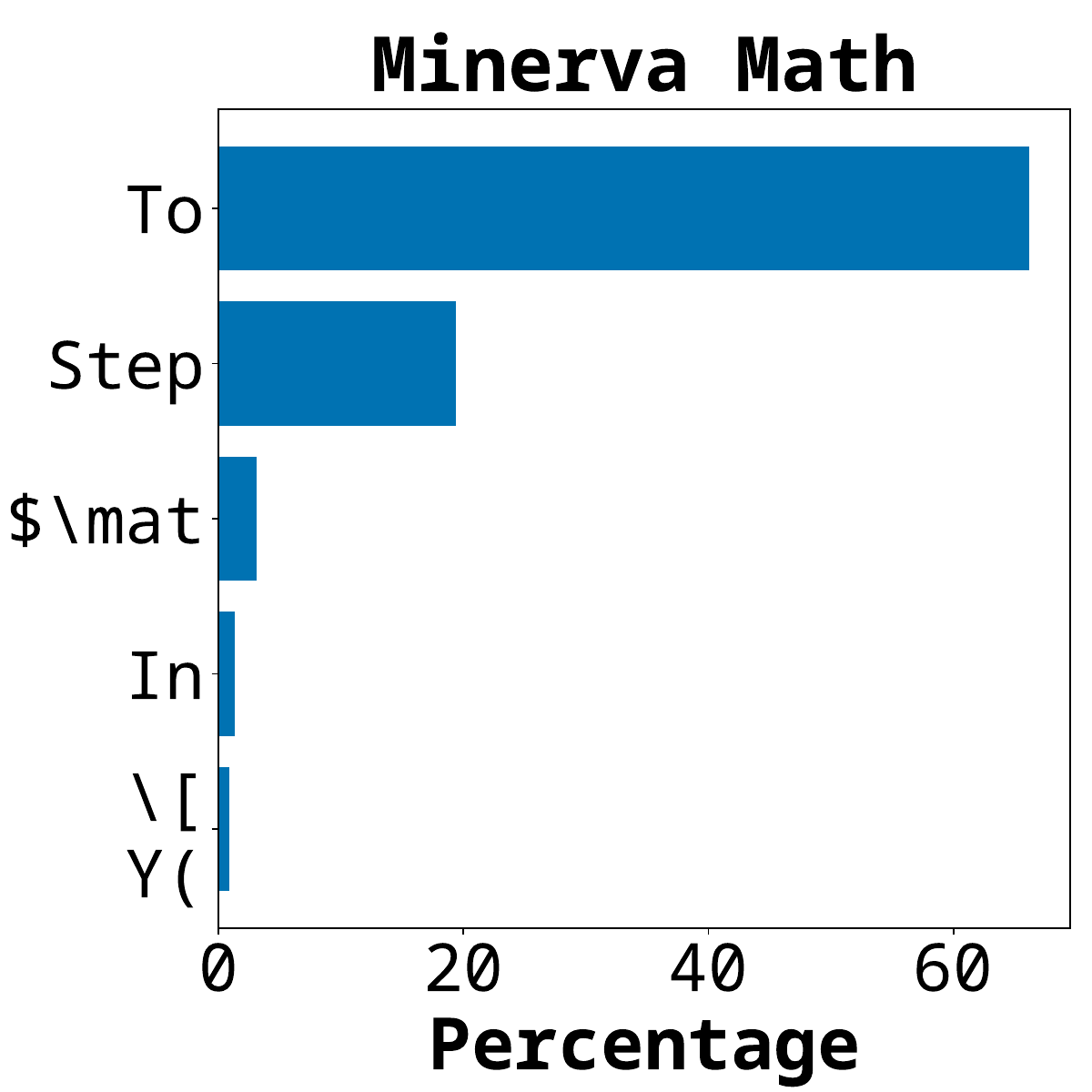}
		\caption{\centering\transfer{\texttt{Qwen2.5-14B}}{7B}}
	\end{subfigure}
    \\[0.7em]
    \begin{subfigure}[t]{0.24\linewidth}
		\centering
		\includegraphics[width=\linewidth]{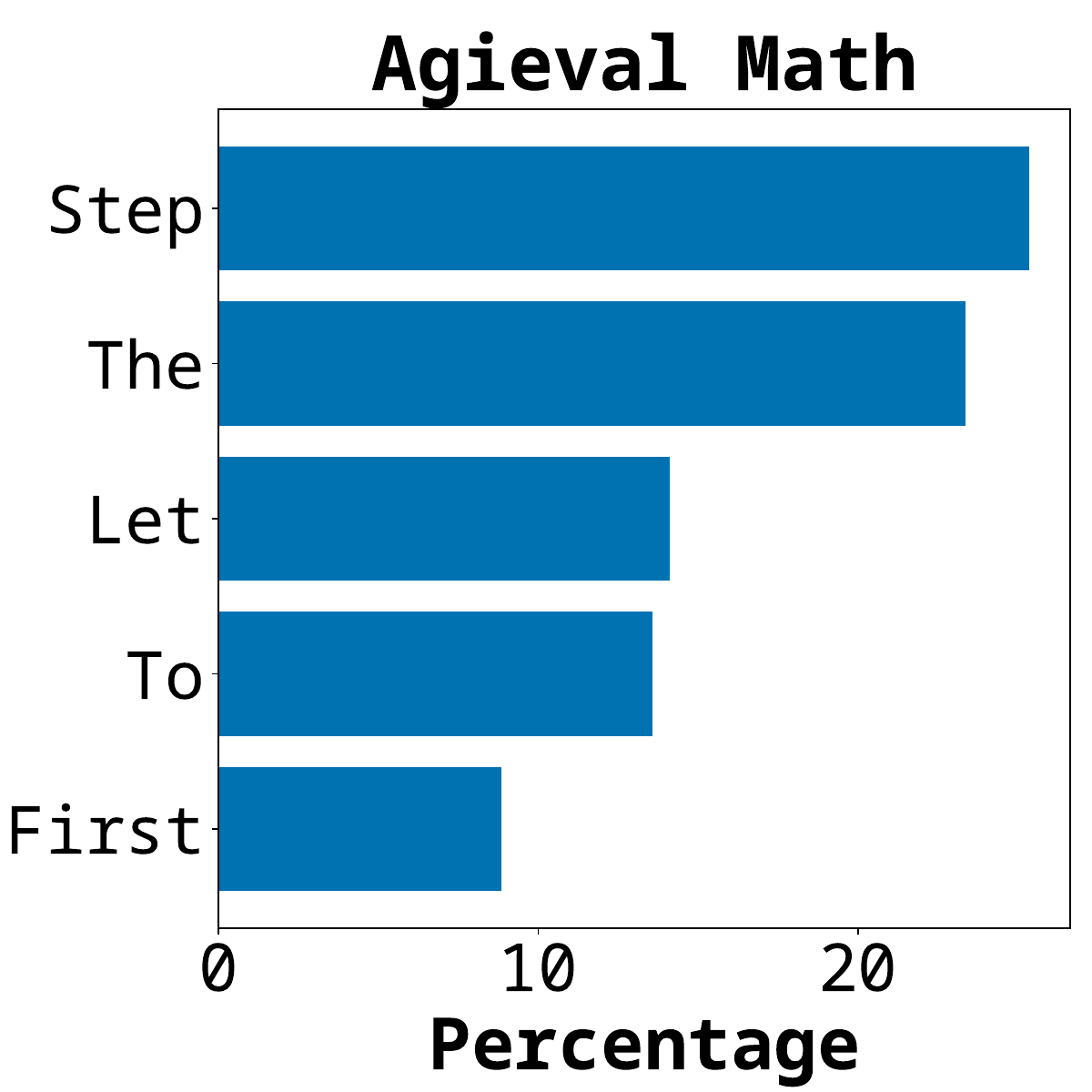}
		\caption{\centering\texttt{Qwen2.5-14B}}
	\end{subfigure}
	\hfill
	\begin{subfigure}[t]{0.24\linewidth}
		\centering
		\includegraphics[width=\linewidth]{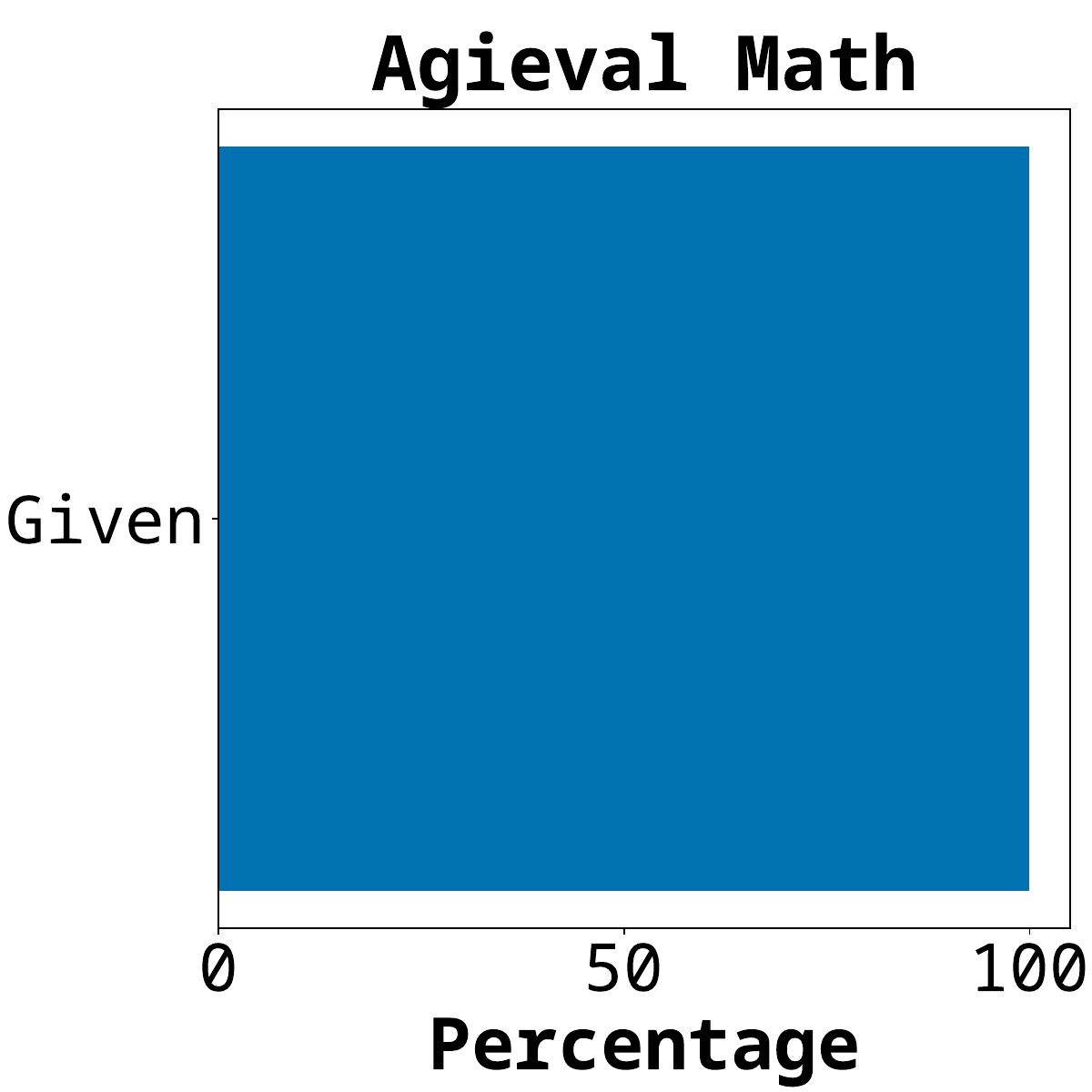}
		\caption{\centering\transfer{\texttt{Qwen2.5-14B}}{7B}}
	\end{subfigure}
    \hfill
    \begin{subfigure}[t]{0.24\linewidth}
		\centering
		\includegraphics[width=\linewidth]{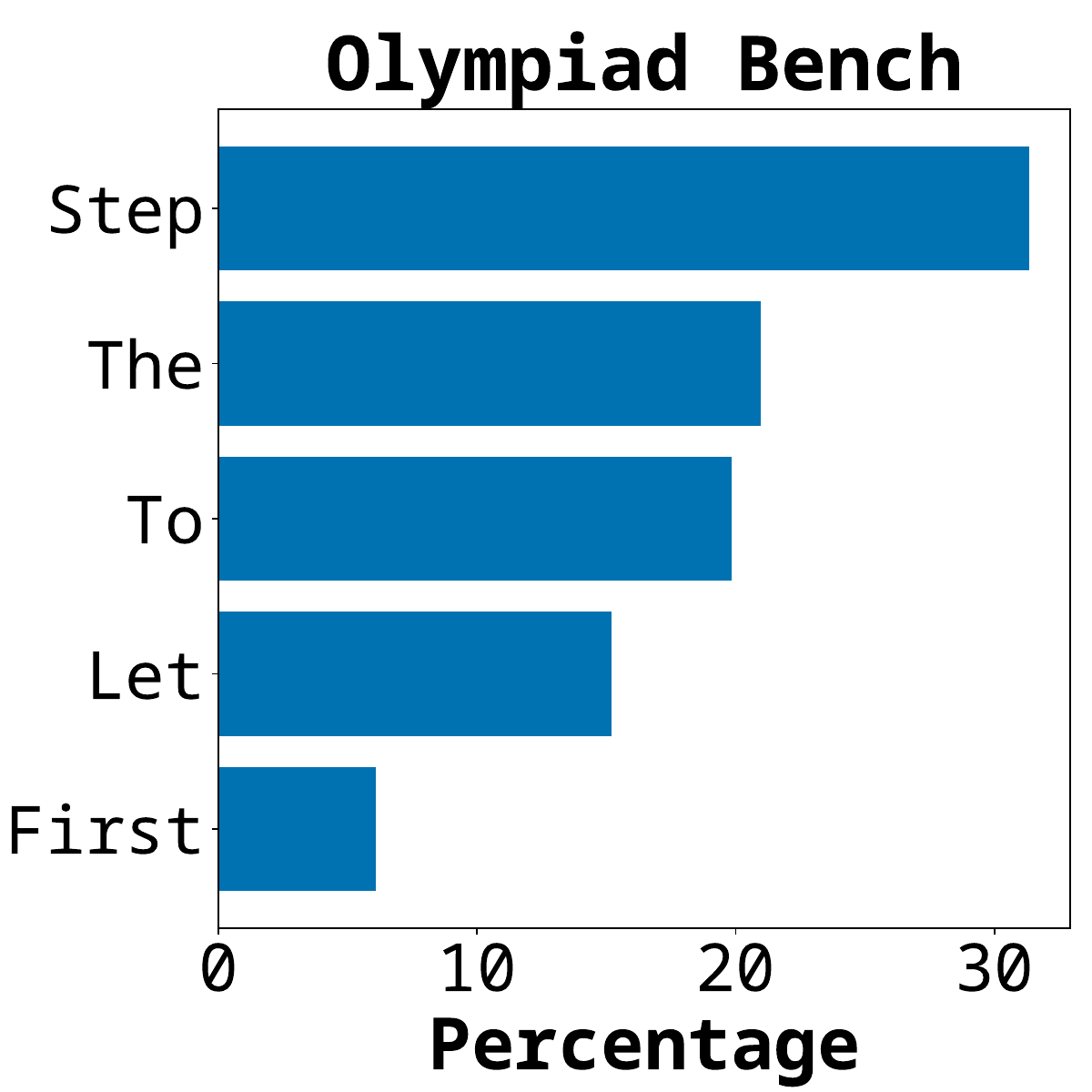}
		\caption{\centering\texttt{Qwen2.5-14B}}
	\end{subfigure}
	\hfill
	\begin{subfigure}[t]{0.24\linewidth}
		\centering
		\includegraphics[width=\linewidth]{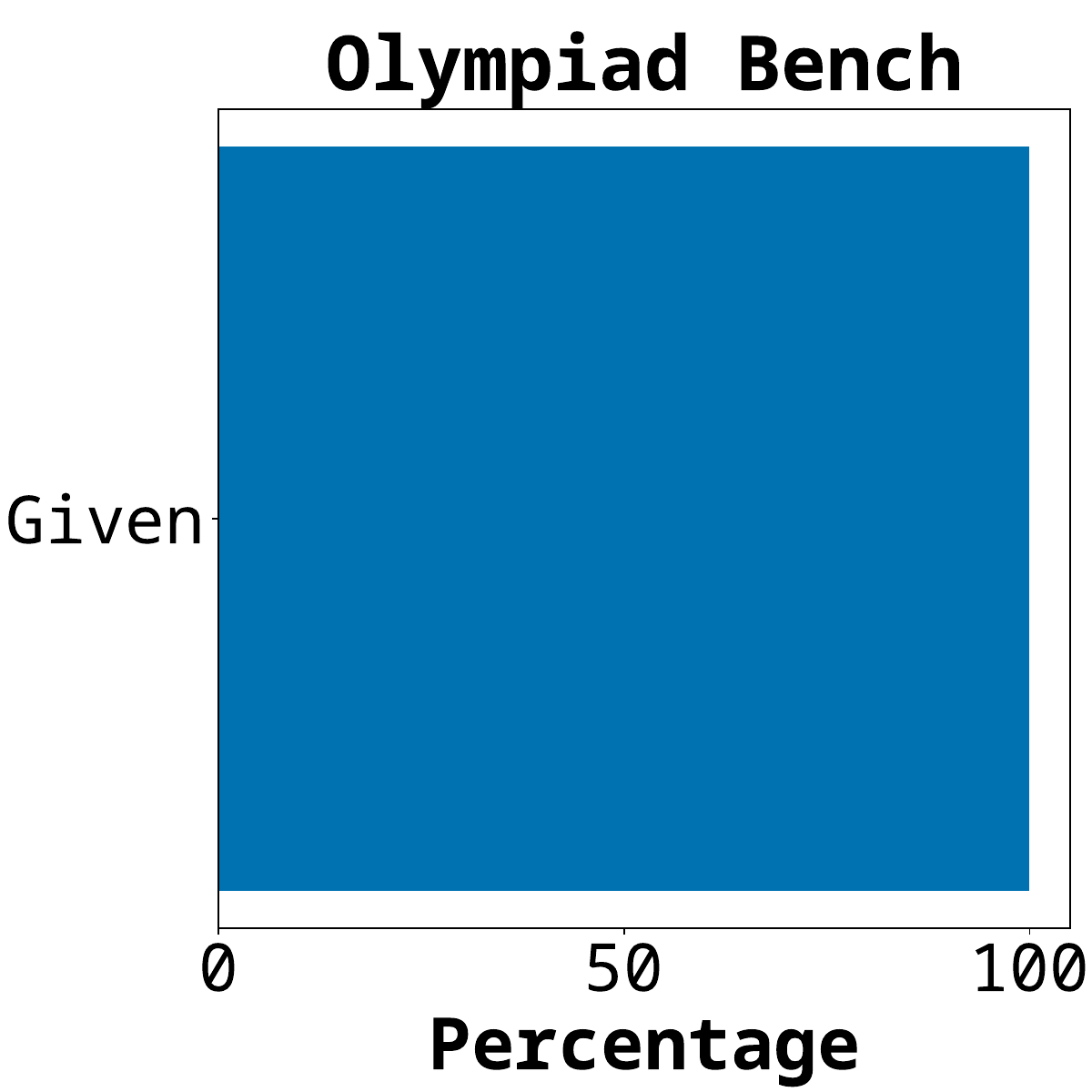}
		\caption{\centering\transfer{\texttt{Qwen2.5-14B}}{7B}}
	\end{subfigure}
    \\[0.7em]
	\caption{
        \textbf{Additional Statistics of the First-generated Word:} A clear shift in the distribution of first-generated word is observed after applying \method
    }
	\label{fig:first-word-dist-appendix}
\end{figure*}

Figure~\ref{fig:first-word-dist-appendix} plots the distribution of first generated words for $\TL$ and $\TU$.
We find that \method sharpens the output distribution toward a small set of recurring openings.
Across model--dataset pairs, the Unlocked model frequently begins with similar phrases (e.g., ``To solve the \dots'' or ``Step 1: \dots'').
In contrast, $\TL$ exhibits a more diffuse distribution over opening tokens.

These patterns suggest that \method increases the likelihood of producing plausible reasoning traces by consolidating representations and reducing variability in early trajectory selection.
We leave a more thorough analysis of diversity and mode coverage under Unlocking, and similarity to various post-training methods to future work.

\emph{(II.) Improved mathematical reliability:}
Example~\ref{ex:math-reasoning} highlights cases where both $\TL$ and $\TU$ generate step-by-step reasoning yet arrive at different conclusions.
$\TL$ often invokes relevant intermediate concepts but fails to reliably build on them to reach a valid solution.
By shifting internal representations during generation, the \masterkey increases the probability that the model follows mathematically sound trajectories.

\begin{figure*}[!hp]
	\centering

	\begin{subfigure}{\linewidth}
		\centering
		\includegraphics[width=\linewidth]{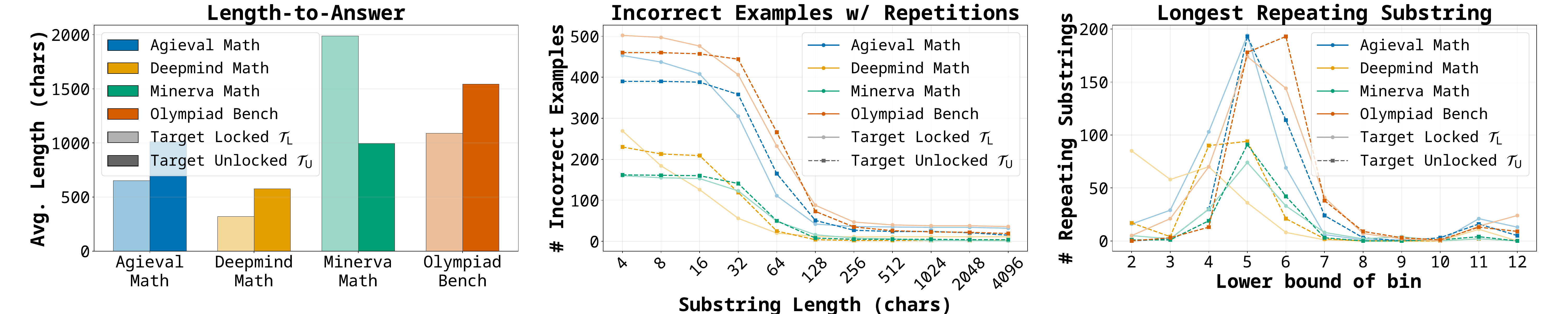}
		\caption{\transfer{\texttt{Qwen3-4B}}{14B}}
	\end{subfigure}
    \\
    \begin{subfigure}{\linewidth}
		\centering
		\includegraphics[width=\linewidth]{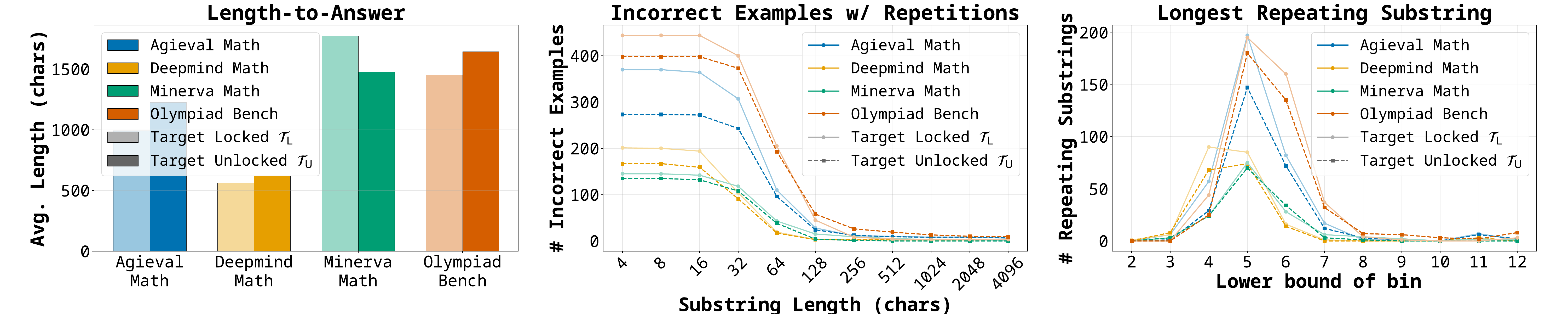}
		\caption{\transfer{\texttt{Qwen3-14B}}{4B}}
	\end{subfigure}
    \\
    \begin{subfigure}{\linewidth}
		\centering
		\includegraphics[width=\linewidth]{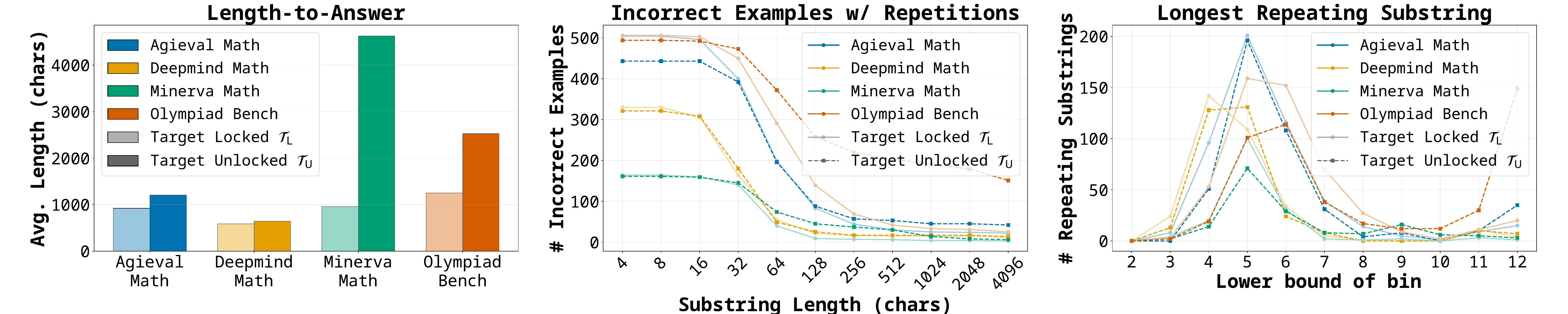}
		\caption{\transfer{\texttt{Ministral-3-3B}}{8B}}
	\end{subfigure}
    \\
    \begin{subfigure}{\linewidth}
		\centering
		\includegraphics[width=\linewidth]{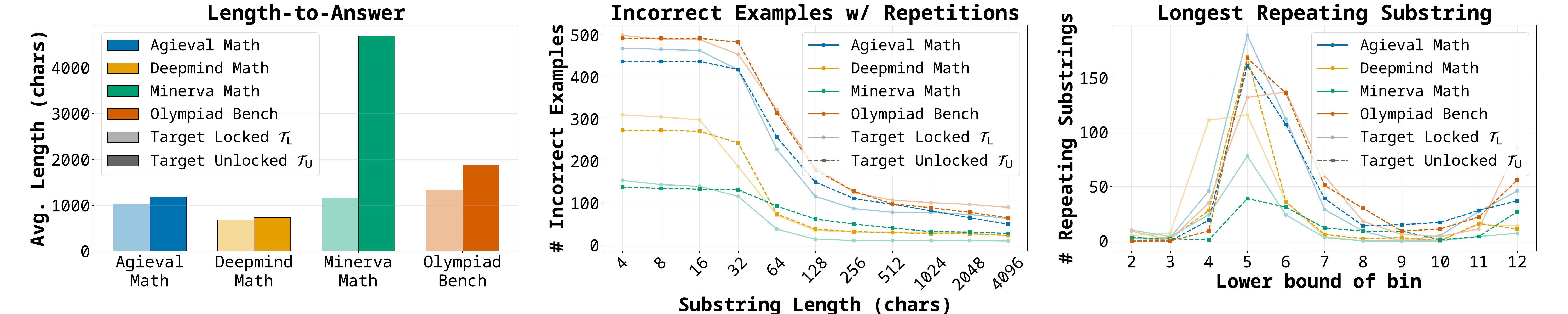}
		\caption{\transfer{\texttt{Ministral-3-8B}}{3B}}
	\end{subfigure}
    \\
	\caption{
        \textbf{Length to answer} (left); and \textbf{Number of repeating substrings} (middle and right) for \texttt{Qwen3} and \texttt{Ministral-3} families.
    }
	\label{fig:substring-analysis}
\end{figure*}

To characterize this effect, we first analyze generation length after unlocking.
Because many models can hallucinate or repeat, we measure length only up to the point at which the final answer is produced, and only for outputs marked correct; we refer to this metric as \emph{length-to-answer}.
Across tasks and model families, \method typically increases length-to-answer (with the exception of Minerva Math), indicating that $\TU$ more often sustains longer, explicit reasoning traces before committing to an answer (Figure~\ref{fig:substring-analysis}, left).

For incorrect solutions, we further quantify degeneration by computing the number of repeated substrings as a function of substring length $l$ (Figure~\ref{fig:substring-analysis}, middle and right).
We find that repetitions peak around $l \in [128,256]$ characters for solutions marked incorrect, indicating substantial repeated fragments in the generated text.
Moreover, $\TL$ exhibits significantly more repetition than $\TU$.
This provides evidence that \method reduces repetition and consolidates the model's internal representations, steering generation more successful reasoning patterns.

\emph{(III.) More consistent formatting:}
A common objective of post-training is to enforce stable output formats so that responses can be parsed and evaluated reliably.
We observe that $\TL$ occasionally deviates from the required format (Example~\ref{ex:math-formatting}), likely because it was not explicitly trained to follow a strict response schema.
In contrast, $\TU$ adheres to the expected format more consistently, reducing format violations.
\emph{We note that these formatting differences are rarely observed for models larger than 7B, suggesting that at this scale the primary gains from \method stem from improved reasoning behavior rather than format compliance.}

\begin{table}
\normalsize
\caption{\textbf{Mathematical Reasoning Transfer From Instruction-Tuned Unlocked Models $\SU$:} Performance of the baselines, \sethlcolor{cyan!15}\hl{task-conditioned}, and \sethlcolor{purple!15}\hl{task-agnostic} transfer.}
\label{tab:math-results-appendix}
\centering

\begin{tabular}{
    >{\centering\arraybackslash}m{1cm}
    >{\centering\arraybackslash}m{2cm}
    >{\centering\arraybackslash}m{2cm}
    >{\centering\arraybackslash}m{2cm}
    ccccccc
}
\toprule
\textbf{Model} & $\TL$ & $\SL$ & $\SU$ & AGI-M & D-M & M-M & OB \\

\midrule

\multirow{15}{*}{\rotatebox[origin=c]{90}{\texttt{Qwen2.5}}}
& 1.5B & -- & --  & 35.9  & 45.3  & 10.8  & 9.9 \\
& 7B & -- & --  & 48.2  & 67.7  & 22.5  & 20.8 \\
& 14B  & -- & --  & 52.2  & 70.7  & 18.5  & 20.0 \\

\cmidrule(lr){2-8}

& 1.5B-Instruct  & -- & --  & 37.8  & 46.6  & 12.6  & 13.3 \\
& 7B-Instruct  & -- & --  & 54.7  & 71.9  & 27.5  & 26.1 \\
& 14B-Instruct  & -- & --  & 65.7  & 78.5  & 29.7  & 33.8 \\

\cmidrule(lr){2-8}

& \cellcolor{cyan!15}1.5B  & \cellcolor{cyan!15}7B & \cellcolor{cyan!15}7B-Instruct & \cellcolor{cyan!15}41.4  & \cellcolor{cyan!15}46.1  & \cellcolor{cyan!15}16.7  & \cellcolor{cyan!15}13.6 \\
& \cellcolor{cyan!15}1.5B  & \cellcolor{cyan!15}14B & \cellcolor{cyan!15}14B-Instruct & \cellcolor{cyan!15}38.6  & \cellcolor{cyan!15}43.5  & \cellcolor{cyan!15}16.2  & \cellcolor{cyan!15}12.3 \\
& \cellcolor{cyan!15}7B  & \cellcolor{cyan!15}1.5B & \cellcolor{cyan!15}1.5B-Instruct & \cellcolor{cyan!15}52.0 & \cellcolor{cyan!15}68.8  & \cellcolor{cyan!15}25.2  & \cellcolor{cyan!15}22.2 \\
& \cellcolor{cyan!15}14B  & \cellcolor{cyan!15}1.5B & \cellcolor{cyan!15}1.5B-Instruct  & \cellcolor{cyan!15}50.3  & \cellcolor{cyan!15}73.4  & \cellcolor{cyan!15}23.9  & \cellcolor{cyan!15}23.0 \\

\cmidrule(lr){2-8}

& \cellcolor{purple!15}1.5B  & \cellcolor{purple!15}7B & \cellcolor{purple!15}7B-Instruct & \cellcolor{purple!15}41.1  & \cellcolor{purple!15}45.7  & \cellcolor{purple!15}18.0  & \cellcolor{purple!15}14.6 \\
& \cellcolor{purple!15}1.5B  & \cellcolor{purple!15}14B & \cellcolor{purple!15}14B-Instruct & \cellcolor{purple!15}38.5  & \cellcolor{purple!15}49.2  & \cellcolor{purple!15}14.9  & \cellcolor{purple!15}12.5 \\
& \cellcolor{purple!15}7B  & \cellcolor{purple!15}1.5B & \cellcolor{purple!15}1.5B-Instruct & \cellcolor{purple!15}50.0 & \cellcolor{purple!15}68.1  & \cellcolor{purple!15}24.3  & \cellcolor{purple!15}21.8 \\
& \cellcolor{purple!15}14B  & \cellcolor{purple!15}1.5B & \cellcolor{purple!15}1.5B-Instruct  & \cellcolor{purple!15}55.5  & \cellcolor{purple!15}72.7  & \cellcolor{purple!15}25.2  & \cellcolor{purple!15}23.5 \\

\midrule

\multirow{15}{*}{\rotatebox[origin=c]{90}{\texttt{Qwen3}}}
& 4B-Base & -- & --  & 52.3  & 71.3  & 27.5  & 19.7 \\
& 8B-Base & -- & --  & 53.6  & 77.1  & 24.3  & 23.0 \\
& 14B-Base  & -- & --  & 61.1  & 78.8  & 34.7  & 29.0 \\

\cmidrule(lr){2-8}

& 4B  & -- & --  & 75.6  & 88.4  & 31.5  & 39.8 \\
& 8B  & -- & --  & 64.0  & 77.6  & 25.2  & 31.4 \\
& 14B  & -- & --  & 67.8  & 80.1  & 27.9  & 37.8 \\

\cmidrule(lr){2-8}

& \cellcolor{cyan!15}4B-Base  & \cellcolor{cyan!15}8B-Base & \cellcolor{cyan!15}8B & \cellcolor{cyan!15}53.1  & \cellcolor{cyan!15}76.4  & \cellcolor{cyan!15}29.3  & \cellcolor{cyan!15}26.6 \\
& \cellcolor{cyan!15}4B-Base  & \cellcolor{cyan!15}14B-Base & \cellcolor{cyan!15}14B & \cellcolor{cyan!15}58.9  & \cellcolor{cyan!15}75.8  & \cellcolor{cyan!15}27.0  & \cellcolor{cyan!15}26.4 \\
& \cellcolor{cyan!15}8B-Base  & \cellcolor{cyan!15}4B-Base & \cellcolor{cyan!15}4B & \cellcolor{cyan!15}54.4  & \cellcolor{cyan!15}73.8  & \cellcolor{cyan!15}26.1  & \cellcolor{cyan!15}20.5 \\
& \cellcolor{cyan!15}14B-Base  & \cellcolor{cyan!15}4B-Base & \cellcolor{cyan!15}4B  & \cellcolor{cyan!15}64.1  & \cellcolor{cyan!15}79.9  & \cellcolor{cyan!15}31.5  & \cellcolor{cyan!15}35.4 \\

\cmidrule(lr){2-8}

& \cellcolor{purple!15}4B-Base  & \cellcolor{purple!15}8B-Base & \cellcolor{purple!15}8B & \cellcolor{purple!15}52.4  & \cellcolor{purple!15}76.5  & \cellcolor{purple!15}28.4  & \cellcolor{purple!15}21.8 \\
& \cellcolor{purple!15}4B-Base  & \cellcolor{purple!15}14B-Base & \cellcolor{purple!15}14B & \cellcolor{purple!15}49.5  & \cellcolor{purple!15}72.9  & \cellcolor{purple!15}25.7  & \cellcolor{purple!15}20.8 \\
& \cellcolor{purple!15}8B-Base  & \cellcolor{purple!15}4B-Base & \cellcolor{purple!15}4B & \cellcolor{purple!15}57.6  & \cellcolor{purple!15}80.9  & \cellcolor{purple!15}27.9  & \cellcolor{purple!15}25.1 \\
& \cellcolor{purple!15}14B-Base  & \cellcolor{purple!15}4B-Base & \cellcolor{purple!15}4B  & \cellcolor{purple!15}71.3  & \cellcolor{purple!15}82.4  & \cellcolor{purple!15}39.2  & \cellcolor{purple!15}36.3 \\

\midrule

\multirow{9}{*}{\rotatebox[origin=c]{90}{\texttt{gemma-3}}}
& 4B-PT & -- & --  & 15.5  & 14.9  & 10.8  & 1.9 \\
& 12B-PT  & -- & --  & 33.1  & 48.4  & 18.9  & 9.1 \\

\cmidrule(lr){2-8}

& 4B-IT  & -- & --  & 62.0  & 74.1  & 17.7  & 29.0 \\
& 12B-IT  & -- & --  & 76.7  & 85.7  & 29.7  & 44.0 \\

\cmidrule(lr){2-8}

& \cellcolor{cyan!15}4B-PT & \cellcolor{cyan!15}12B-PT & \cellcolor{cyan!15}12B-IT & \cellcolor{cyan!15}17.4  & \cellcolor{cyan!15}25.6  & \cellcolor{cyan!15}7.7  & \cellcolor{cyan!15}3.0 \\
& \cellcolor{cyan!15}12B-PT  & \cellcolor{cyan!15}4B-PT & \cellcolor{cyan!15}4B-IT  & \cellcolor{cyan!15}33.5 & \cellcolor{cyan!15}54.7  & \cellcolor{cyan!15}19.4  & \cellcolor{cyan!15}9.6 \\

\cmidrule(lr){2-8}

& \cellcolor{purple!15}4B-PT & \cellcolor{purple!15}12B-PT & \cellcolor{purple!15}12B-IT & \cellcolor{purple!15}16.6  & \cellcolor{purple!15}25.1  & \cellcolor{purple!15}9.0  & \cellcolor{purple!15}3.4 \\
& \cellcolor{purple!15}12B-PT  & \cellcolor{purple!15}4B-PT & \cellcolor{purple!15}4B-IT  & \cellcolor{purple!15}33.7 & \cellcolor{purple!15}53.5  & \cellcolor{purple!15}20.3  & \cellcolor{purple!15}10.1 \\

\midrule

\multirow{9}{*}{\rotatebox[origin=c]{90}{\texttt{Ministral-3}}}
& 3B  & -- & --  & 46.9  & 65.3  & 26.1  & 19.0 \\
& 8B  & -- & --  & 50.7  & 67.4  & 29.3  & 20.0 \\
\cmidrule(lr){2-8}
& $\mathcal{I}$(3B)  & -- & --  & 68.7  & 84.2  & 26.6  & 33.9 \\
& $\mathcal{I}$(8B)  & -- & --  & 70.6  & 87.2  & 29.3  & 37.0 \\
\cmidrule(lr){2-8}
& \cellcolor{cyan!15}3B  & \cellcolor{cyan!15}8B & \cellcolor{cyan!15}8B-Instruct & \cellcolor{cyan!15}53.4  & \cellcolor{cyan!15}66.2  & \cellcolor{cyan!15}27.5  & \cellcolor{cyan!15}21.0 \\
& \cellcolor{cyan!15}8B  & \cellcolor{cyan!15}3B & \cellcolor{cyan!15}3B-Instruct & \cellcolor{cyan!15}51.9  & \cellcolor{cyan!15}71.3  & \cellcolor{cyan!15}37.4  & \cellcolor{cyan!15}20.2 \\
\cmidrule(lr){2-8}
& \cellcolor{purple!15}3B  & \cellcolor{purple!15}8B & \cellcolor{purple!15}8B-Instruct  & \cellcolor{purple!15}49.9  & \cellcolor{purple!15}65.5  & \cellcolor{purple!15}27.5  & \cellcolor{purple!15}21.0 \\
& \cellcolor{purple!15}8B  & \cellcolor{purple!15}3B & \cellcolor{purple!15}3B-Instruct & \cellcolor{purple!15}54.0  & \cellcolor{purple!15}70.7  & \cellcolor{purple!15}34.7  & \cellcolor{purple!15}21.1 \\

\bottomrule
\end{tabular}
\end{table}

\newpage

\begin{table}
\normalsize
\caption{\textbf{Mathematical Reasoning Transfer From Math-specific Unlocked Models $\SU$:} Performance of the baselines, \sethlcolor{yellow!15}\hl{math post-trained models}, \sethlcolor{cyan!15}\hl{task-conditioned}, and \sethlcolor{purple!15}\hl{task-agnostic} transfer.}
\label{tab:math-specific-model-results-appendix}
\centering

\begin{tabular}{
    >{\centering\arraybackslash}m{1cm}
    >{\centering\arraybackslash}m{2cm}
    >{\centering\arraybackslash}m{2cm}
    >{\centering\arraybackslash}m{2cm}
    ccccccc
}
\toprule
\textbf{Model} & $\TL$ & $\SL$ & $\SU$ & AGI-M & D-M & M-M & OB \\

\midrule

\multirow{14}{*}{\rotatebox[origin=c]{90}{\texttt{Qwen2.5}}}
& 7B & -- & --  & 48.2  & 67.7  & 22.5  & 20.8 \\
& 14B  & -- & --  & 52.2  & 70.7  & 18.5  & 20.0 \\

\cmidrule(lr){2-8}

& 7B-Instruct  & -- & --  & 54.7  & 71.9  & 27.5  & 26.1 \\
& 14B-Instruct  & -- & --  & 65.7  & 78.5  & 29.7  & 33.8 \\

\cmidrule(lr){2-8}

& \cellcolor{yellow!15}Nemotron-14B  & \cellcolor{yellow!15}-- & \cellcolor{yellow!15}-- & \cellcolor{yellow!15}58.1  & \cellcolor{yellow!15}82.2  & \cellcolor{yellow!15}10.8  & \cellcolor{yellow!15}9.1 \\

& \cellcolor{yellow!15}DLER-R1-7B  & \cellcolor{yellow!15}-- & \cellcolor{yellow!15}-- & \cellcolor{yellow!15}80.7  & \cellcolor{yellow!15}88.6  & \cellcolor{yellow!15}40.5  & \cellcolor{yellow!15}50.2 \\

\cmidrule(lr){2-8}

& \cellcolor{cyan!15}7B  & \cellcolor{cyan!15}14B & \cellcolor{cyan!15}Nemotron-14B & \cellcolor{cyan!15}52.8  & \cellcolor{cyan!15}71.6  & \cellcolor{cyan!15}21.6  & \cellcolor{cyan!15}20.2 \\
& \cellcolor{cyan!15}14B  & \cellcolor{cyan!15}7B & \cellcolor{cyan!15}DLER-R1-7B & \cellcolor{cyan!15}55.5  & \cellcolor{cyan!15}73.7  & \cellcolor{cyan!15}24.8  & \cellcolor{cyan!15}25.4 \\

\cmidrule(lr){2-8}

& \cellcolor{purple!15}7B  & \cellcolor{purple!15}14B & \cellcolor{purple!15}Nemotron-14B & \cellcolor{purple!15}50.1  & \cellcolor{purple!15}69.9  & \cellcolor{purple!15}23.0  & \cellcolor{purple!15}21.3 \\
& \cellcolor{purple!15}14B  & \cellcolor{purple!15}17B & \cellcolor{purple!15}DLER-R1-7B & \cellcolor{purple!15}58.0  & \cellcolor{purple!15}78.2  & \cellcolor{purple!15}26.1  & \cellcolor{purple!15}26.1 \\

\midrule

\multirow{12}{*}{\rotatebox[origin=c]{90}{\texttt{Qwen3}}}
& 4B-Base & -- & --  & 52.3  & 71.3  & 27.5  & 19.7 \\
& 8B-Base & -- & --  & 53.6  & 77.1  & 24.3  & 23.0 \\

\cmidrule(lr){2-8}

& \cellcolor{yellow!15}Nemotron-Cascade-8B  & \cellcolor{yellow!15}-- & \cellcolor{yellow!15}-- & \cellcolor{yellow!15}80.1  & \cellcolor{yellow!15}89.7 & \cellcolor{yellow!15}36.5  & \cellcolor{yellow!15}45.1 \\

& \cellcolor{yellow!15}4B-Thinking  & \cellcolor{yellow!15}-- & \cellcolor{yellow!15}-- & \cellcolor{yellow!15}60.5  & \cellcolor{yellow!15}75.2  & \cellcolor{yellow!15}26.6  & \cellcolor{yellow!15}36.2 \\

\cmidrule(lr){2-8}

& \cellcolor{cyan!15}4B  & \cellcolor{cyan!15}8B & \cellcolor{cyan!15}Nemotron-Cascade-8B & \cellcolor{cyan!15}56.1  & \cellcolor{cyan!15}78.9  & \cellcolor{cyan!15}27.5  & \cellcolor{cyan!15}24.5 \\
& \cellcolor{cyan!15}8B  & \cellcolor{cyan!15}4B & \cellcolor{cyan!15}4B-Thinking & \cellcolor{cyan!15}55.5  & \cellcolor{cyan!15}80.0  & \cellcolor{cyan!15}28.4  & \cellcolor{cyan!15}22.6 \\

\cmidrule(lr){2-8}

& \cellcolor{purple!15}4B  & \cellcolor{purple!15}8B & \cellcolor{purple!15}Nemotron-Cascade-8B & \cellcolor{purple!15}54.9  & \cellcolor{purple!15}77.2  & \cellcolor{purple!15}27.0  & \cellcolor{purple!15}23.4 \\
& \cellcolor{purple!15}8B  & \cellcolor{purple!15}4B & \cellcolor{purple!15}4B-Thinking & \cellcolor{purple!15}53.6  & \cellcolor{purple!15}79.4  & \cellcolor{purple!15}30.6  & \cellcolor{purple!15}24.3 \\

\bottomrule
\end{tabular}
\end{table}

\onecolumn
\clearpage

\subsection{Examples of Math Reasoning Transfer}
\begin{center}

\begin{tcolorbox}[
	breakable,
    exbox,
	colback=blue!10,
	colframe=blue!100,
	title=\centering\textbf{Question}
]

Find the solution to the equation $\log_{3x} 4 = \log_{2x} 8$, where $x$ is a positive real number other than $\tfrac{1}{3}$ or $\tfrac{1}{2}$.

\end{tcolorbox}

\begin{tcolorbox}[
	breakable,
    exbox,
	colback=red!10,
	colframe=red!70!black!90,
	title=\centering\textbf{Locked Model $\TL$}
]

1. We are given the equation $\log_{3x} 4 = \log_{2x} 8$.\\2. To solve this equation, we can use the change of base formula for logarithms: $\log_a b = \frac{\log_c b}{\log_c a}$, where $c$ is any positive real number other than 1.\\3. Applying the change of base formula to both sides of the equation, we get $\frac{\log 4}{\log (3x)} = \frac{\log 8}{\log (2x)}$.\\4. Now, we can cross-multiply to get rid of the fractions: $\log 4 \cdot \log (2x) = \log 8 \cdot \log (3x)$.\\5. We can simplify the equation by using the properties of logarithms: $\log 4 + \log (2x) = \log 8 + \log (3x)$.\\6. Using the property $\log a + \log b = \log (ab)$, we can rewrite the equation as $\log (4 \cdot 2x) = \log (8 \cdot 3x)$.\\7. Simplifying further, we get $\log (8x) = \log (24x)$.\\8. Since the logarithms are equal, their arguments must also be equal: $8x = 24x$.\\9. Solving for $x$, we get $x = \frac{1}{3}$.\\10. However, we are given that $x$ is a positive real number other than $\tfrac{1}{3}$ or $\tfrac{1}{2}$, so $x = \frac{1}{3}$ is not a valid solution.\\11. Therefore, there is no solution to the equation $\log_{3x} 4 = \log_{2x} 8$.\\\\The final answer is <atok> no solution </atok>.

\end{tcolorbox}

\begin{tcolorbox}[
	breakable,
    exbox,
	colback=green!10,
	colframe=green!40!black!90,
	title=\centering\textbf{Unlocked Model $\TU$}
]

To solve the equation \(\log_{3x} 4 = \log_{2x} 8\), we will follow these steps:\\\\1. **Express the logarithms in terms of natural logarithms:**\\   \[\\   \log_{3x} 4 = \frac{\ln 4}{\ln (3x)} \quad \text{and} \quad \log_{2x} 8 = \frac{\ln 8}{\ln (2x)}\\   \]\\\\2. **Set the two expressions equal to each other:**\\   \[\\   \frac{\ln 4}{\ln (3x)} = \frac{\ln 8}{\ln (2x)}\\   \]\\\\3. **Cross-multiply to eliminate the fractions:**\\   \[\\   \ln 4 \cdot \ln (2x) = \ln 8 \cdot \ln (3x)\\   \]\\\\4. **Use the properties of logarithms to simplify:**\\   \[\\   \ln 4 = 2 \ln 2 \quad \text{and} \quad \ln 8 = 3 \ln 2\\   \]\\   Substituting these into the equation, we get:\\   \[\\   2 \ln 2 \cdot \ln (2x) = 3 \ln 2 \cdot \ln (3x)\\   \]\\\\5. **Divide both sides by \(\ln 2\) (since \(\ln 2 = 0\)):**\\   \[\\   2 \ln (2x) = 3 \ln (3x)\\   \]\\\\6. **Expand the logarithms:**\\   \[\\   2 (\ln 2 + \ln x) = 3 (\ln 3 + \ln x)\\   \]\\\\7. **Distribute the constants:**\\   \[\\   2 \ln 2 + 2 \ln x = 3 \ln 3 + 3 \ln x\\   \]\\\\8. **Isolate the terms involving \(\ln x\):**\\   \[\\   2 \ln 2 - 3 \ln 3 = 3 \ln x - 2 \ln x\\   \]\\   \[\\   2 \ln 2 - 3 \ln 3 = \ln x\\   \]\\\\9. **Solve for \(x\):**\\   \[\\   \ln x = 2 \ln 2 - 3 \ln 3\\   \]\\   \[\\   \ln x = \ln 2^2 - \ln 3^3\\   \]\\   \[\\   \ln x = \ln 4 - \ln 27\\   \]\\   \[\\   \ln x = \ln \left(\frac{4}{27}\right)\\   \]\\   \[\\   x = \frac{4}{27}\\   \]\\\\The final answer is \(\boxed{\frac{4}{27}}\).

\end{tcolorbox}

\captionof{example}{\textbf{Improved Coherence in Reasoning Traces:} $\TL$ and $\TU$ generate similar reasoning steps up till Step 4 in $\TL$'s output and Step 3 in $\TU$'s output. At this point $\TL$ skips intermediate reasoning steps, and thus makes mistakes in all further steps. On the other hand $\TU$ continues to reason step by step and arrives at the correct answer}
\label{ex:math-consistent-cot}

\end{center}
\clearpage

\begin{center}

\begin{tcolorbox}[
	breakable,
    exbox,
	colback=blue!10,
	colframe=blue!100,
	title=\centering\textbf{Question}
]
What is 1134508901 to the power of 1/4, to the nearest integer?
\end{tcolorbox}

\begin{tcolorbox}[
	breakable,
    exbox,
	colback=red!10,
	colframe=red!70!black!90,
	title=\centering\textbf{Locked Model $\TL$}
]

To find the fourth root of 1134508901 to the nearest integer, we can follow these steps:\\\\1. **Understand the problem**: We need to find a number \( x \) such that \( x^4 \) is approximately equal to 1134508901. In other words, we are looking for \( x = \sqrt[4]{1134508901} \).
\\\\2. **Estimate the value**: Since \( 10^4 = 10000 \) and \( 100^4 = 100000000 \), the fourth root of 1134508901 must be between 100 and 1000. Let's narrow it down further.\\\\3. **Calculate \( 300^4 \)**:\\   \[\\   300^4 = (300^2)^2 = 90000^2 = 8100000000\\   \]\\   This is much larger than 1134508901, so the fourth root must be less than 300.\\\\4. **Calculate \( 200^4 \)**:\\   \[\\   200^4 = (200^2)^2 = 40000^2 = 1600000000\\   \]\\   This is still larger than 1134508901, so the fourth root must be less than 200.\\\\5. **Calculate \( 150^4 \)**:\\   \[\\   150^4 = (150^2)^2 = 22500^2 = 506250000\\   \]\\   This is less than 1134508901, so the fourth root must be greater than 150.
\\\\6. **Calculate \( 180^4 \)**:\\   \[\\   180^4 = (180^2)^2 = 32400^2 = 1049760000\\   \]\\   This is close to 1134508901, so the fourth root is likely around 180.\\\\7. **Calculate \( 185^4 \)**:\\   \[\\   185^4 = (185^2)^2 = 34225^2 = 1171215625\\   \]\\   This is slightly larger than 1134508901, so the fourth root must be less than 185.\\\\8. **Calculate \( 183^4 \)**:\\   \[\\   183^4 = (183^2)^2 = 33489^2 = 1121088121\\   \]\\   This is very close to 1134508901, so the fourth root is likely around 183.\\\\9. **Calculate \( 184^4 \)**:\\   \[\\   184^4 = (184^2)^2 = 33856^2 = 1146617616\\   \]\\   This is slightly larger than 1134508901, so the fourth root must be less than 184.
\\\\10. **Conclusion**: The fourth root of 1134508901 is between 183 and 184. Since \(183^4\) is closer to 1134508901 than \(184^4\), the nearest integer is 183.\\\\The final answer is <atok> 183 </atok>.

\end{tcolorbox}

\begin{tcolorbox}[
	breakable,
    exbox,
	colback=green!10,
	colframe=green!40!black!90,
	title=\centering\textbf{Unlocked Model $\TU$}
]

1. First, we need to find the fourth root of 1134508901.\\2. To do this, we can use a calculator or a computer program that can handle large numbers and roots.\\3. Using a calculator, we find that the fourth root of 1134508901 is approximately 183.57.\\4. Now, we need to round this number to the nearest integer.\\5. Since 183.57 is closer to 184 than it is to 183, we round up to 184.\\\\The final answer is <atok> 184 </atok>.

\end{tcolorbox}

\captionof{example}{\textbf{Improved Coherence in Reasoning Traces:} $\TL$ arrives at approximately the correct answer by making a mistake in floating point estimations. On the other hand, $\TU$ is more vary of the rounding off specified in the question, and thus reaches the correct solution.}
\label{ex:math-correct-reasoning}

\end{center}
\clearpage

\begin{center}

\begin{tcolorbox}[
	breakable,
    exbox,
	colback=blue!10,
	colframe=blue!100,
	title=\centering\textbf{Question}
]

In the $n^{th}$ row of Pascal's Triangle where the first row is $n=0$, the arithmetic mean of the elements is $51.2$. What is the value of $n$?

\end{tcolorbox}

\begin{tcolorbox}[
	breakable,
    exbox,
	colback=red!10,
	colframe=red!70!black!90,
	title=\centering\textbf{Locked Model $\TL$}
]

1. We know that the sum of the elements in the $n^{th}$ row of Pascal's Triangle is $2^n$.\\2. The arithmetic mean of the elements in the $n^{th}$ row is the sum of the elements divided by the number of elements, which is $2^n / (n+1)$.\\3. We are given that the arithmetic mean is $51.2$, so we can set up the equation: $2^n / (n+1) = 51.2$.\\4. To solve for $n$, we can multiply both sides of the equation by $(n+1)$: $2^n = 51.2(n+1)$.\\5. Now, we can try different values of $n$ to find the one that satisfies the equation. We can start with $n=10$ since $2^{10} = 1024$ and $51.2(10+1) = 563.2$, which is close to $1024$.\\6. We can check if $n=10$ is the correct value by plugging it back into the equation: $2^{10} / (10+1) = 1024 / 11 = 93.09$, which is not equal to $51.2$.\\7. We can try $n=11$ since $2^{11} = 2048$ and $51.2(11+1) = 614.4$, which is closer to $2048$.\\8. We can check if $n=11$ is the correct value by plugging it back into the equation: $2^{11} / (11+1) = 2048 / 12 = 170.67$, which is not equal to $51.2$.\\9. We can try $n=12$ since $2^{12} = 4096$ and $51.2(12+1) = 665.6$, which is closer to $4096$.\\10. We can check if $n=12$ is the correct value by plugging it back into the equation: $2^{12} / (12+1) = 4096 / 13 = 315.08$, which is not equal to $51.2$.\\ $\cdots$

\end{tcolorbox}

\begin{tcolorbox}[
	breakable,
    exbox,
	colback=green!10,
	colframe=green!40!black!90,
	title=\centering\textbf{Unlocked Model $\TU$}
]

To solve this problem, we need to understand the properties of Pascal's Triangle and how to calculate the arithmetic mean of its elements in the $n^{th}$ row.
\\\\\#\#\# Step 1: Understand Pascal's Triangle\\Pascal's Triangle is a triangular array of numbers where each number is the sum of the two directly above it. The $n^{th}$ row of Pascal's Triangle corresponds to the coefficients of the binomial expansion $(a + b)^n$. The elements in the $n^{th}$ row are given by the binomial coefficients $\binom{n}{k}$ for $k = 0, 1, 2, \dots, n$\\\\\#\#\# Step 2: Calculate the Sum of the Elements in the $n^{th}$ Row\\The sum of the elements in the $n^{th}$ row of Pascal's Triangle is $2^n$. This is because the sum of the binomial coefficients $\binom{n}{k}$ for $k = 0$ to $n$ is equal to $2^n$.\\\\\#\#\# Step 3: Calculate the Arithmetic Mean of the Elements in the $n^{th}$ Row\\The arithmetic mean of the elements in the $n^{th}$ row is the sum of the elements divided by the number of elements. Since there are $n + 1$ elements in the $n^{th}$ row, the arithmetic mean is:\\\[\\\text{Arithmetic Mean} = \frac{2^n}{n + 1}\\\]\\\\\#\#\# Step 4: Set Up the Equation\\We are given that the arithmetic mean is $51.2$. Therefore, we can set up the equation:\\\[\\\frac{2^n}{n + 1} = 51.2\\\]\\\\\#\#\# Step 5: Solve for $n$\\We need to find the value of $n$ that satisfies the equation. Let's test integer values of $n$:\\\\- For $n = 6$:\\  \[\\  \frac{2^6}{6 + 1} = \frac{64}{7} \approx 9.14 \neq 51.2\\  \]\\  \\- For $n = 7$:\\  \[\\  \frac{2^7}{7 + 1} = \frac{128}{8} = 16 \neq 51.2\\  \]\\  \\- For $n = 8$:\\  \[\\  \frac{2^8}{8 + 1} = \frac{256}{9} \approx 28.44 \neq 51.2\\  \]\\  \\- For $n = 9$:\\  \[\\  \frac{2^9}{9 + 1} = \frac{512}{10} = 51.2\\  \]\\\\\#\#\# Step 6: Conclusion\\The value of $n$ that satisfies the equation is $9$.\\\\The final answer is <atok> 9 </atok>.

\end{tcolorbox}

\captionof{example}{\textbf{Improved Math Reasoning Capability:} $\TL$ completely hallucinates, and never reaches the solution, while $\TU$ is able to generate complete reasoning and the correct solution.}
\label{ex:math-reasoning}

\end{center}
\clearpage

\begin{center}

\begin{tcolorbox}[
	breakable,
    exbox,
	colback=blue!10,
	colframe=blue!100,
	title=\centering\textbf{Question}
]
Compute $\begin{pmatrix} 1 & 2 \\ 4 & 8 \\ \end{pmatrix} \begin{pmatrix} 5 \\ 3 \end{pmatrix}.$
\end{tcolorbox}

\begin{tcolorbox}[
	breakable,
    exbox,
	colback=red!10,
	colframe=red!70!black!90,
	title=\centering\textbf{Locked Model $\TL$}
]

To compute the product of the given matrices, we need to follow the rules of matrix multiplication. The first matrix is a $2 \times 2$ matrix, and the second matrix is a $2 \times 1$ matrix. The result will be a $2 \times 1$ matrix.

Step 1: Multiply the first row of the first matrix by the column of the second matrix.  
\[
(1 \times 5) + (2 \times 3) = 5 + 6 = 11
\]

Step 2: Multiply the second row of the first matrix by the column of the second matrix.  
\[
(4 \times 5) + (8 \times 3) = 20 + 24 = 44
\]

The final answer is \textless atok\textgreater\ $[11, 44]$\ \textless/atok\textgreater.

\end{tcolorbox}

\begin{tcolorbox}[
	breakable,
    exbox,
	colback=green!10,
	colframe=green!40!black!90,
	title=\centering\textbf{Unlocked Model $\TU$}
]

Step 1: Identify the matrices involved in the multiplication.

Matrix A = 
\[
\begin{pmatrix}
1 & 2 \\
4 & 8 \\
\end{pmatrix}
\]

Matrix B = 
\[
\begin{pmatrix}
5 \\
3
\end{pmatrix}
\]

Step 2: Check if the multiplication is possible.
Matrix A is a $2 \times 2$ matrix, and Matrix B is a $2 \times 1$ matrix. Since the number of columns in Matrix A is equal to the number of rows in Matrix B, the multiplication is possible.

Step 3: Perform the matrix multiplication.
To multiply the matrices, we need to multiply the elements of each row in Matrix A by the corresponding elements in Matrix B and then sum the results.

Row 1:
\[
(1 \times 5) + (2 \times 3) = 5 + 6 = 11
\]

Row 2:
\[
(4 \times 5) + (8 \times 3) = 20 + 24 = 44
\]

Step 4: Write the result as a matrix.

The final answer is \textless atok\textgreater\ 
\[
\begin{pmatrix}
11 \\
44
\end{pmatrix}
\]
\ \textless/atok\textgreater.

\end{tcolorbox}

\captionof{example}{\textbf{Improved Formatting:} Both $\TL$ and $\TU$ achieve the same answer but $\TL$ incorrectly formats the final answer. This phenomenon is rarely observed in 14B models (Example taken from Qwen-2.5-7B)}
\label{ex:math-formatting}

\end{center}

\clearpage

\section{Model Family Transfer}
\label{sec:model-family}

\subsection{Experimental Setup}

We now investigate the efficacy of cross-family transfer, where the \emph{Source} and \emph{Target} models belong to different architectural families. 
From Section~\ref{sec:cot-main}, we observed that \texttt{Qwen-1.5} family of models possesses CoT as an atomic ability, and consistently demonstrates robust performance across the evaluation settings.
We thus select \texttt{Qwen1.5} checkpoints as the \emph{Source} models. In contrast to the intra-family configuration used in the CoT setting, we utilize the stronger post-trained variant (\texttt{-Chat}) model for $\SU$, while maintaining all other experimental settings.

\subsection{Results \& Observations}

We report our results in Table~\ref{tab:model-family-results}. 
First, we observe that \emph{cross-family transfer} can elicit significant CoT behavior from $\TL$, confirming that Chain-of-Thought capabilities are often latent.
Next, we find that cross-family transfer achieves comparable performance to prompting $\TL$ with a \emph{CoT} prompt.
Surprisingly, we find that cross-family transfer performs comparably to intra-family transfer, providing evidence of converging representations of capabilities across models.
These results further support our hypothesis that if the Target model contains sufficient representational capacity, it is possible to isolate and apply capabilities and directions in latent space.
We leave further exploration into this space, and the more complex challenge of non-atomic capability transfer to future work.

\begin{table}
\normalsize
\caption{\textbf{Cross-Family CoT Transfer:} Results when transferring Chain-of-Thought capabilities from the \texttt{Qwen1.5} family onto \texttt{gemma-2} and \texttt{OLMo-2} families.
For simplicity, we use $\mathcal{I}(x)$ to denote the instruction-tuned version of the corresponding model.
\sethlcolor{gray!15}\hl{Baseline comparisons shown in gray}.}
\label{tab:model-family-results}
\centering
\begin{tabular}{
    >{\centering\arraybackslash}m{1.2cm}
    >{\centering\arraybackslash}m{1.6cm}
    ccccc
}
\toprule
\textbf{Model} & \textbf{Prompt} & $\TL$ & $\SU \equiv \mathcal{I}(\SL)$ & \makecell{GSM8K} & \makecell{MATH} & \makecell{SVAMP} \\
\midrule

\multirow{5}{*}{\rotatebox[origin=c]{90}{\texttt{gemma-2}}}
&\cellcolor{gray!15}Direct &\cellcolor{gray!15}9B  & \cellcolor{gray!15}--  & \cellcolor{gray!15}3.0  & \cellcolor{gray!15}3.5  & \cellcolor{gray!15}21.0 \\

&\cellcolor{gray!15}CoT &\cellcolor{gray!15}9B  & \cellcolor{gray!15}--  & \cellcolor{gray!15}66.6  & \cellcolor{gray!15}26.4  & \cellcolor{gray!15}79.3 \\

&\cellcolor{gray!15}CoT &\cellcolor{gray!15}9B-Instruct & \cellcolor{gray!15}--  & \cellcolor{gray!15}87.6  & \cellcolor{gray!15}43.5  & \cellcolor{gray!15}85.3 \\

\cmidrule(lr){2-7}

&Direct &9B  & 14B  & 60.5  & 24.2  & 76.7 \\
&Direct &9B & 7B  & 43.7  & 26.3  & 72.3 \\

\midrule

\multirow{4}{*}{\rotatebox[origin=c]{90}{\texttt{OLMo-2}}}
&\cellcolor{gray!15}Direct &\cellcolor{gray!15}7B  & \cellcolor{gray!15}--  & \cellcolor{gray!15}10.0  & \cellcolor{gray!15}9.7  & \cellcolor{gray!15}43.7 \\

&\cellcolor{gray!15}CoT &\cellcolor{gray!15}7B  & \cellcolor{gray!15}--  & \cellcolor{gray!15}53.8 & \cellcolor{gray!15}15.3  & \cellcolor{gray!15}71.0 \\

&\cellcolor{gray!15}CoT &\cellcolor{gray!15}7B-Instruct & \cellcolor{gray!15}--  & \cellcolor{gray!15}79.4  & \cellcolor{gray!15}24.9  & \cellcolor{gray!15}78.7 \\

\cmidrule(lr){2-7}

&Direct &7B  & 14B  & 51.6  & 15.5  & 58.7 \\

\bottomrule
\end{tabular}
\end{table}

\end{document}